\documentclass[letterpaper]{article} %
\usepackage{aaai25}  %
\usepackage{times}  %
\usepackage{helvet}  %
\usepackage{courier}  %
\usepackage[hyphens]{url}  %
\usepackage{graphicx} %
\usepackage{natbib}  %
\usepackage{caption} %
\usepackage{algorithm}
\usepackage{algorithmic}
\usepackage{pifont}
\usepackage{color}

\usepackage{newfloat}
\usepackage{listings}
\DeclareCaptionStyle{ruled}{labelfont=normalfont,labelsep=colon,strut=off} %
\floatstyle{ruled}
\newfloat{listing}{tb}{lst}{}
\floatname{listing}{Listing}
\usepackage[export]{adjustbox} %
\usepackage{booktabs} %

\usepackage{amssymb}
\usepackage{mathtools}
\usepackage{stmaryrd}
\usepackage{tabularx}
\usepackage{makecell}

\usepackage{url}            %
\usepackage{booktabs}       %
\usepackage{amsfonts}       %
\usepackage{nicefrac}       %
\usepackage{microtype}      %
\usepackage{multirow}       %

\usepackage{enumitem}%
\usepackage{comment}
\usepackage{amsmath,amssymb} %
\usepackage{symbols}

\usepackage{amsmath,amsfonts,bm}

\def\1{\bm{1}}

\newcommand{\test}{\mathcal{D_{\mathrm{test}}}}

\def\eps{{\epsilon}}
\def\sp{space}

\def\rvc{{\mathbf{c}}}

\def\rvs{{\mathbf{s}}}
\def\rvt{{\mathbf{t}}}

\def\rvx{{\mathbf{x}}}

\def\rmC{{\mathbf{C}}}

\def\rmM{{\mathbf{M}}}

\def\rmO{{\mathbf{O}}}

\def\rmQ{{\mathbf{Q}}}

\def\rmS{{\mathbf{S}}}

\def\rmW{{\mathbf{W}}}

\def\vc{{\bm{c}}}

\def\vt{{\bm{t}}}

\def\mQ{{\bm{Q}}}

\def\mW{{\bm{W}}}

\DeclareMathAlphabet{\mathsfit}{\encodingdefault}{\sfdefault}{m}{sl}
\SetMathAlphabet{\mathsfit}{bold}{\encodingdefault}{\sfdefault}{bx}{n}

\def\gA{{\mathcal{A}}}

\def\gC{{\mathcal{C}}}

\def\gI{{\mathcal{I}}}

\def\gL{{\mathcal{L}}}
\def\gM{{\mathcal{M}}}
\def\gN{{\mathcal{N}}}

\def\gS{{\mathcal{S}}}

\def\gW{{\mathcal{W}}}
\def\gX{{\mathcal{X}}}

\def\sR{{\mathbb{R}}}

\newcommand{\R}{\mathbb{R}}

\DeclareMathOperator*{\argmin}{arg\,min}

\usepackage[multiple]{footmisc}

\newcommand{\myparagraph}[1]{\vspace*{0pt}{\bf #1}}

\definecolor{imma}{rgb}{0.23, 0.45, 0.69}

\newcommand{\msgr}{\texttt{MSGR}\xspace}

\newcommand{\mrsgr}{\texttt{MRSGR}\xspace}

\newcommand{\bsla}{\texttt{JT}\xspace}
\newcommand{\bslb}{\texttt{CP}\xspace}
\newcommand{\ml}{\texttt{Merge}\xspace}

\title{Multi-concept Model Immunization through Differentiable Model Merging}

\author {
    Amber Yijia Zheng \quad\quad
    Raymond A. Yeh
}
\affiliations {
    Department of Computer Science, Purdue University \\
    {\tt\small \{zheng709,rayyeh\}@purdue.edu}
}

\begin{document}
\twocolumn[{%
\renewcommand\twocolumn[1][]{#1}%
\maketitle
\centering
\vspace{-1cm}
\setlength{\tabcolsep}{3pt}
\begin{tabular}{ccc}
\includegraphics[width=5.5cm,trim={0.6cm 0 0.2cm 0},clip]{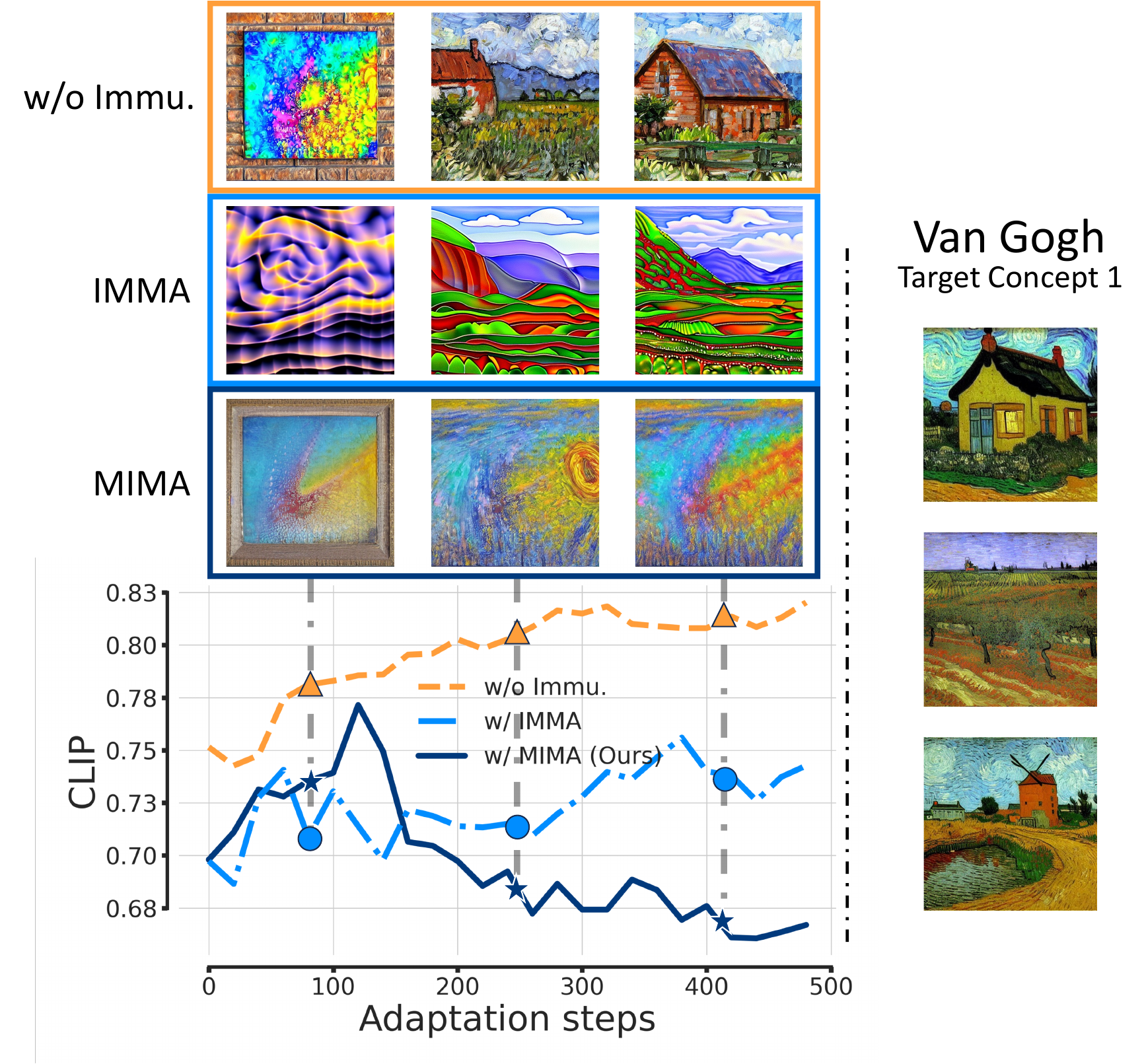} & 
\includegraphics[width=5.5cm,trim={0.6cm 0 0.2cm 0},clip]{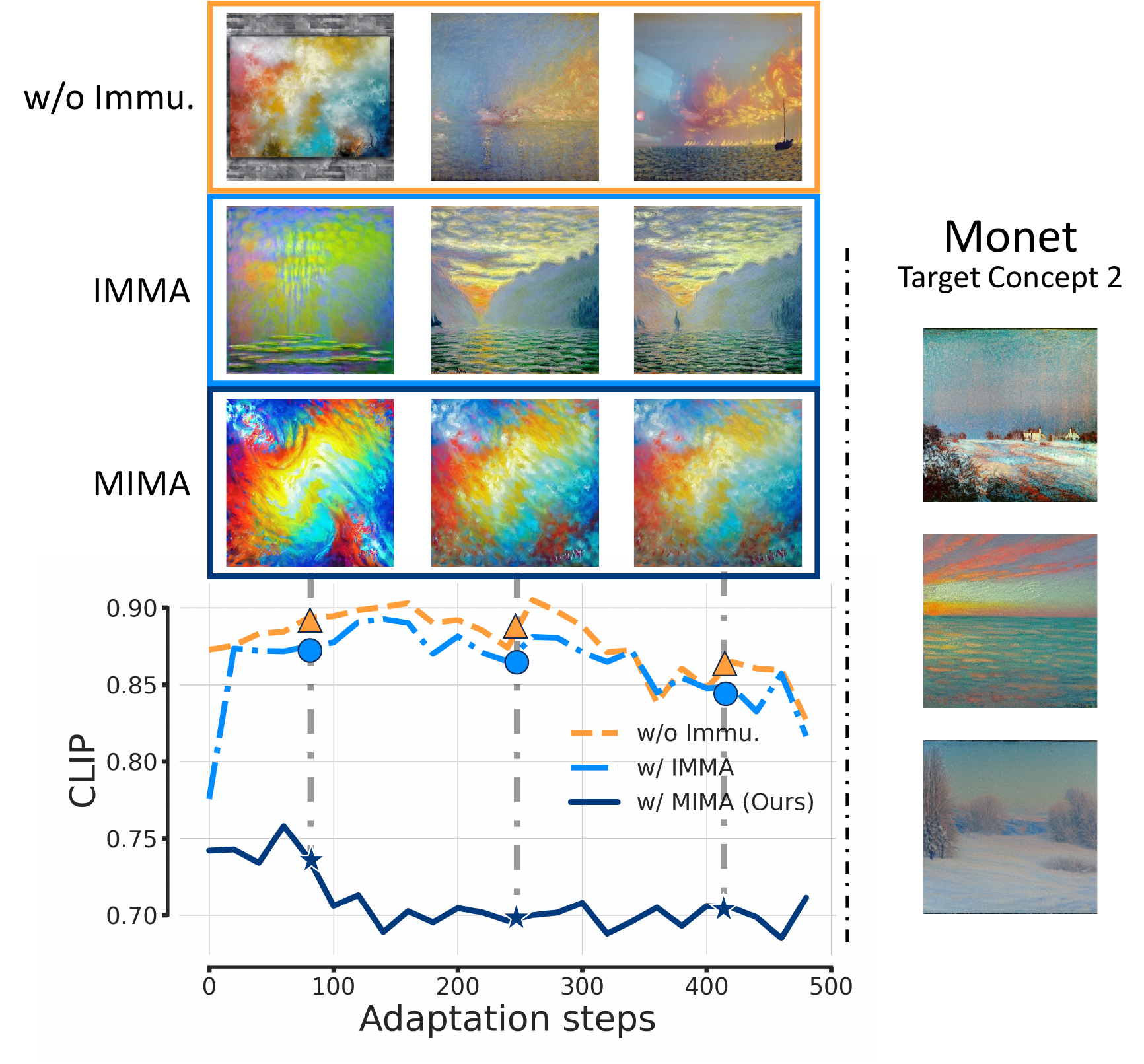} &
\includegraphics[width=5.5cm,trim={0.6cm 0 0.2cm 0},clip]{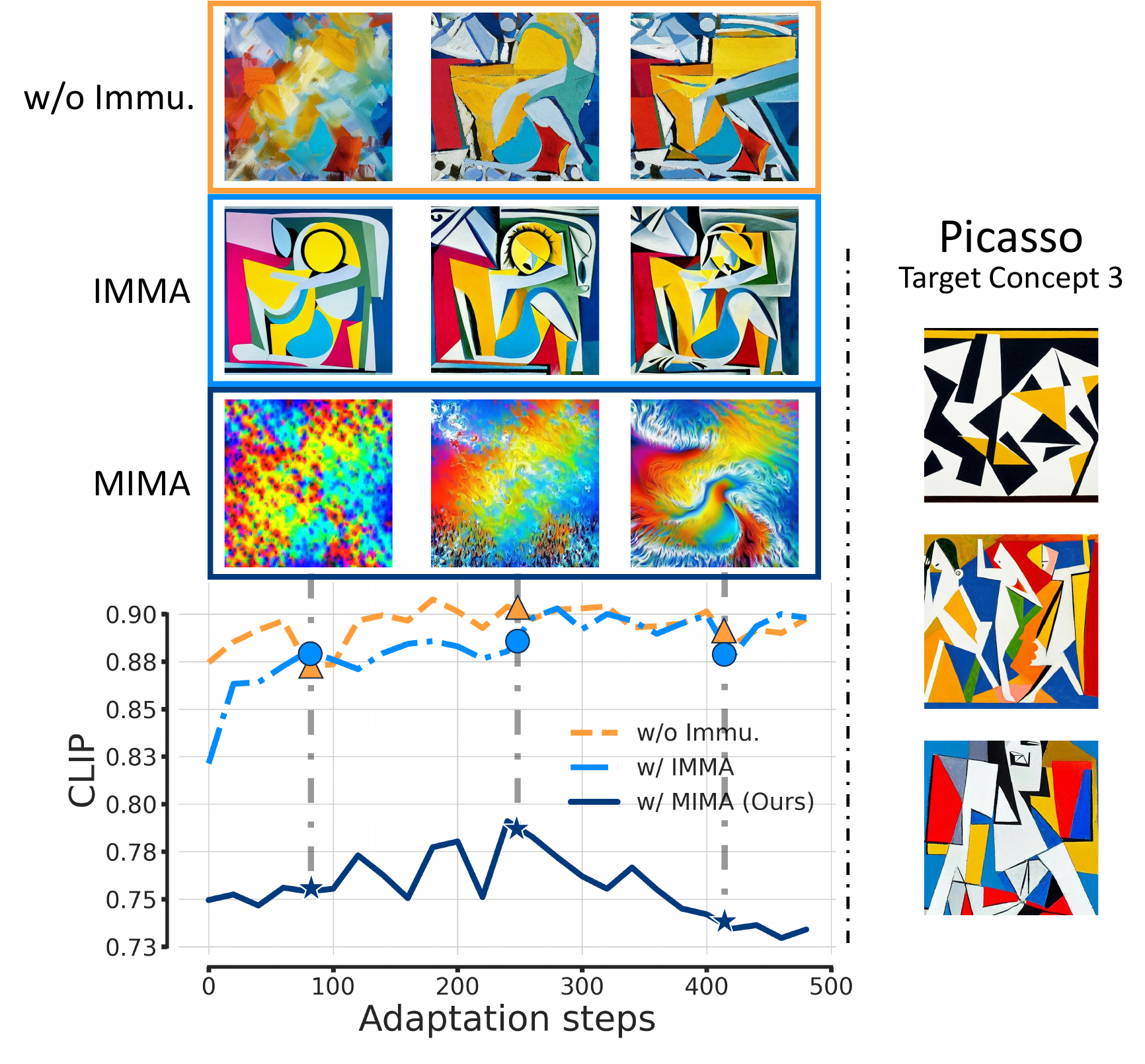}
\end{tabular}
\vspace{-0.27cm}
\captionof{figure}{We propose MIMA an immunize algorithm that protects a model against the adaptation on harmful concepts. Here we show an experiment on immunization against the re-learning of multiple artistic styles and report the CLIP similarity between the generations and the target concept at different adaption steps. A lower CLIP similarity indicates a more effective immunization, as the images semantically differ more from the references. As can be seen, MIMA offers protection over all three concepts of Van Gogh, Monet, and Picasso. In comparison, IMMA~\cite{zheng2023imma}, designed to immunize over a single concept, only offers protection against the re-learning of Van Gogh.
}
\label{fig:teaser}
\vspace{0.5cm}

}]
\begin{abstract}
Model immunization is an emerging direction that aims to mitigate the potential risk of misuse associated with open-sourced models and advancing adaptation methods. The idea is to make the released models' weights difficult to fine-tune on certain harmful applications, hence the name ``immunized''. Recent work on model immunization focuses on the \textit{single-concept} setting. However, models need to be immunized against multiple concepts in real-world situations. To address this gap, we propose an immunization algorithm that, simultaneously, learns a single ``difficult initialization'' for adaptation methods over \textit{a set of concepts}. We achieve this by incorporating a differentiable merging layer that combines a set of model weights adapted over multiple concepts.
In our experiments, we demonstrate the effectiveness of multi-concept immunization by generalizing prior work's experiment setup of re-learning and personalization adaptation to multiple concepts.

\begin{links} 
\link{Project page}{https://www.amberyzheng.com/mima}
\end{links}
\end{abstract}

\section{Introduction}

With the advancements in effective adaptation techniques, such as DreamBooth~\cite{ruiz2022dreambooth} or Textual Inversion~\cite{gal2022textual}, there are increasing risks of misuse for open-sourced text-to-image models. As the models are released, an ill-intended person can leverage adaptation methods to tune unsafe content into the models and perform malicious acts, \eg, generating unsafe or sexual content~\cite{harwell2023wash}. 

To tackle these risks,~\citet{zheng2023imma} propose to ``immunize'' the open-sourced models before releasing them. The idea is to learn models that are resistant (``immuned'') to adaptations on harmful concepts. \citet{zheng2023imma} refer to their approach as \textit{Immunizing text-to-image Models against Malicious Adaptation}, in short, IMMA.
While IMMA shows a promising direction for mitigation, IMMA's method and experiments focus on the immunization of a \textit{single concept.} However, in most practical settings, a model needs to be immunized against multiple harmful concepts. 

To address this gap, we study Multi-concept Immunization against Malicious Adaptation (MIMA).
We aim to make a \textit{single model} resilient to adaptation on more than one concept. We propose a model immunization algorithm that 
meta-learns a ``difficult initialization'' for adaptation methods over \textit{a set of concepts} formulated as a bi-level optimization with multiple lower-level tasks. We accomplish this by introducing a differentiable model merging layer that combines the individual lower-level task's weights from each target concept.
The bi-level optimization is solved by backpropagating through this merging layer to immunize the model over a set of concepts. 
This approach is inspired by the success of model merging for multi-concept customization~\cite{kumari2022customdiffusion}, we hypothesize that model merging would also benefit immunization as it captures the relationships among concepts.

Empirically, we experiment with several adaptation methods, including, Textual Inversion~\cite{gal2022textual}, DreamBooth~\cite{ruiz2022dreambooth}, LoRA~\cite{hu2022lora}, and CustomDiffusion~\cite{kumari2022customdiffusion} over two applications: (a) restoring erased concepts such as artistic styles or object categories, and (b) learning personalized concepts. We found that MIMA successfully immunizes a model against \textit{multiple} malicious concepts and outperforms IMMA-inspired baselines.

{\noindent\bf Our contributions are summarized as follows:}
\begin{itemize}[topsep=2pt]
\item We generalize the task of model immunization from a single concept to multiple concepts that more closely match the real-world scenario.
\item We propose MIMA, a novel model immunization algorithm for multi-concept immunization. MIMA leverages a differentiable model merging layer that combines multiple adapted weights, enabling backpropagation to meta-learn an immunized model. 
\item We conduct experiments over two tasks and four adaptation methods demonstrating the efficacy of MIMA.
\end{itemize}

\section{Related Work}

\myparagraph{Towards safer generative AI.} 
Several directions have been proposed to make generative AI safer. One direction that has received attention is removing inappropriate content from pre-trained models~\cite{schramowski2023safe, gandikota2023erasing,gandikota2024unified,zhang2023forgetmenot,kumari2023conceptablation,heng2023selective}. Another direction is to protect the data sources by using adversarial examples~\cite{goodfellow2015explaining} to achieve data-poisioning~\cite{biggio2011support,mei2015using}, such that when adapted on these protected images, the diffusion model fails~\cite{shan2023glaze,liang2023adversarial, liang2023mist,zhao2023unlearnable}. However, these approaches have limitations when dealing with open-sourced models. For content removal, adaptation methods can quickly relearn the removed content. For data poisoning techniques, it requires poisoning the content which may not be feasible depending on the content sources. 

Most related to our work is the model immunization paradigm introduced by~\citet{zheng2023imma} which aims to learn poor initialization such that adaptation methods fail on a single concept. In this work, we generalize the study of model immunization to multi-concept and propose an algorithm MIMA that incorporates a differentiable model merging layer to enable multi-concept immunization. Also, model immunization has been considered in the few-shot classification setting~\cite{zheng2025learning}.

\myparagraph{Model adaptation and editing methods.}
With the open-source of high-quality text-to-image models, \eg, Stalbe Diffusion~\cite{rombach2022high,deep-floyd-if}, there is a surge of interest in how to adapt these pre-trained models for different applications, \eg, adapting the model for a customized generation of personalized items~\cite{ruiz2022dreambooth, gal2022textual, kumari2022customdiffusion}, efficient fine-tuning of these models~\cite{hu2022lora}, or adding extra control to the generation~\cite{zhang2023adding}. Closely related is model editing which aims to directly modify model parameters to achieve new generative capabilities~\cite{bau2020rewriting,gal2022stylegan,kumari2022customdiffusion,nitzan2023domain}.

\myparagraph{Optimization layers.} 
We briefly discuss optimization layers as the differentiable model merging can be viewed as an optimization layer. The literature of optimization layers views an optimization problem as a differentiable function, \ie, mapping its input to its exact solution~\cite{Domke_2012,Amos_Kolter_2017,ren2020not,Gould_Hartley_Campbell_2021,Agrawal_Amos_Barratt_Boyd_Diamond_Kolter_2019,liu2023differentiable}. Depending on the exact optimization program, the gradient of this mapping can either be computed analytically or via implicit differentiation. Optimization layers have found applications in AI~\cite{Amos_Kolter_2017,Tschiatschek_Sahin_Krause_2018,Wang_Donti_Wilder_Kolter_2019, Amos_Jimenez_Sacks_Boots_Kolter_2018,zheng2024graph} and computer vision~\cite{yeh2022total,hu2023surface,Bai_Kolter_Koltun_2019,Bai_Koltun_Kolter_2020,Rolinek_Swoboda_Zietlow_Paulus_Musil_Martius_2020,wang2023deep,geng2023one,zheng2025learning}. 

\section{Background}
\label{sec:prelim}

\myparagraph{Model immunization.}
Given pre-trained diffusion model weights $\theta^p$, an adaptation method $\gA$, and its corresponding loss function $L_\gA$, IMMA~\cite{zheng2023imma} aims to prevent $\gA$ from fine-tuning $\theta^p$, such that the fine-tuned model fails to generate images of a single target (harmful) concept $\rvc'$. IMMA is formulated as a bi-level optimization with the following objective:
\bea
\label{eq:imma}
\underbrace{\max_{\theta_{\in\gS}} L_{\gA}(\rvx'_{\gI}, \rvc'; \theta, \phi^\star)}_{\text{upper-level task}}
\text{~s.t.~}\phi^\star =
\underbrace{
\argmin_\phi L_{\gA}(\rvx'_{\gA}, \rvc'; \theta, \phi)}_{\text{lower-level task}}.
\eea%
Intuitively, the upper-level task aims to find the worst parameters $\theta$ for the adaptation algorithm $\gA$ by taking into the $\gA$'s update in the lower-level task. Note that, IMMA requires the following: (a) $\gA$ is known at immunization time, and (b) there is a single-concept $\vc'$ to be immunized. In this work, we propose an immunization algorithm that does not require $\gA$ and immunizes a model over \textit{a set} of concepts.

\myparagraph{Model merging.} To achieve a unified fine-tuned model capable of generating multiple target concepts, one approach is to fine-tune multiple models for each concept and combine them. \citet{kumari2022customdiffusion} propose to merge models by solving an optimization problem, where the keys and values of cross-attention weights~\cite{dosovitskiy2020image,vaswani2017attention} are modified then merged. Recall, cross attention layers take in text embeddings $\rvc \in \R^{l \times c}$, where $l$ is the number of tokens with $c$ denoting the embedding size, and project them into keys, values with the projection matrics $\rmW^k \in \sR^{c \times d}$ and $\rmW^v \in \sR^{c \times d'}$, where $d$ and $d'$ are the dimensions of the keys and values. We subsume these projection matrices into a compact notation $\rmW$.

\begin{figure*}[t]
\vspace{-0.05cm}
\centering
\small
\setlength{\tabcolsep}{8pt}
\begin{tabular}{c@{\hskip 1.5cm}c}
\bf Multi-concept Immunization (\algref{alg:mima}) & \bf Generation \\
\includegraphics[height=5.7cm]{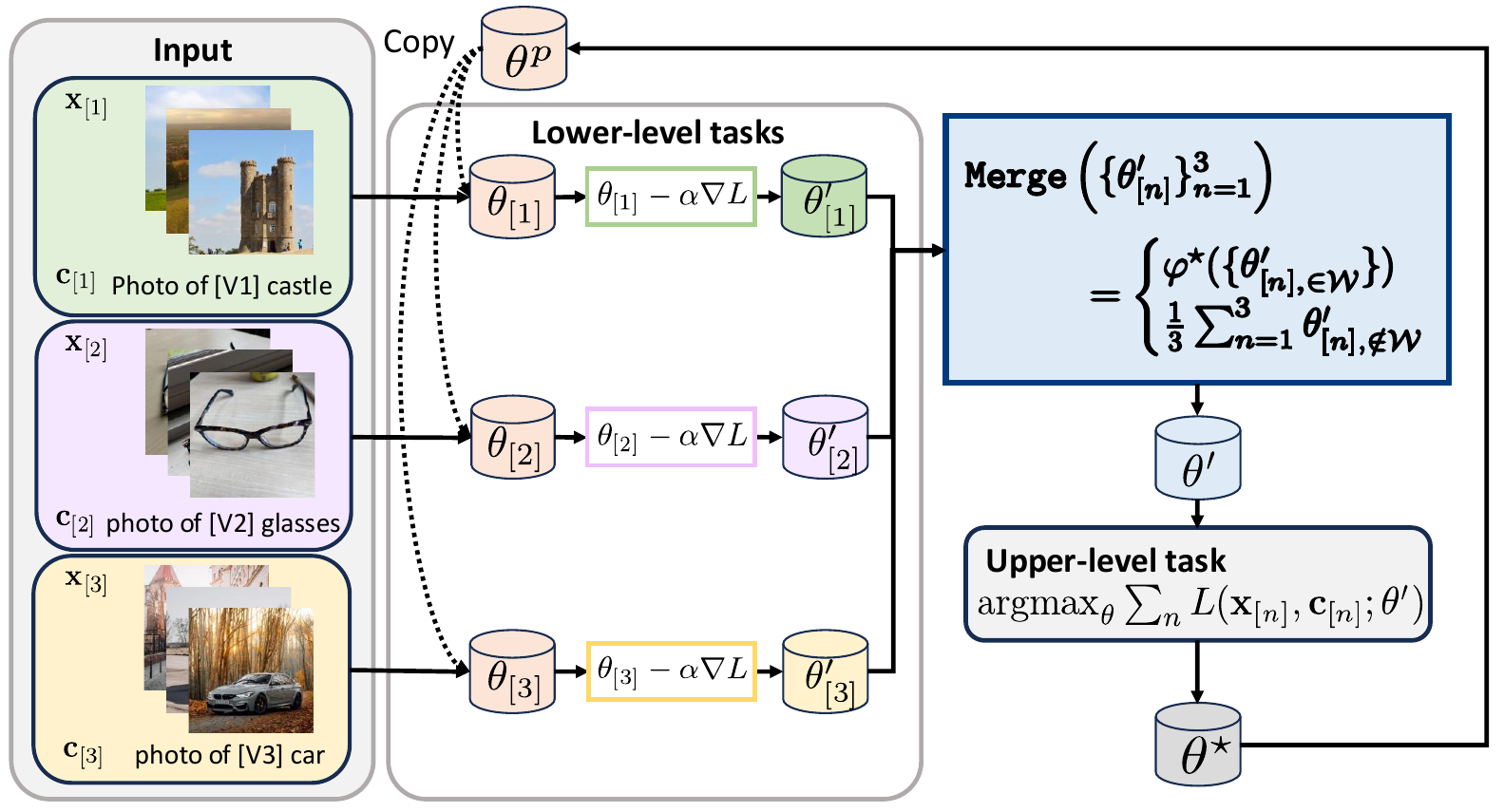} &
\includegraphics[height=5.7cm]{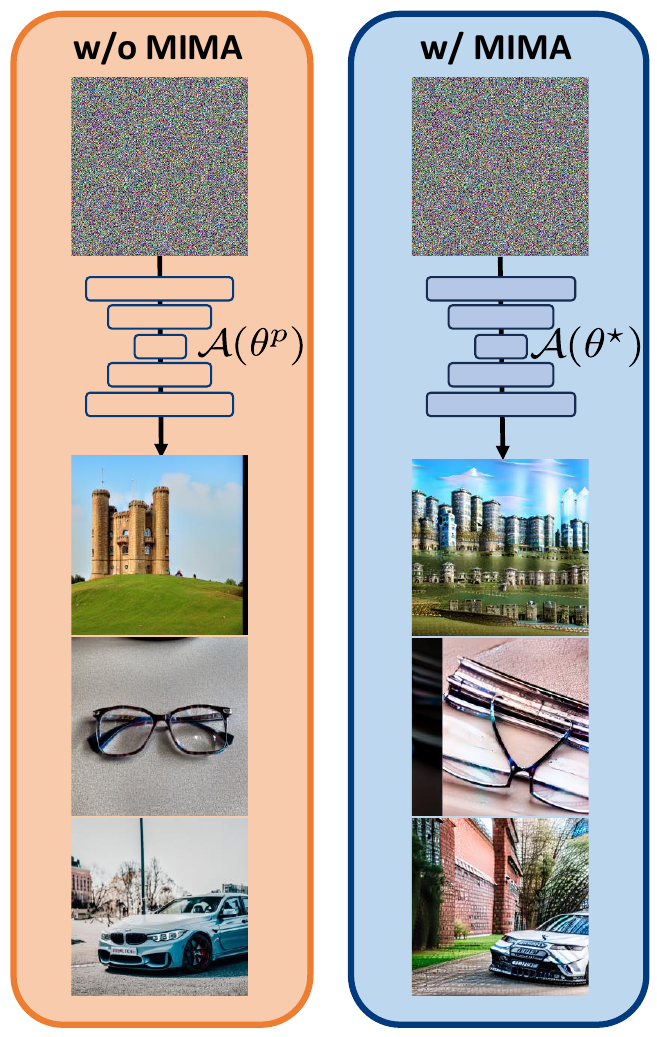}
\end{tabular}
\vspace{-0.15cm}
\caption{Method overview. \textit{Left:} MIMA is formulated as a bi-level optimization program. 
For the lower-level, we unroll loss $L$ for the copied weights of each concept. Next, we combine the individual weights $\theta'_{[n]}$ via our proposed {\tt Merge} layer defined in~\equref{eq:my_merge}. For the upper-level, we maximize the diffusion loss $L$ with respect to the parameters $\theta$ by backpropagating through $\theta'$. \textit{Right:} During generation, a model $\theta^\star$ immunized with MIMA fails to be adapted by $\gA$ on all of the target concepts, \ie, the generations do not contain good quality images of castles, glasses, or cars.
}
\label{fig:intro_comp}
\vspace{-0.2cm}
\end{figure*}

Given $N$ model weights $\{\rmW_{[n]}\}_{n=1}^N$, that has each been adapted to a concept embedding $\rvc_{[n]}$ from the target concept set $\gC$,
model merging aims to find a single optimal weight $\varphi^\star$ that mimics the mapping of all weights on their corresponding concepts while maintaining proximity to a set of $N'$ regularization concepts $\gC_{\text{reg}}$. This merging process is formulated as a constrained optimization program:
\bea
&\varphi^\star = \argmin_{\varphi}  ||\rmC_{\text{reg}} \varphi  - \rmC_{\text{reg}} \rmW^p||_F^2 \text{~~s.t.~~}\rmC \varphi = \rmO^\ast, \\
\nonumber
&\text{~where~} \rmC \triangleq {\tt Concat}(\gC)\in \mathbb{R}^{(N \cdot l) \times c}\text{~,~} \\ 
\nonumber
&\rmC_{\text{reg}} \triangleq {\tt Concat}(\gC_{\text{reg}})\in \mathbb{R}^{(N' \cdot l) \times c}\text{~,~} \\ 
\nonumber
&\text{~~~~~~~~and~}\rmO^\ast \triangleq {\tt Concat}\left(\{\rvc_{[n]}\rmW_{[n]}\}_{n=1}^N\right) \in \mathbb{R}^{(N \cdot l) \times d}.
\label{eq:optim_layer}
\eea
Here ${\tt Concat(\cdot)}$ concatenates a set $\gS$ of embeddings into a stack of embeddings $\rmS = [\rvs_{[1]}^\top, \cdots, \rvs_{[N]}^\top]^\top$ and $\rmW^p$ corresponds to the pre-trained weight of a model before the adaptation. 

The constraint matches the output of $\varphi$ on target concepts fine-tuned for each of the $\mW_{[n]}$. Next, the objective maintains the model's generation capability on the other concepts not in $\gC$. It encourages the model $\varphi$ to be similar to the corresponding pre-trained weight $\rmW^p$ on a set of regularization concepts $\gC_{\text{reg}}$.

\section{Approach}
We introduce Multi-concept model Immunization against Malicious Adaptation (MIMA). As the name suggests, the goal is to protect a pre-trained model with weights $\theta^p$ from being fine-tuned by adaptation methods to generate images containing harmful concepts. Formulated as a bi-level optimization, MIMA meta-learns a difficult initialization for downstream adaptation methods on all the concepts within a target concept set. The key idea is to treat model merging as a differentiable optimization layer that allows for gradients to be backpropagated through model merging to provide updated directions for immunization. Please see overview in~\figref{fig:intro_comp}.

\subsection{Multi-concept Model Immunization}
\myparagraph{Problem formulation.} As in IMMA~\cite{zheng2023imma}, the general immuimization process is formulated as a bi-level optimization problem. Given pre-trained model weights $\theta^p$, a set of target concept embeddings $\gC = \{\rvc_{[n]}\}_{n=1}^N$, and the image set $\gX = \cup_n \gX_{[n]}$ where $ \gX_{[n]} = \{\rvx_{[n]}\}$ is a set of images representative of the concept $\rvc_{[n]}$, we optimize:%
\bea\nonumber
        && \hspace{-1cm}\underbrace{\max_{\theta_{\in \gS^u}} \sum_{n=1}^{|\gC|} L(\rvx^{u}_{[n]}, \rvc_{[n]}; \texttt{Merge}\left(\left\{ \theta'_{[n]} \right\}\right) )}_{\text{upper-level task}}, \\
        && \hspace{-.35cm} ~~\text{s.t.}~~ \underbrace{
        \theta'_{[n]} \triangleq \argmin_{\theta_{\in \gS^l}} L(\rvx^{l}_{[n]}, \rvc_{[n]}; \theta)~~\forall n
        }_{\text{multiple lower-level tasks}},
        \label{eq:mima}
\eea
where $\rvx^u_{[n]}$ and $\rvx^l_{[n]}$ are independently sampled from $\gX_{[n]}$. The sets $\gS^u$ and $\gS^l$ denote the subset of model parameters that are being updated in the upper and lower tasks. Next, the \texttt{Merge} function, defined formally in~\equref{eq:my_merge}, combines a set of model weights into a single model, 
and 
$L$ denotes the standard loss for training a diffusion model given by
\bea \label{eq:sd}
    L(\rvx, \rvc; \theta) = \sE_{t, \eps \sim \gN(0, I)} \left[ w_t \|\eps_\theta(\rvx_t, \rvc, t) - \eps\|_2^2 \right].
\eea
Here, $\eps_\theta$ is the denoising network with weights $\theta$ conditioned on the timestep $t$ sampled from a discrete uniform distribution, $\rvx_t$ is the noisy image, and $w_t$ is a loss weight.

To achieve multi-concept immunization, the upper-level task aims to make the standard diffusion loss~(\equref{eq:sd}) high when being adapted to any of the target concepts. Hence, in the lower-level tasks, we perform a set of updates, one for each concept $\vc_{[n]}$, leading to $N$ different models $\theta'_{[n]}$. 

However, model immunization requires a single model. It is unclear how to perform the maximization in the upper-level task given $N$ separate models. Specifically, we need to merge these $N$ models into one. 
The main challenges are: {\bf (a)} How to merge these $N$ models? The procedure proposed by~\citet{kumari2022customdiffusion} only merges the projection matrices of keys and values, what about the other parameters? {\bf (b)} How do we backpropagate through the model merging operation, such that gradient-based optimization can be performed? We now answer these two questions.

\myparagraph{Differentiable model merging.} 
To merge the weights that are fine-tuned on different concepts, we split the model parameters into two sets: key and value project matrices subsumed in $\gW$ and the rest $\notin \gW$. For parameters within $\gW$, we combine the parameters following the optimization in~\equref{eq:optim_layer}, and for the other parameters, we perform a simple average. More formally, the \texttt{Merge} operation is defined as:
\bea
\theta' \triangleq \ml\left(\{\theta'_{[n]}\}_{n=1}^N\right) \triangleq \begin{cases}
    \varphi^\star(\{\theta'_{[n], \in \gW}\}) \\
    \frac{1}{N} \sum_{n=1}^{N} \theta'_{[n], \notin \gW}
\end{cases},
\label{eq:my_merge}
\eea
where we view the solution $\varphi^\star$ in~\equref{eq:optim_layer} \textit{as a function of the input} model weights to the optimization problem. The reason why we perform a simple average over the parameters $\notin \gW$ is that they are shared across the different lower-level tasks. With a simple average, \ie, the gradients will also be shared.

\begin{figure}[t]
\vspace{-.3cm}
\begin{algorithm}[H]
\caption{MIMA (Our method)}
\begin{algorithmic}[1]
\renewcommand{\algorithmicrequire}{\textbf{Input:}}
\renewcommand{\algorithmicensure}{\textbf{Output:}}
\REQUIRE pre-trained model $\theta^p$, images $\gX = \cup_n \gX_{[n]}$ concepts $\gC = \{\rvc_{[n]}\}_{n=1}^N$, learning rates $\alpha$ and $\beta$, modified parameters set $\gS^l$ and $\gS^u$ in lower and upper tasks, loss function $L$,  $\ml$ layer, training epochs $K$
\ENSURE  Immunized model $\theta^\star$
\STATE Initialize $\theta^0 = \theta^p$
\FOR {$k = 1$ to $K$}
\STATE Sample batches of each concept $\{(\rvx^u_{[n]}, \rvc_{[n]})\}_{n = 1}^N$ from $\gX$ and $\gC$
\STATE {\it \color{imma}\# Solve the lower-level tasks for one step.}
\FOR {$n = 1$ to $N$}
\STATE Sample batch $\rvx^l_{[n]}$ from $\gX_{[n]}$
\STATE $\theta'_{[n], \in \gS^l} \leftarrow \theta^{k-1}_{\in \gS^l} - \alpha \nabla_{\theta} L(\rvx^l_{[n]}, \rvc_{[n]}; \theta^{k-1})$
\ENDFOR
\STATE {\it \color{imma}\# Each $\theta'_{[n]}$ is a function of $\theta^{k-1}$}
\STATE $\theta' \leftarrow \ml(\{\theta_{[1]}', \dots, \theta_{[N]}' \})$ 
\STATE $\theta^{k}_{\in \gS^u} \leftarrow \theta^{k-1}_{\in \gS^u} + \beta \nabla_\theta L(\rvx^u_{[n]}, \rvc_{[n]}; \theta')$
\ENDFOR
\STATE $\theta^\star \leftarrow \theta^K$
\RETURN $\theta^\star$
\end{algorithmic}
\label{alg:mima}
\end{algorithm}
\vspace{-0.8cm}
\end{figure}

To backpropagate through~\equref{eq:my_merge}, we need to compute the gradient through $\varphi^\star$. %
To obtain $\varphi^\star$ from~\equref{eq:optim_layer}, we can solve its Lagrange form of:
\bea
\label{eq:lagrange_form}
L(\varphi, \rmM) = ||\rmC_{\text{reg}} \varphi  - \rmC_{\text{reg}} \rmW^p||_F^2 - \text{tr}((\rmC \varphi - \rmO^\ast)\rmM^\top)
\eea
where $\rmM \in \mathbb{R}^{(N \times l) \times d}$ represents the matrix of Lagrange multipliers associated with the constraint. 
Taking a closer look at~\equref{eq:lagrange_form}, we can express it as a linear system  $\mQ\varphi=\vt$ with
\bea
\label{eq:linear_reg_form}
\mQ \triangleq \rmC_{\text{reg}}^\top \rmC_{\text{reg}}
~\text{ and }~
\vt \triangleq  \rmC_{\text{reg}}^\top \rmC_{\text{reg}} \rmW^p + \frac{1}{2}\rmC^\top \rmM,\\
\label{eq:optim_m}
~\text{ where }~ \rmM = 2\left(\rmC(\rmC_{\text{reg}}^\top\rmC_{\text{reg}})C^\top\right)^{-1}(\rmO^* - \rmC \rmW^p).
\eea

In other words, the solution has the form
\bea
\varphi^\star = \mQ^{-1}\vt ~\text{ and }~ \frac{\partial \gL}{\partial \varphi^\star} = \mQ^\intercal\frac{\partial \gL}{\partial \vt}
\eea
from chain-rule. 
We note that this gradient has been previously studied in the optimization layer literature~\cite{Amos_Kolter_2017,Barratt_Boyd_2021} in more generic forms.

Putting everything together,
from chain rule, the gradient of $L$ \wrt to $\theta$ is:
\bea
\label{eq:grad}
\frac{\partial L(\theta')}{\partial \theta} = \frac{\partial L}{\partial \theta'} \cdot  \sum_{n=1}^{N}  \frac{\partial \theta'}{\partial \theta'_{[n]}} \cdot \frac{\partial \theta'_{[n]}}{\partial \theta},
\eea
where $\frac{\partial \theta'}{\partial \theta'_{[n], \in \gW}}$ is computed through $\varphi^\star$ and $\frac{\partial \theta'}{\partial \theta'_{[n], \notin \gW}}$ is a scaled identity matrix. %

\myparagraph{Solving bi-level optimization.} 
We solve the bi-level optimization program in~\equref{eq:mima} using gradient-based methods~\cite{maclaurin2015gradient,shaban2019truncated} commonly used in meta-learning~\cite{finn2017model}. We provide a summary in~\algref{alg:mima}.
The lower-level tasks in~\equref{eq:mima} are solved approximately per $\theta'_{[n]}$ with a single step of gradient update. After collecting all $\{\theta_{[n]}' \forall n\}$, we aggregate these weights with the proposed $\ml$ layer, which leads to an aggregated parameter $\theta'$ as a function of the original $\theta$. Next, we iteratively solve the upper-level task using gradient descent by backpropagating through $\theta'$ to update $\theta$, \ie, unrolled gradient. 

\section{Experiments}
As in IMMA~\cite{zheng2023imma}, we consider two categories of malicious adaptation:~\ding{182} immunization for protecting against re-learning concepts from an erased model and~\ding{183}  immunization against personalized content. Different from IMMA, we generalize immunization experiment settings to multiple concepts.

\begin{figure*}[t]
    \centering
    \setlength{\tabcolsep}{0.5pt}
    \begin{tabular}{cc@{\hskip 3pt}|@{\hskip 3pt} ccc}
    \multicolumn{2}{c}{\textbf{2-concept}} & \multicolumn{3}{c}{\textbf{3-concept}} \\
      Kelly Mckernan  & Kilian Eng & Van Gogh & Claude Monet & Pablo Picasso  \\
     \includegraphics[height=2.0cm, trim={10 0.4cm 10 0.4cm},clip]{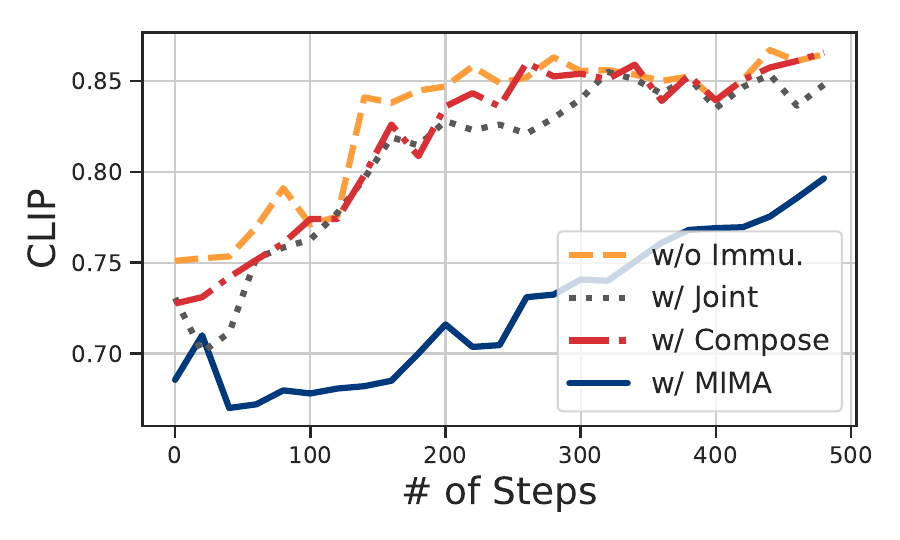} & 
     \includegraphics[height=2.0cm, trim={10 0.4cm 10 0.4cm},clip]{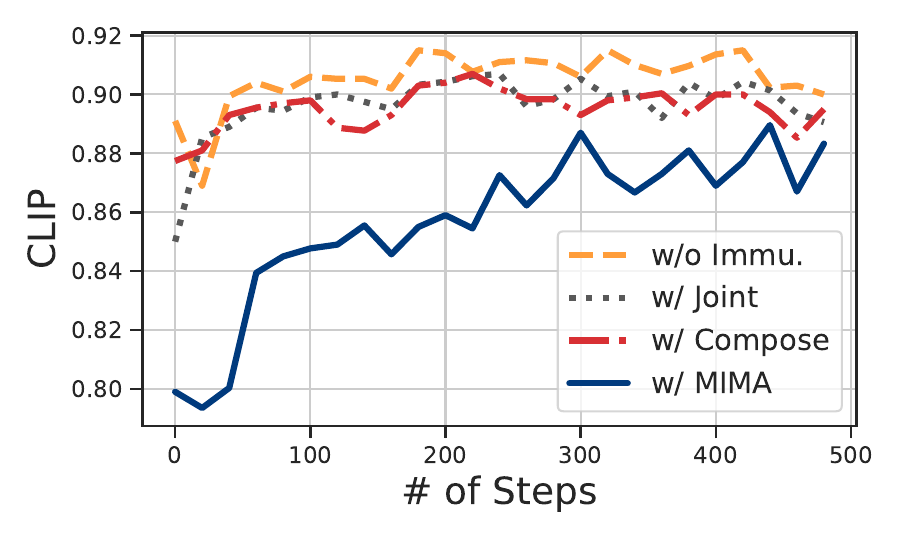} &
     \includegraphics[height=2.0cm, trim={10 0.4cm 10 0.4cm},clip]{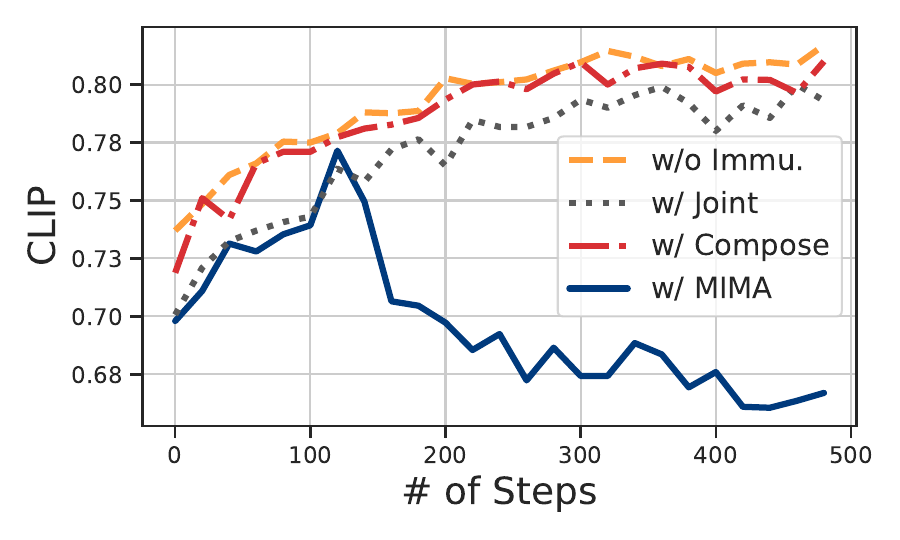} & \includegraphics[height=2.0cm, trim={10 0.4cm 10 0.4cm},clip]{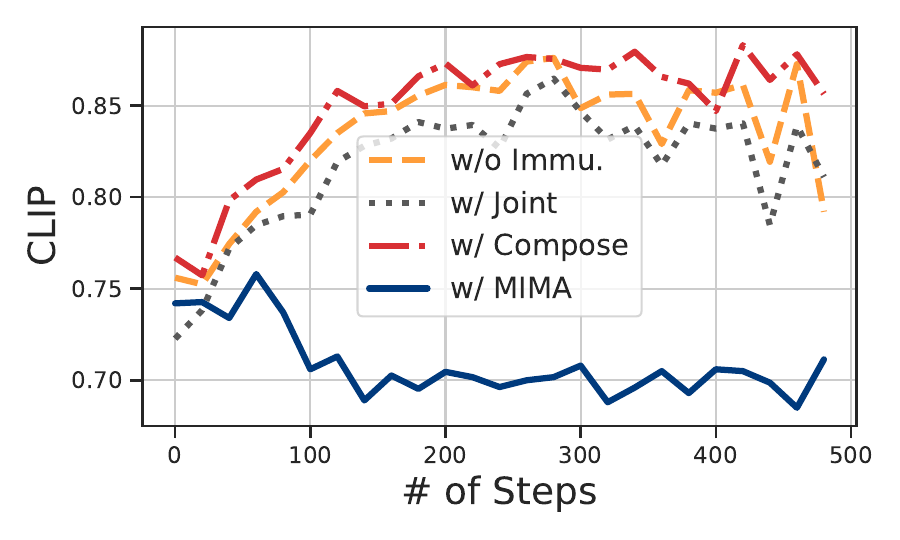} &
     \includegraphics[height=2.0cm, trim={10 0.4cm 10 0.4cm},clip]{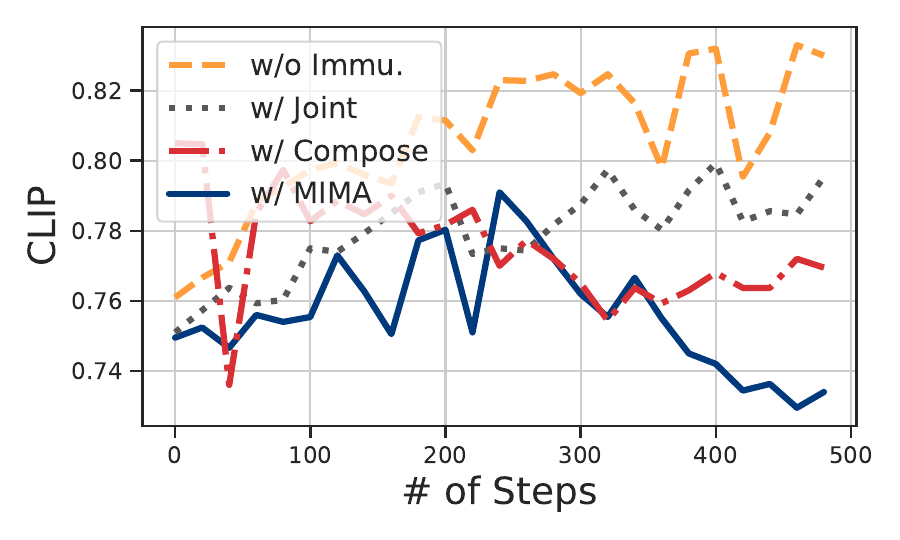} \\
     \includegraphics[height=2.0cm, trim={10 0.4cm 10 0.4cm},clip]{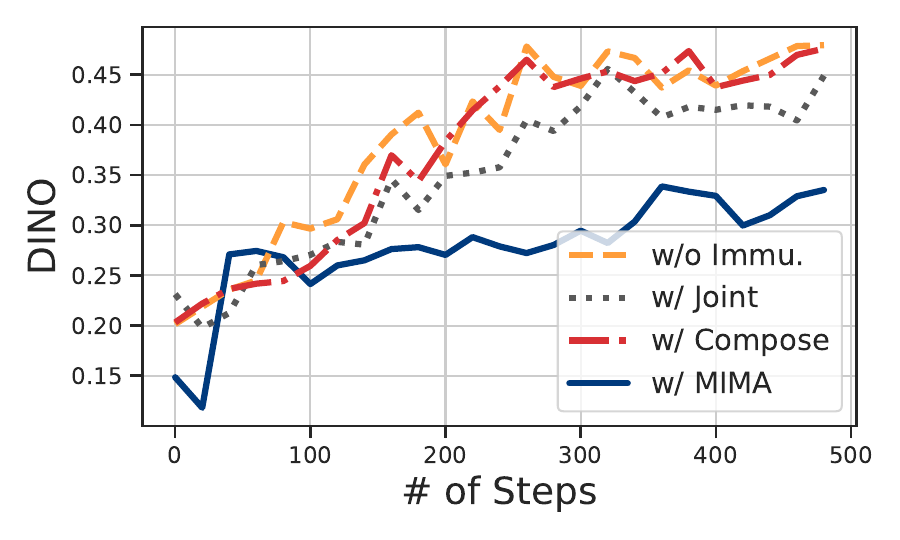} & \includegraphics[height=2.0cm, trim={10 0.4cm 10 0.4cm},clip]{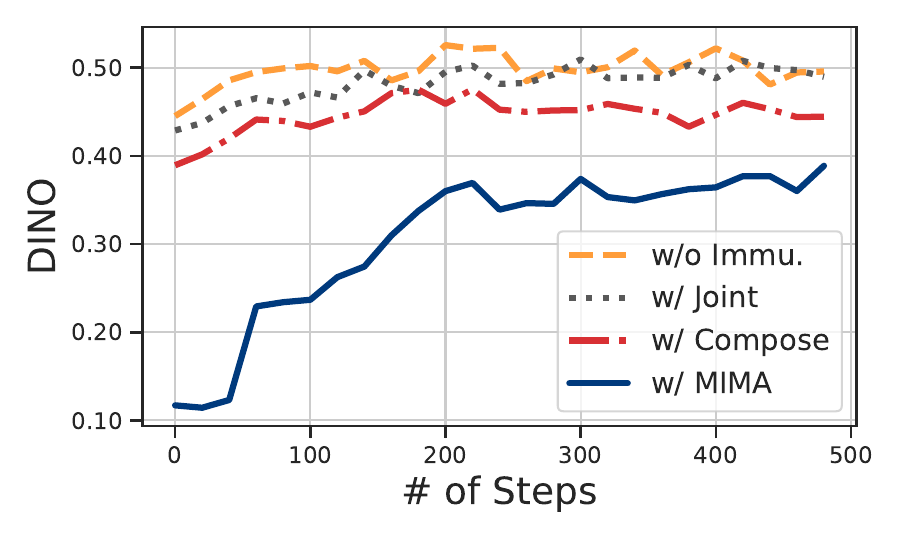} & \includegraphics[height=2.0cm, trim={10 0.4cm 10 0.4cm},clip]{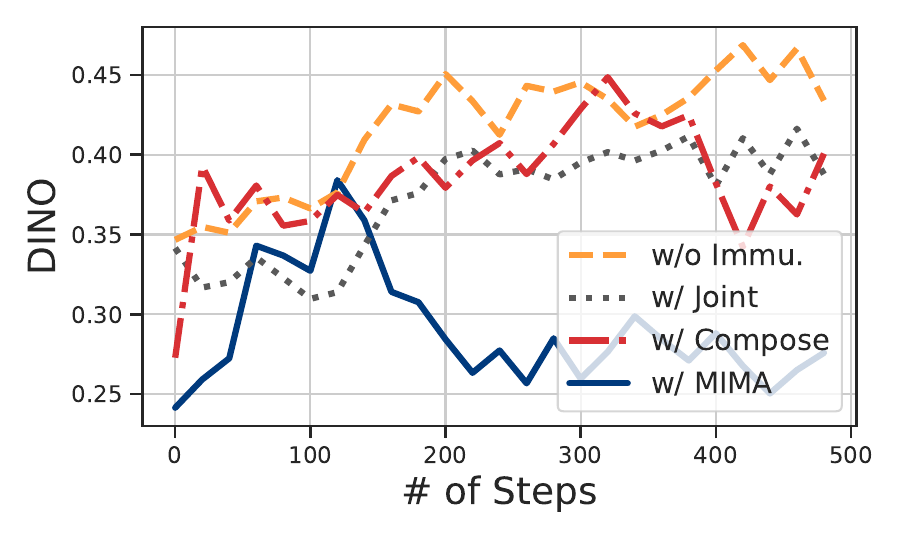} & \includegraphics[height=2.0cm, trim={10 0.4cm 10 0.4cm},clip]{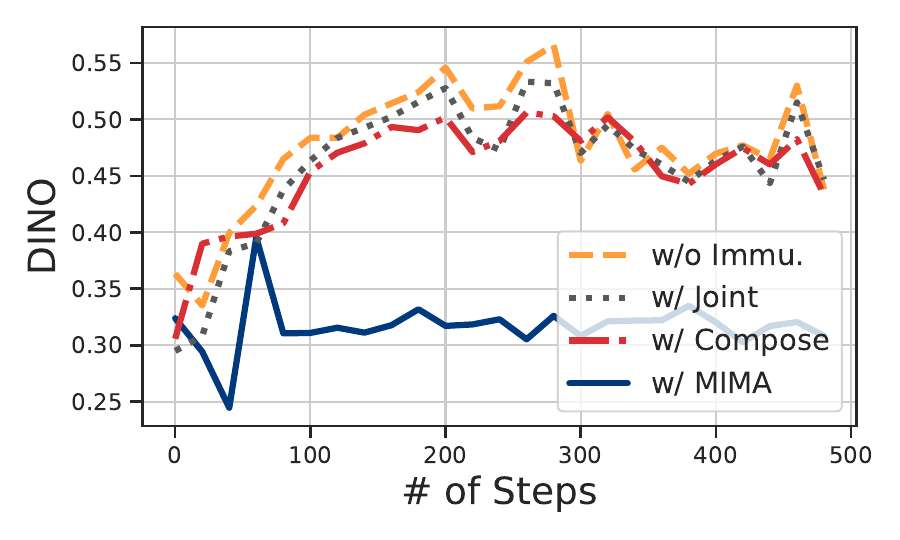} & 
     \includegraphics[height=2.0cm, trim={10 0.4cm 10 0.4cm},clip]{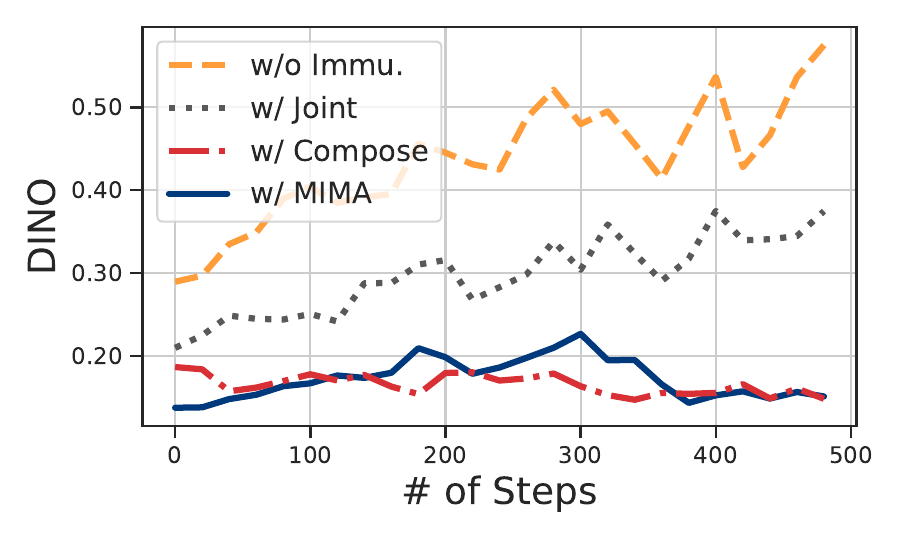} \\
     \includegraphics[height=2.0cm, trim={10 0.4cm 10 0.4cm},clip]{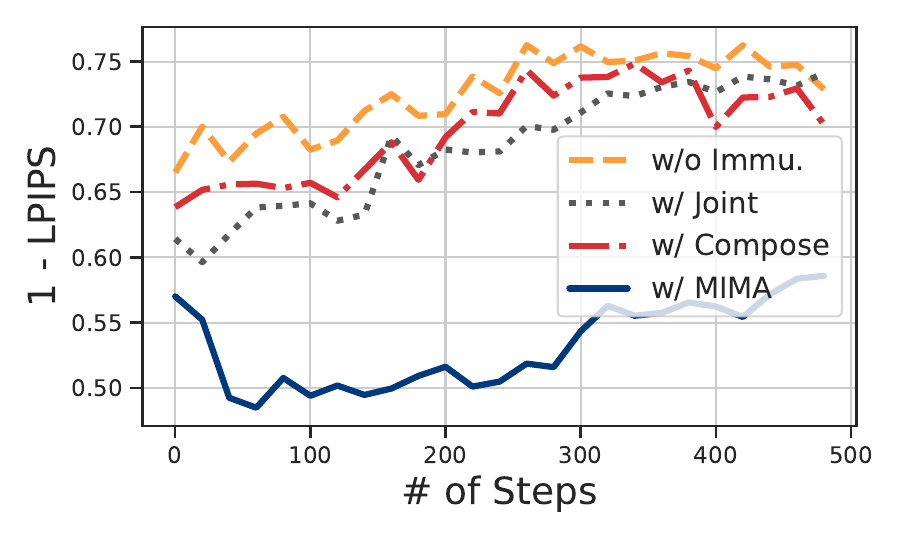} & \includegraphics[height=2.0cm, trim={10 0.4cm 10 0.4cm},clip]{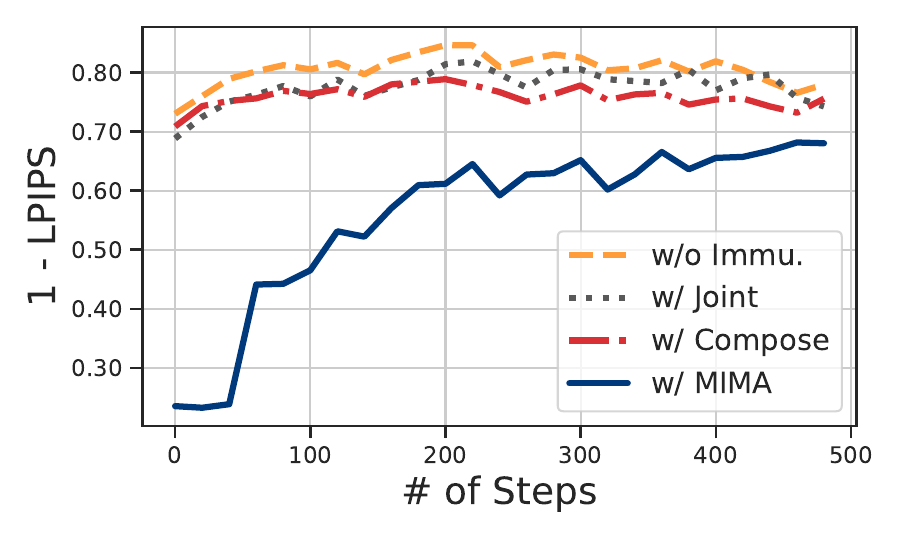} & \includegraphics[height=2.0cm, trim={10 0.4cm 10 0.4cm},clip]{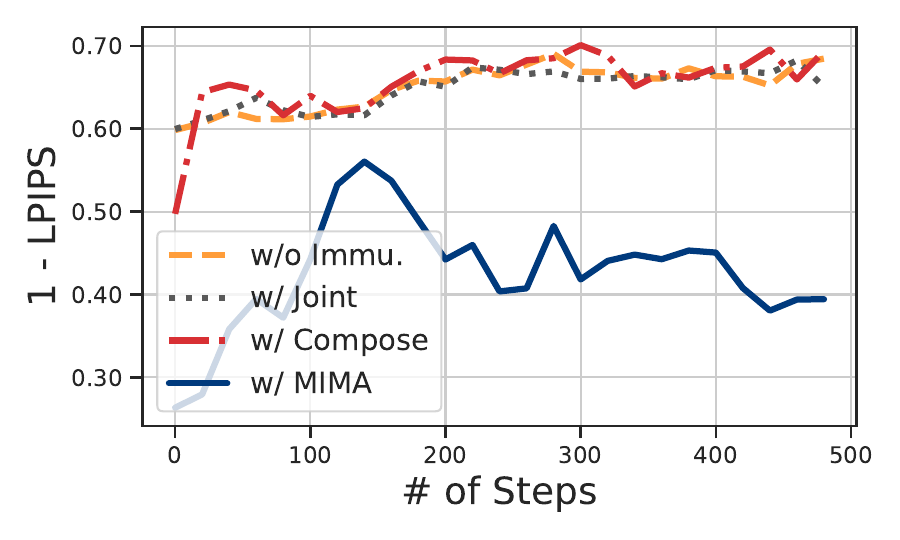} & \includegraphics[height=2.0cm, trim={10 0.4cm 10 0.4cm},clip]{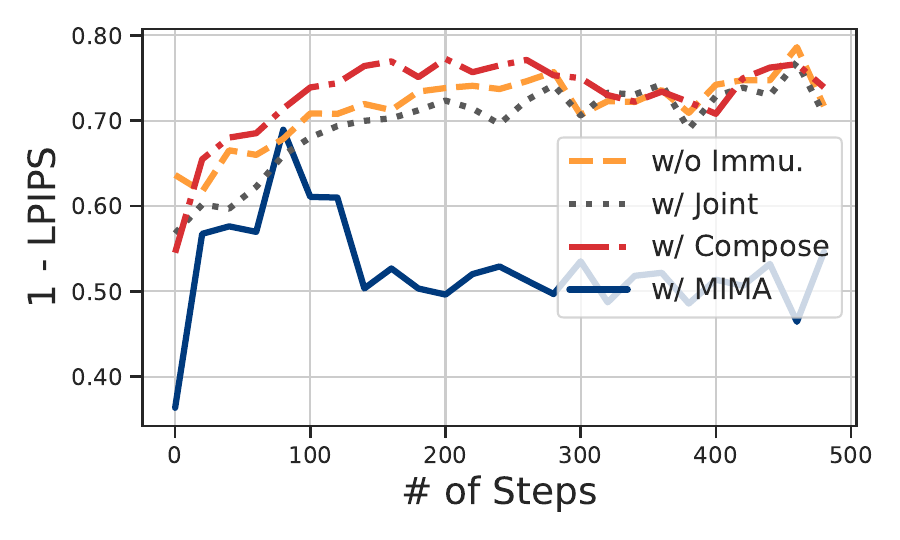} & 
     \includegraphics[height=2.0cm, trim={10 0.4cm 10 0.4cm},clip]{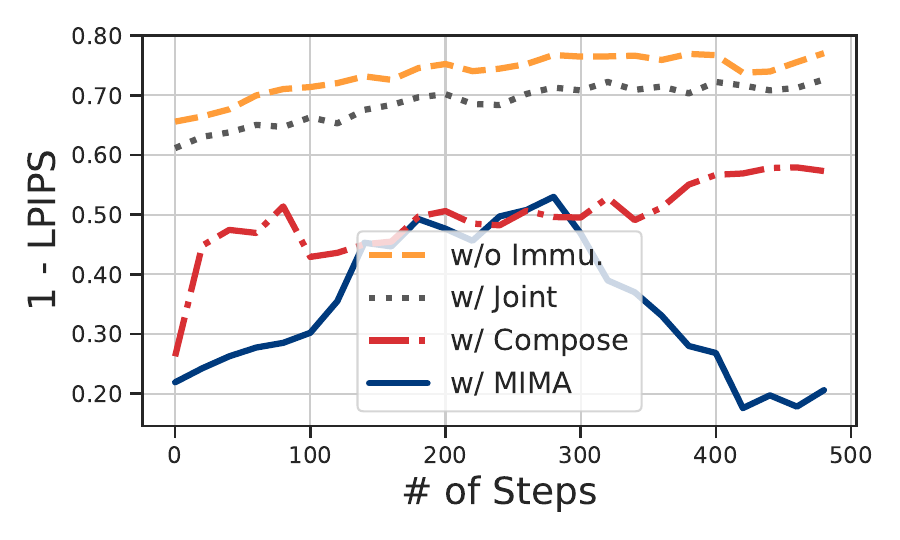}
    \end{tabular}
        \vspace{-0.2cm}
    \caption{{Similarity \vs epochs for LoRA on styles.} Each row shows one metric. Models with MIMA achieve lower similarity throughout LoRA's steps. This means that on the target concepts, MIMA generates images less similar to the references.
    }
    \vspace{-0.25cm}
    \label{fig:sgr_erase_style}
\end{figure*}

\myparagraph{Baselines.} 
In our experiments, we compare MIMA against two baselines extended from IMMA:
\begin{itemize}[topsep=0pt, ]
    \item Joint (\bsla) performs multi-concept immunization by joint training all concepts by combining the training datasets into one. For the $\gA$, we choose DreamBooth as the inner loop adaptation algorithm, \ie, modifying the whole U-Net, which gives the best immunization performance among the different adaptation methods used in IMMA.
    \item Compose (\bslb) only aggregates the cross-attention key and value weights via~\equref{eq:optim_layer} and freezes all the other weights of $\theta$ during immunization training. This is equivalent to IMMA choosing Custom Diffusion~\cite{kumari2022customdiffusion} %
    as the adaptation algorithm $\gA$ during immunization. As Custom Diffusion supports customization on multiple concepts, \bslb is a natural generalization of IMMA to multi-concept immunization.
\end{itemize}

\subsection{Multi-concept Re-learning Immunization}

Following IMMA~\cite{zheng2023imma}, we perform experiments on eight artistic styles and ten classes spanning various categories from a subset of ImageNet~\cite{deng2009imagenet}. 

\myparagraph{Experiment details.} We choose the concept sets by randomly sampling two or three concepts from the eight styles or the ten classes. The pre-trained weights are from 
UCE~\cite{gandikota2024unified}, an algorithm that erases multiple concepts from a pre-trained Diffusion model. 
For immunization, we generate 20 images for each target concept from Stable Diffusion V1-4 (SD V1-4) with the prompts of the target artistic styles and objects. Specpficially, the prompts are \textit{``an artwork of \{artist name\}''} and \textit{``a photo of \{object name\}''} rescptively.

As in IMMA, we consider the risk of re-learning the concept using the efficient adaptation method of LoRA~\cite{hu2022lora}.
We generate another 20 images to be used as the training images for LoRA. 
To maintain the model capability of being finetuned to learn other concepts, we generate 200 regularization images for \ml using either the prompt \textit{``artwork''} or \textit{``object''} for each of the corresponding settings. The results for re-learning style are presented in this section, and the results for objects can be found in the appendix.
\begin{table}[t]
\footnotesize
\setlength{\tabcolsep}{2pt}
\centering
\resizebox{1.\linewidth}{!}{
\begin{tabular}{llccccc@{\hskip 10pt}ccccc}
\specialrule{.15em}{.05em}{.05em}
           \multicolumn{2}{c}{\multirow{2}{*}{Group \#}} & \multicolumn{5}{c}{2-concept} & \multicolumn{5}{c}{3-concept} \\ \cline{3-7} \cline{8-12}
           && 1  & 2 & 3 & 4  & 5 & 1  & 2 & 3 & 4 & 5  \\
\hline
\multirow{3}{*}{\bsla} 
&\texttt{(C)}     & 1.26     & 1.80          & 0.81        & 3.13           & 2.16       & 0.55        & 3.45           & -1.32    & 3.11  & 5.38 \\
 &\texttt{(D)}     & 9.87     & 10.6         & 6.00       & 21.0          & 1.88      & 6.34        & 29.2          & 0.61     & 15.8 & 12.3 \\
 &\texttt{(L)}     & 2.82     & 0.31         & 4.23       & 4.65          & 2.80      & 1.51           & 5.78           & -1.17    & 2.18  & 5.90\\
 \hline
\multirow{3}{*}{\bslb} 
&\texttt{(C)}     & 5.31     & 1.15          & 0.82         & \bf7.38    &\bf 7.45       & 6.13       & 7.44           & 6.43      & 3.28 & 7.10  \\
 &\texttt{(D)}     & 24.8     & -5.30         & 7.40       & 31.1          & 19.5      & 19.1      & 26.1          & 31.3     & 29.5 &\bf 26.3 \\
  &\texttt{(L)}     & 24.9     & 3.02         & 6.96       & 8.32          & 8.37      & 13.3       & 17.4          & 32.3    & 9.73 & 16.0 \\
 \hline
\multirow{3}{*}{Ours} 
&\texttt{(C)}     &\bf 6.66     &\bf 12.2    &\bf 6.84        & 3.89           & 7.26       &\bf 6.92       &\bf 13.2          &\bf 11.5     &\bf 14.8 & 7.67 \\
 &\texttt{(D)}     &\bf 25.3     &\bf 22.0  & \bf 27.6       & \bf 42.4     &\bf 20.0      &\bf 50.4     &\bf 46.9          &\bf 39.4     &\bf 46.6 &\bf 19.5 \\
 &\texttt{(L)}     &\bf 41.4     &\bf 12.3  &\bf 22.2       &\bf 11.7    &\bf 28.7      &\bf 38.9     &\bf 44.8          &\bf 38.1     &\bf 42.6 &\bf 19.6  \\
\specialrule{.15em}{.05em}{.05em}
\end{tabular}
}
    \vspace{-0.15cm}
\caption{\msgr$\uparrow$(\%) on artistic styles for UCE with LoRA. MIMA shows an average \msgr improvement of 18.95\% over \bsla and 10.94\% over \bslb across all three similarity metrics.
}
\vspace{-0.15cm}
\label{tab:erase_style}
\end{table}

\myparagraph{Evaluation metrics.}
IMMA explained that the effectiveness of an immunization can be evaluated by quantifying the performance \textit{gap} with and without the immunization. Following this intuition, we propose \textit{Mean Similarity Gap Ratio} (\msgr) between the generation with and without MIMA for \textit{all target concepts} from $\gC$ as an evaluation metric. 

Given a metric $\gM$ that captures image similarity, $\msgr\left(\{\rvx^\gI_{[n]}\}, \{\rvx^\gA_{[n]}\}, \{\rvx^r_{[n]}\}\right)$ is defined as
\be
        \label{eq:sgr}
    \frac{1}{|\gC|} \sum_{n=1}^{|\gC|}\frac{
    \overbrace{\gM(\rvx^r_{[n]}, \rvx^\gA_{[n]})}^{\text{w/o immunization} (\uparrow)} - \overbrace{\gM(\rvx^r_{[n]}, \rvx^\gI_{[n]})}^{\text{w/ immunization}(\downarrow)}}{\gM(\rvx^r_{[n]}, \rvx^\gA_{[n]})}.
\ee
Here, $\rvx^\gI_{[n]}$ and $\rvx^\gA_{[n]}$ denote the generated images with and without immunization of the $n$-th target concept, and $\rvx^r_{[n]}$ denotes the corresponding reference images of the target concept. A larger \msgr indicates a stronger effect of MIMA as the performance gap is larger.

\begin{figure}[t]
    \hspace{-.5cm}
    \centering
    \footnotesize
    \setlength{\tabcolsep}{1.2pt}
    \renewcommand{\arraystretch}{1.2}
    \begin{tabular}{lcc@{\hskip 5pt}cc}
    & \multicolumn{2}{c}{Stable Diffusion} & \multicolumn{2}{c}{Re-learning}\vspace{-3pt}  \\
    & Reference  & Erased & w/o Immu. &  w/ MIMA \\
    \multirow{2}{*}[1.cm]{\rotatebox[origin=c]{90}{Monet}} & 
    \includegraphics[height=1.8cm,width=1.8cm]{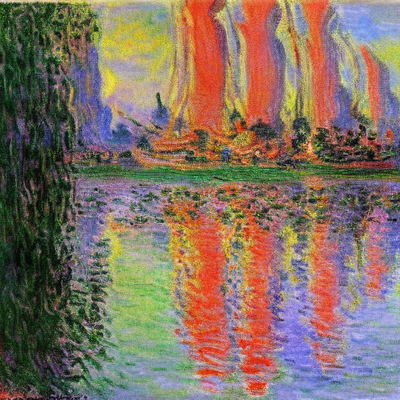} & 
    \includegraphics[height=1.8cm,width=1.8cm]{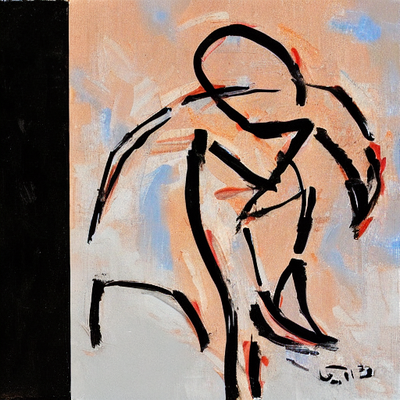}  &
    \includegraphics[height=1.8cm,width=1.8cm]{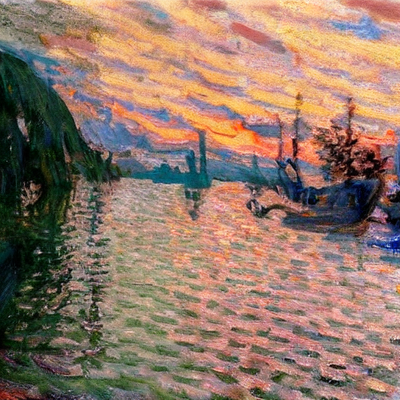}  &
    \includegraphics[height=1.8cm,width=1.8cm]{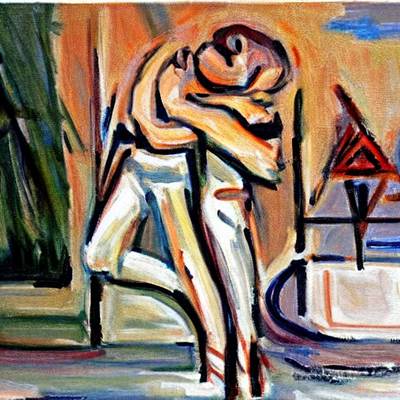}  \\
    \multirow{2}{*}[1.3cm]{\rotatebox[origin=c]{90}{Kilian Eng}} & 
    \includegraphics[height=1.8cm,width=1.8cm]{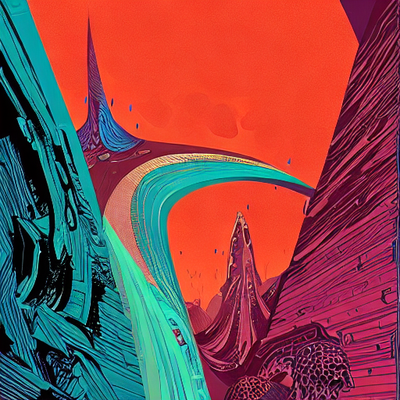} & 
    \includegraphics[height=1.8cm,width=1.8cm]{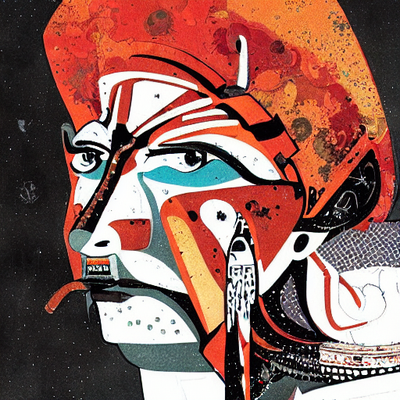}  &
    \includegraphics[height=1.8cm,width=1.8cm]{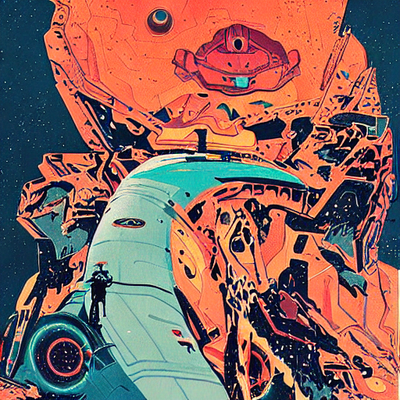}  &
    \includegraphics[height=1.8cm,width=1.8cm]{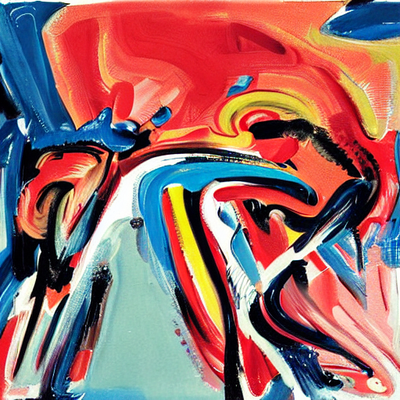}  \\
    \multirow{2}{*}[1.4cm]{\rotatebox[origin=c]{90}{Tyler Edlin }} & 
    \includegraphics[height=1.8cm,width=1.8cm]{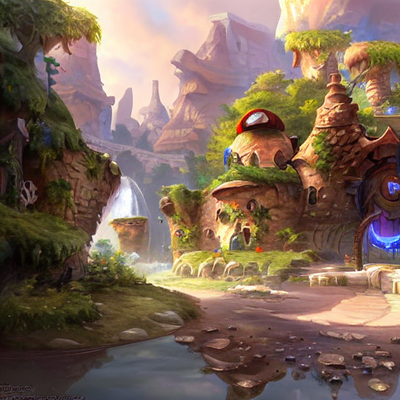} & 
    \includegraphics[height=1.8cm,width=1.8cm]{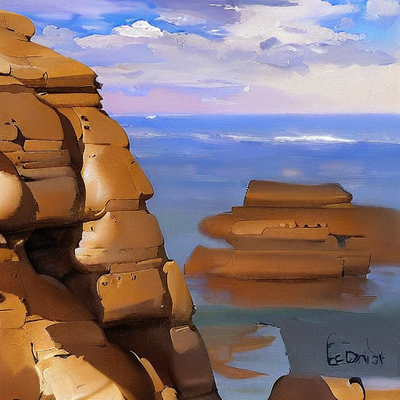}  &
    \includegraphics[height=1.8cm,width=1.8cm]{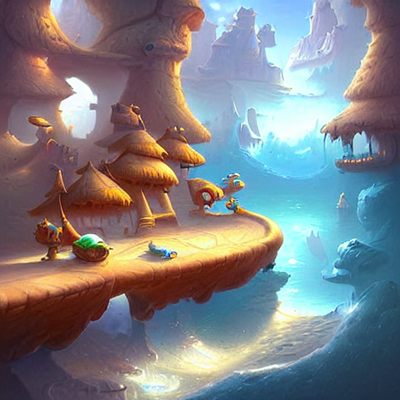}  &
    \includegraphics[height=1.8cm,width=1.8cm]{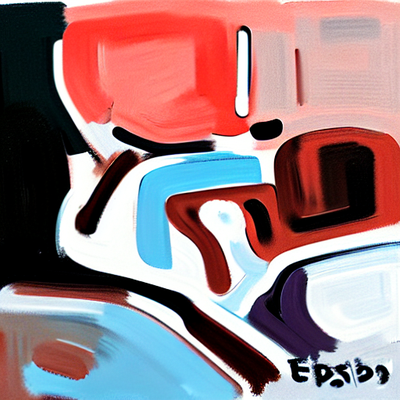} \\
    \end{tabular}
        \vspace{-0.15cm}
    \caption{Qualitative result of MIMA against re-learning artistic styles. Both Erased and MIMA are adapted to all three concepts on a single model.
    }
    \vspace{-0.25cm}
    \label{fig:qual_erase_style}
\end{figure}

Following IMMA, we choose $\gM$ to be one minus the Learned Perceptual Image Patch Similarity~\cite{zhang2018unreasonable} (LPIPS), 
cosine similarity measured in the feature space of CLIP~\cite{Radford_CLIP} or DINO~\cite{caron2021emerging} each denoted as \texttt{MSGR(L)}, \texttt{MSGR(C)} and \texttt{MSGR(D)}.

\myparagraph{Style results.}
In~\tabref{tab:erase_style}, we report the \msgr of re-learning sets of concepts after they were erased by UCE~\cite{gandikota2024unified}. We provide results on five groups of concepts for each of the two or three concept combinations, \eg, group 1 of two concepts corresponds to the artistic style of Monet, and Fagan. The group details are provided in the Appendix. All the numbers are reported at the $400^{\tt th}$ step of LoRA with a batch size of four. 

We observe that the \msgr of MIMA is generally greater than zero. A positive gap between the similarity without and with MIMA indicates the effectiveness of immunization. Overall, we observe that MIMA outperforms \bsla by 18.95\% and \bslb by 10.94\% averaging across all groups and metrics.

To further study the effectiveness of MIMA, we visualize the CLIP, DINO, and LPIPS metrics at each training step for LoRA in~\figref{fig:sgr_erase_style}. The gaps between the lines and the dashed orange line illustrate the \msgr. 
A larger gap means that the immunization method performs better. %
We observe that MIMA outperforms to two compared baselines.

In~\figref{fig:qual_erase_style}, we provide qualitative results and observe the following: 
(a) LoRA can train back the erased concepts on a model without immunization; (b) With the immunization of MIMA, a shows a degree of resistance to LoRA, \ie, the model fails to generate artwork in the style of the multiple protected artists. These observations are consistent with our quantitative findings.

\begin{table}[t]
\small %
\centering
\setlength{\tabcolsep}{2pt}
\resizebox{\columnwidth}{!}{
\begin{tabular}{llccccc@{\hskip 10pt}ccccc}
\specialrule{.15em}{.05em}{.05em}
           \multicolumn{2}{c}{\multirow{2}{*}{ Group \#}} & \multicolumn{5}{c}{2-concept} & \multicolumn{5}{c}{3-concept} \\ \cline{3-7} \cline{8-12}
           && 1  & 2 & 3 & 4  & 5 & 1  & 2 & 3 & 4 & 5  \\
\hline
\multirow{2}{*}{\bsla} 
&\texttt{(C)}     & 3.34     & 3.77          & 3.30        & 2.08           &  5.38      & 2.77               & 3.72        & 2.23     & 6.50 & 3.97 \\
 &\texttt{(D)}     & 11.3     & 16.5         & 14.3       & 7.94         &  17.1     & 14.1           & 10.3          & 14.1     & 20.2 & 13.6  \\
 \hline
\multirow{2}{*}{\bslb} 
&\texttt{(C)}     & 1.26     & 0.21          & 2.34        & 1.88          & 5.04       & 2.21               & 1.25        & 1.58   & 2.07 & 0.77 \\
 &\texttt{(D)}     & 3.08     & 5.11         & 9.28       & 10.7          &  22.7    & 12.0          & 14.6          &\bf 21.6     & 5.45 & 1.22 \\
 \hline
\multirow{2}{*}{Ours} 
&\texttt{(C)}     &\bf 6.57     &\bf 5.49          &\bf 6.23        &\bf 5.57           &\bf  7.48      &\bf 4.97           &\bf 3.75       &\bf 3.35     & \bf6.64 &\bf 6.43  \\
 &\texttt{(D)}     &\bf 14.7     &\bf 18.1         &\bf 17.5       &\bf 28.5        &\bf  24.8    &\bf 17.1           &\bf 19.4          & 13.6     &\bf 22.7 &\bf 18.7  \\

\specialrule{.15em}{.05em}{.05em}
\end{tabular}
}
    \vspace{-0.15cm}
\caption{\msgr$\uparrow$(\%) on personalized adaptation. MIMA shows an average \msgr improvement of 3.75\% over \bsla and 6.36\% over \bslb across all groups.
}
\vspace{-0.15cm}
\label{tab:sgr_per}
\end{table}

\begin{figure*}[t]
    \centering
    \setlength{\tabcolsep}{0.5pt}
    \begin{tabular}{cc @{\hskip 3pt}|@{\hskip 3pt}ccc}
     \multicolumn{2}{c}{\bf 2-concept} & \multicolumn{3}{c}{\bf 3-concept} \\
     plant & castle & wooden pot & guitar & plant \\
     \includegraphics[height=2.0cm, trim={10 0.4cm 10 0.4cm},clip]{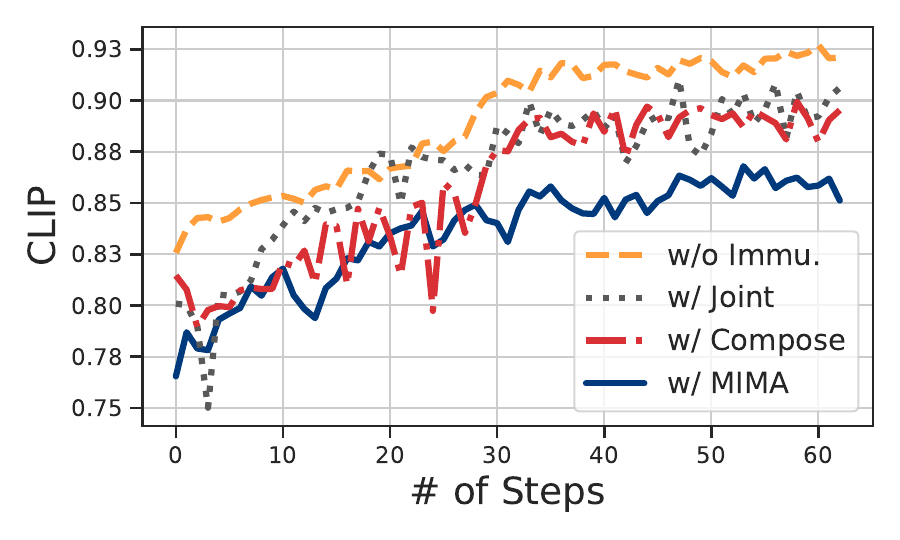} & 
     \includegraphics[height=2.0cm, trim={10 0.4cm 10 0.4cm},clip]{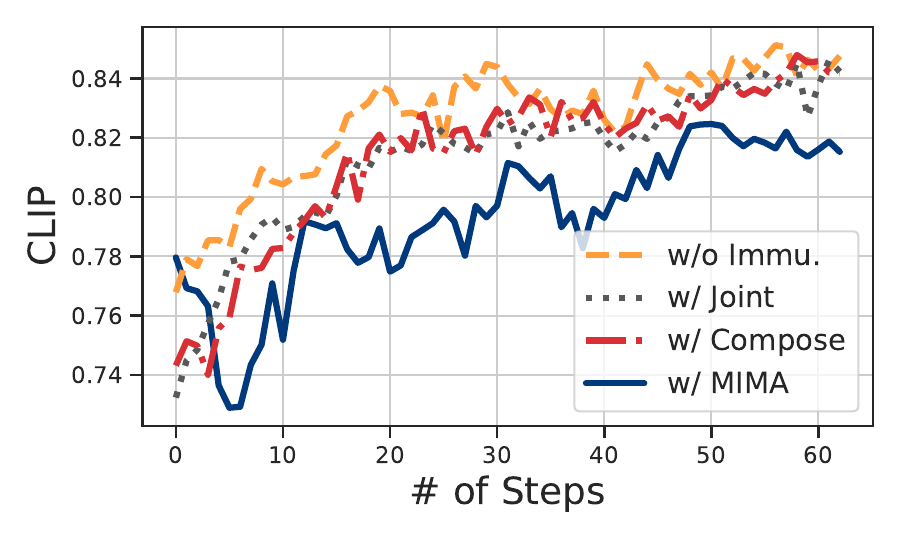} &
     \includegraphics[height=2.0cm, trim={10 0.4cm 10 0.4cm},clip]{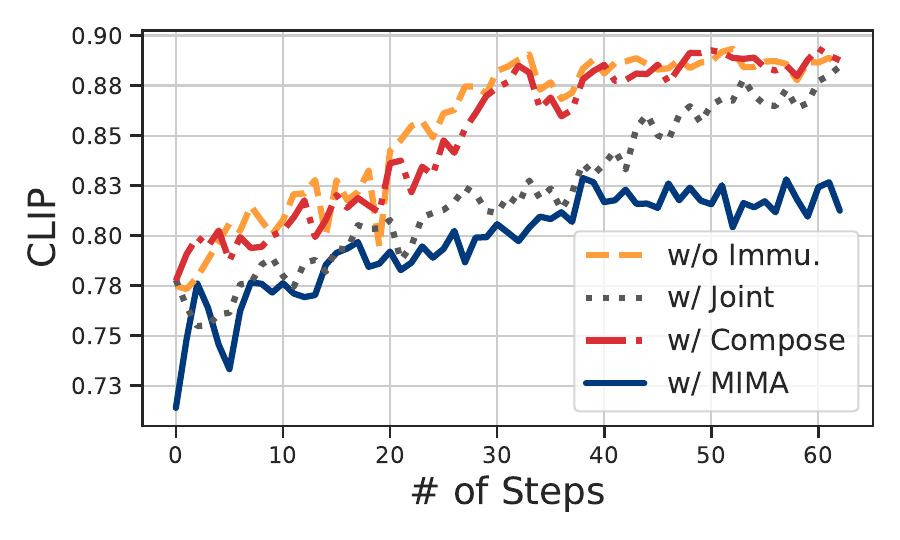} & 
     \includegraphics[height=2.0cm, trim={10 0.4cm 10 0.4cm},clip]{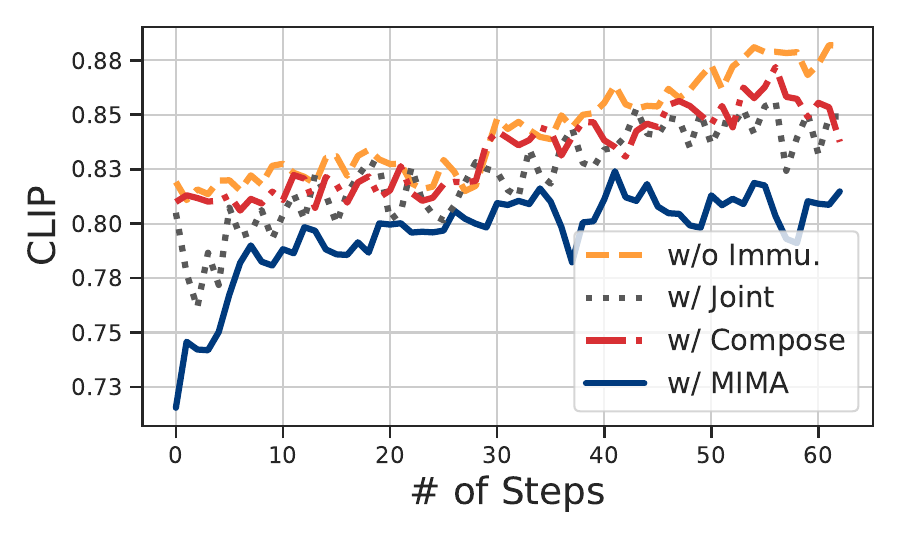} & 
     \includegraphics[height=2.0cm, trim={10 0.4cm 10 0.4cm},clip]{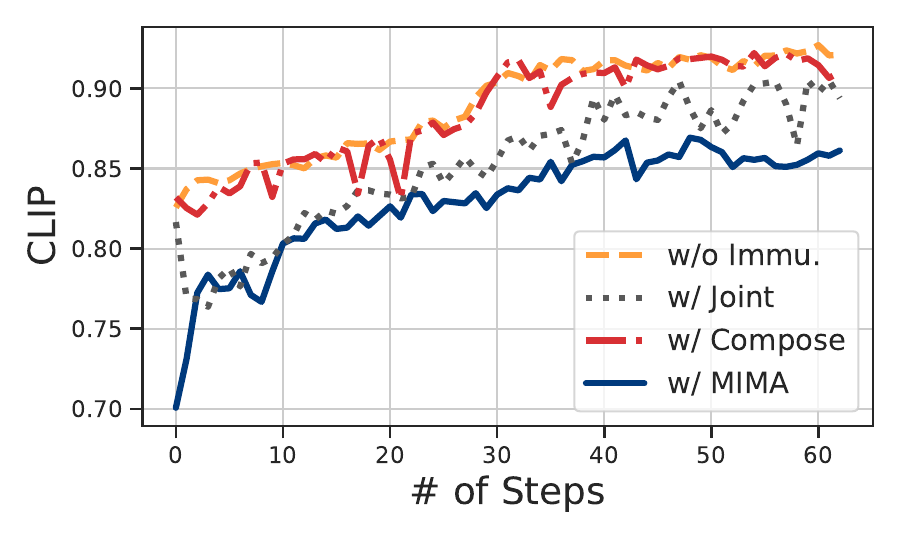} \\
     \includegraphics[height=2.0cm, trim={10 0.4cm 10 0.4cm},clip]{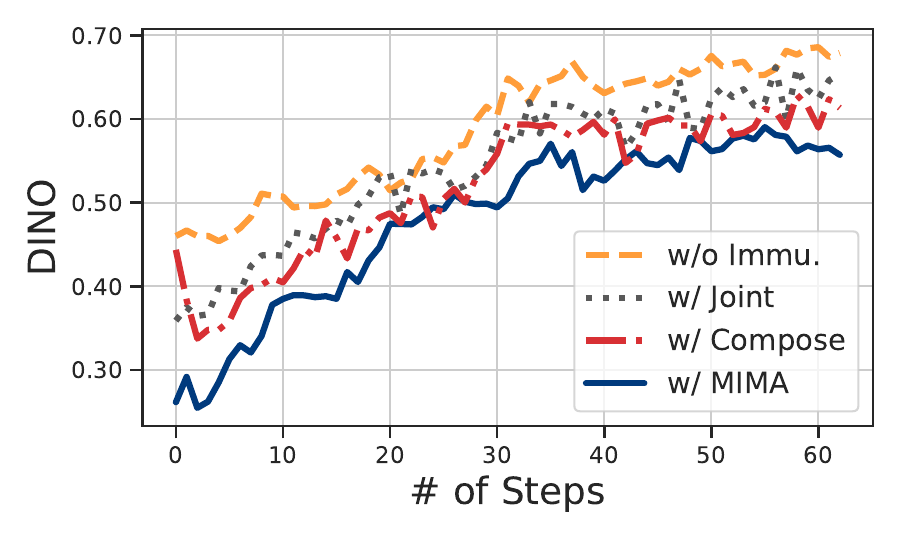} & 
     \includegraphics[height=2.0cm, trim={10 0.4cm 10 0.4cm},clip]{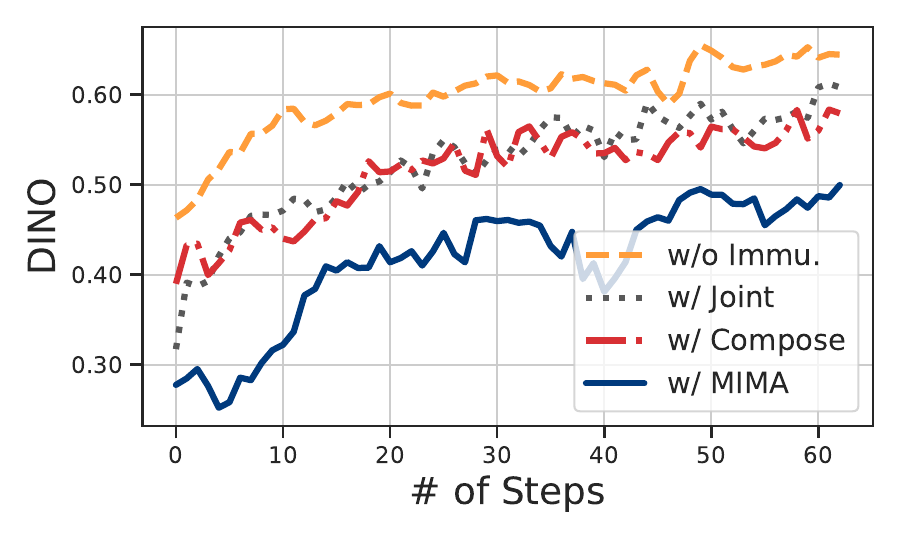} & 
     \includegraphics[height=2.0cm, trim={10 0.4cm 10 0.4cm},clip]{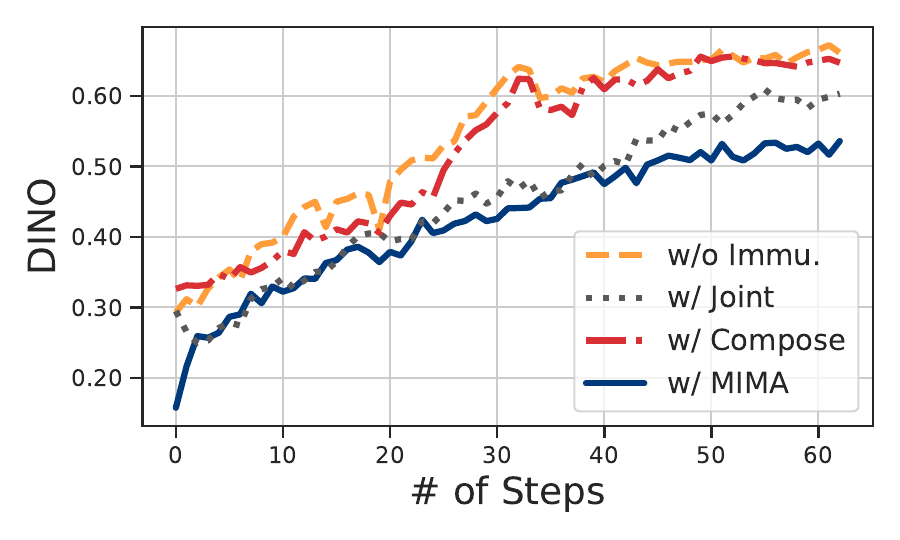} & 
     \includegraphics[height=2.0cm, trim={10 0.4cm 10 0.4cm},clip]{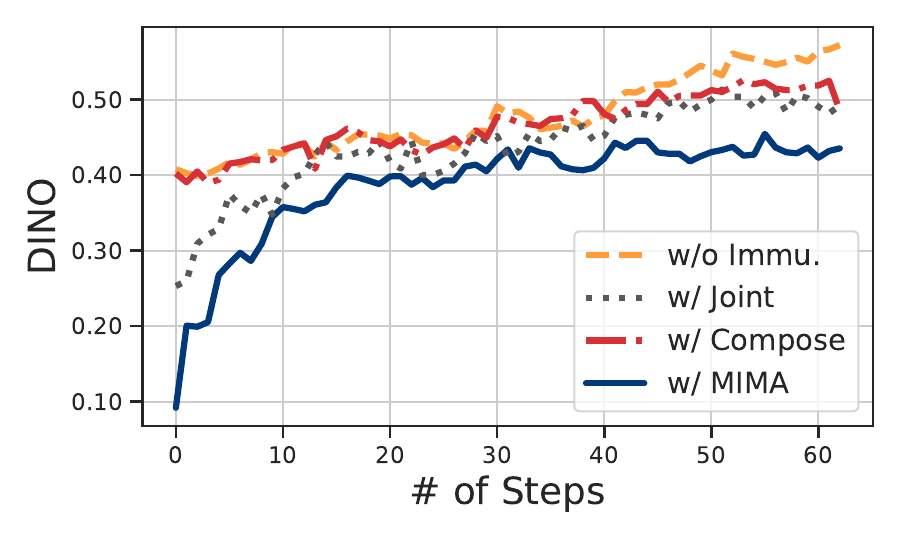} & 
     \includegraphics[height=2.0cm, trim={10 0.4cm 10 0.4cm},clip]{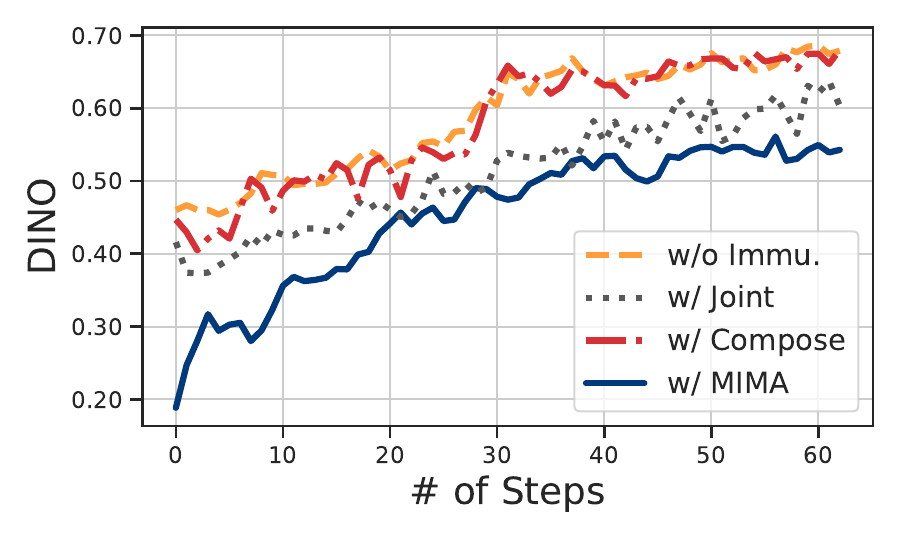} \\
    \end{tabular}
        \vspace{-0.225cm}
    \caption{{CLIP and DINO similarity on personalization concepts. } The \textit{gaps} between the dashed line and solid lines show \msgr$\uparrow$(\%) of different methods. That is, a larger gap indicates stronger immunization.
    }
    \vspace{-0.3cm}
    \label{fig:sgr_per}
\end{figure*}

\begin{table}[t]
\small %
\centering
\setlength{\tabcolsep}{2pt}
\resizebox{\columnwidth}{!}{
\begin{tabular}{llccccc@{\hskip 10pt}ccccc}
\specialrule{.15em}{.05em}{.05em}
           \multicolumn{2}{c}{\multirow{2}{*}{Group \#}} & \multicolumn{5}{c}{2-concept} & \multicolumn{5}{c}{3-concept} \\ \cline{3-7} \cline{8-12}
           && 1  & 2 & 3 & 4  & 5 & 1  & 2 & 3 & 4 & 5  \\
\hline
\multirow{2}{*}{\bsla} 
&\texttt{(C)}     & 1.82     & 4.26          & 1.28        & 0.08           &  4.96      & 2.52          & 3.14           & 2.23       & 3.84 & 1.73   \\
 &\texttt{(D)}     & 14.2     & 16.6         & 7.02       & 0.01          &  8.43     & 7.33           & -10.9          & 8.73     & 4.81 & 6.48  \\
 \hline
\multirow{2}{*}{\bslb} 
&\texttt{(C)}     & 1.10     & 3.52          & -0.03        & -0.67        &  4.05      & 0.49          & -0.58           & 1.58       & 1.83 & -0.55  \\
 &\texttt{(D)}     & 7.46     & 9.83         & 6.88       & -7.95          &  2.18    & -5.93           & -5.76          & -0.30     & 2.36 & 0.77 \\
 \hline
\multirow{2}{*}{Ours} 
&\texttt{(C)}     &\bf 5.08     &\bf 5.60          &\bf 3.16        &\bf 4.38           &\bf  5.11      &\bf 5.79          &\bf 7.10           &\bf 4.36       &\bf 4.03 &\bf 1.92  \\
 &\texttt{(D)}     &\bf 22.4     &\bf 22.9         &\bf 15.0       &\bf 15.7        &\bf  9.36     &\bf 20.7          &\bf 14.8          &\bf 15.7     &\bf 16.2 &\bf 7.20 \\
\specialrule{.15em}{.05em}{.05em}
\end{tabular}
}
    \vspace{-0.15cm}
\caption{\mrsgr$\uparrow$(\%) on personalized adaptation. MIMA shows an average \mrsgr improvement of 5.89\% over \bsla and 9.31\% over \bslb across all groups.}
\vspace{-0.15cm}
\label{tab:rsgr_per}
\end{table}

\begin{figure}[t]
    \centering
    \setlength{\tabcolsep}{0.5pt}
    \begin{tabular}{cc@{\hskip 6pt}cc}
     \multicolumn{2}{c}{plant+castle} \\
     \includegraphics[height=2.2cm, trim={10 0.4cm 10 0.4cm},clip]{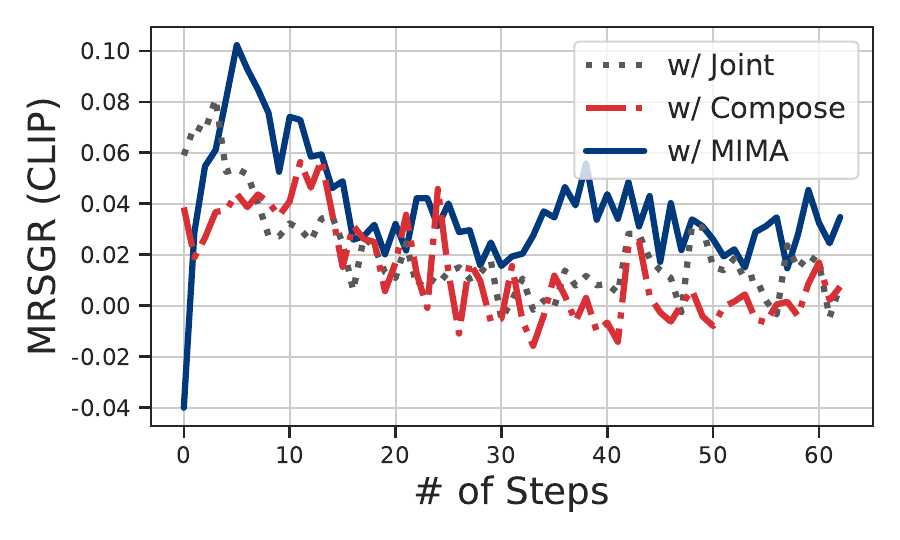} & \includegraphics[height=2.2cm, trim={10 0.4cm 10 0.4cm},clip]{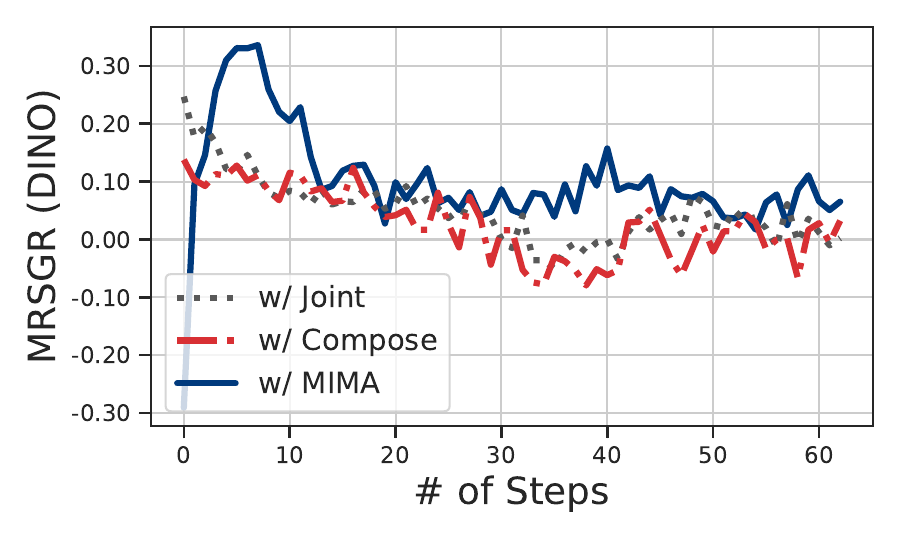}
    \end{tabular}
    \begin{tabular}{cc@{\hskip 6pt}cc}
     \multicolumn{2}{c}{woodenpot+guitar+plant} \\
     \includegraphics[height=2.2cm, trim={10 0.4cm 10 0.4cm},clip]{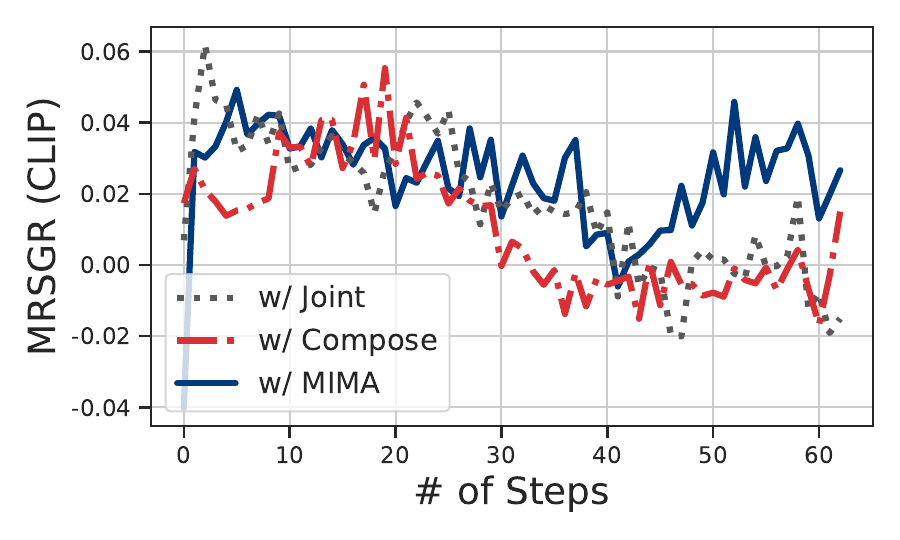} & \includegraphics[height=2.2cm, trim={10 0.4cm 10 0.4cm},clip]{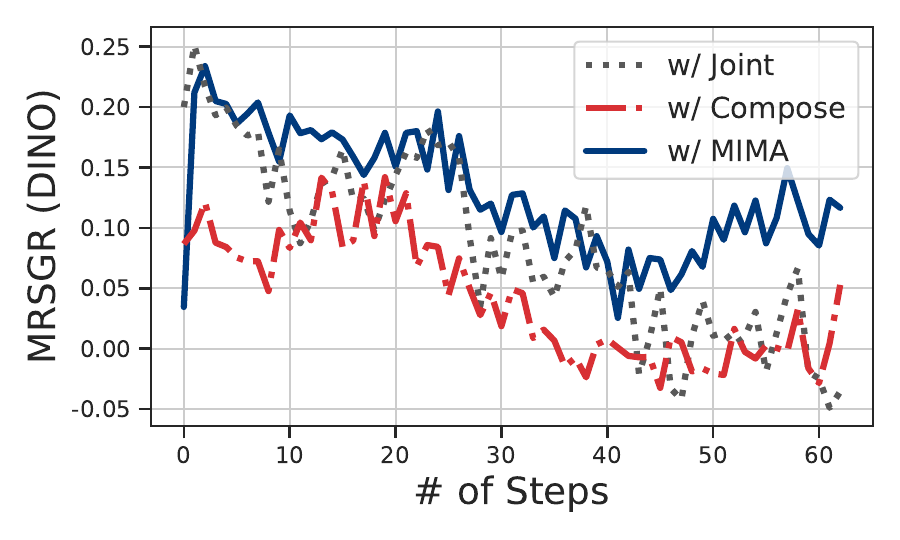} \\
    \end{tabular}
        \vspace{-0.15cm}
    \caption{{\mrsgr$\uparrow$(\%) of MIMA and two baselines on personalization experiment.} 
    }
    \label{fig:rsgr_per}
    \vspace{-0.3cm}
\end{figure}

\begin{figure}[!htb]
    \centering
    \footnotesize
    \setlength{\tabcolsep}{1.3pt}
    \renewcommand{\arraystretch}{1.2}
    \begin{tabular}{c@{\hskip 6pt}ccccc}
      Reference & ~TI & DB & +LoRA & CD\\
    \includegraphics[height=1.51cm]{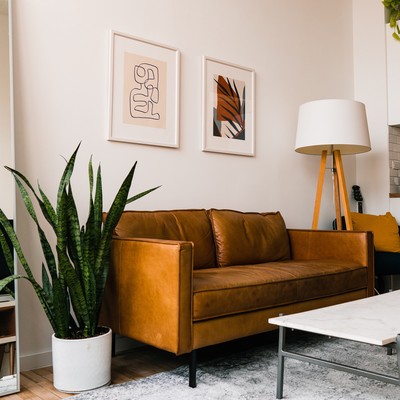} &
    \multirow{1}{*}[1.15cm]{\rotatebox[origin=c]{90}{w/o Immu.}}
    \includegraphics[height=1.51cm]{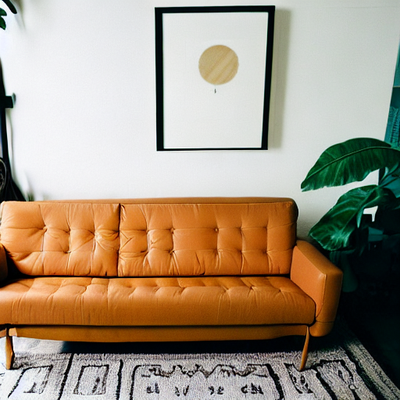} & 
    \includegraphics[height=1.51cm]{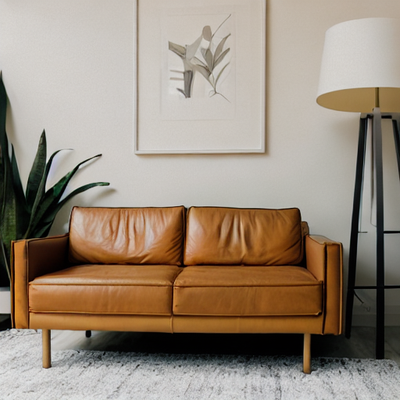} & 
    \includegraphics[height=1.51cm]{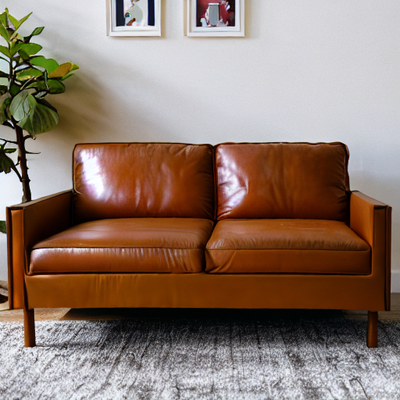} & 
    \includegraphics[height=1.51cm]{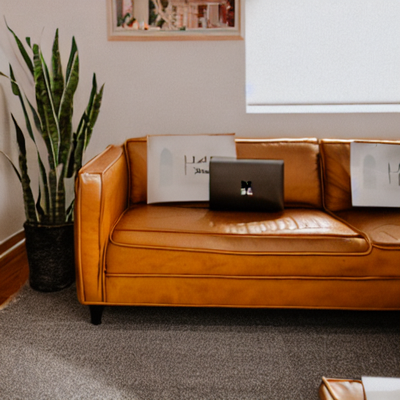} \\
    \includegraphics[height=1.51cm, width=1.51cm]{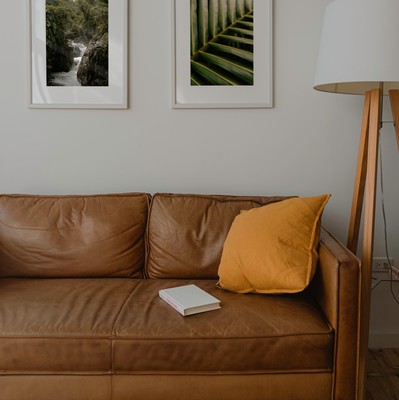} &
                       \vspace{0.15cm}
     \multirow{2}{*}[1.1cm]{\rotatebox[origin=c]{90}{ w/ MIMA}}
     \includegraphics[height=1.51cm]{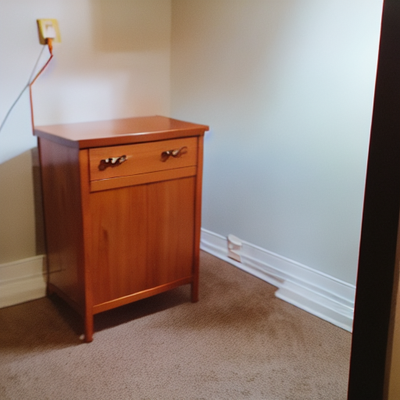} & 
     \includegraphics[height=1.51cm]{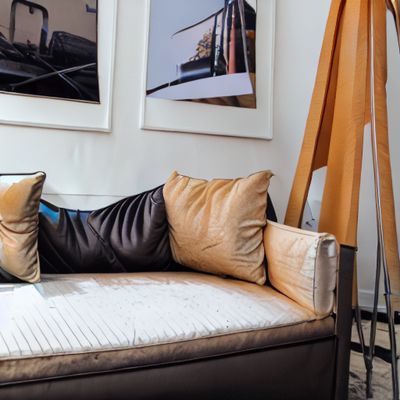} &
     \includegraphics[height=1.51cm]{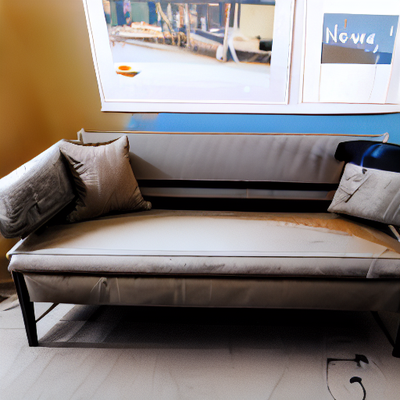} & 
     \includegraphics[height=1.51cm]{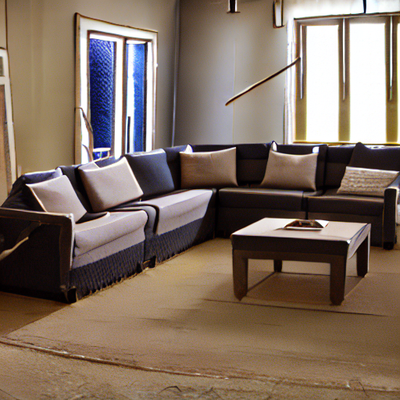} \\
     
    \includegraphics[height=1.51cm]{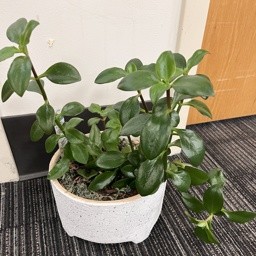} &
    \multirow{1}{*}[1.15cm]{\rotatebox[origin=c]{90}{w/o Immu.}}
    \includegraphics[height=1.51cm]{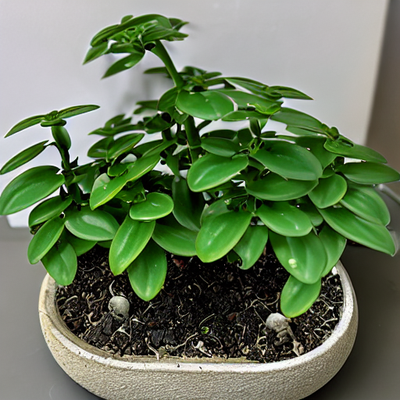} & 
    \includegraphics[height=1.51cm]{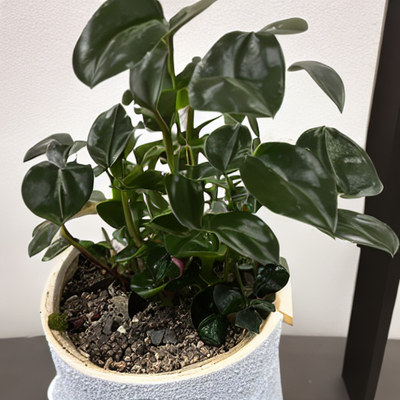} & 
    \includegraphics[height=1.51cm]{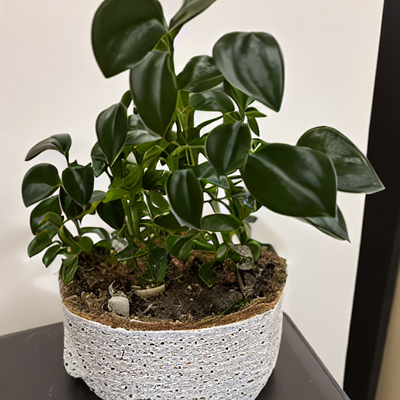} & 
    \includegraphics[height=1.51cm]{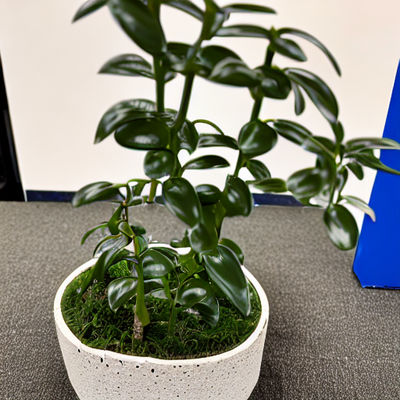} \\
        
    \includegraphics[height=1.51cm]{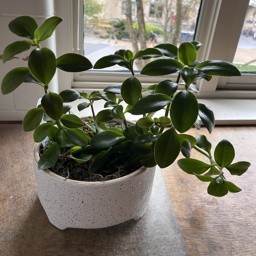} &
                       \vspace{0.15cm}
     \multirow{2}{*}[1.1cm]{\rotatebox[origin=c]{90}{w/ MIMA}}
     \includegraphics[height=1.51cm]{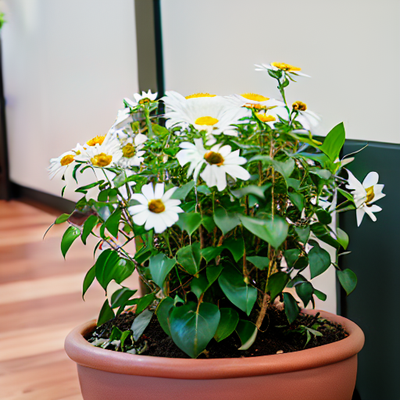} &
     \includegraphics[height=1.51cm]{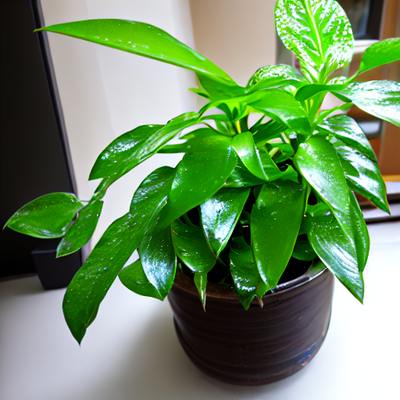} & 
     \includegraphics[height=1.51cm]{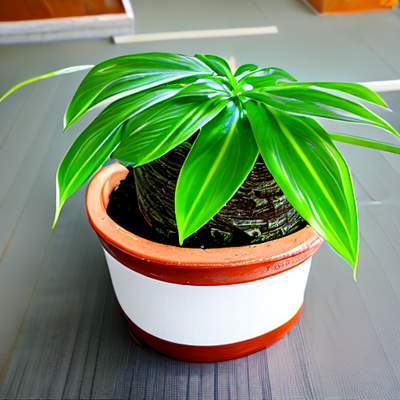} & 
     \includegraphics[height=1.51cm]{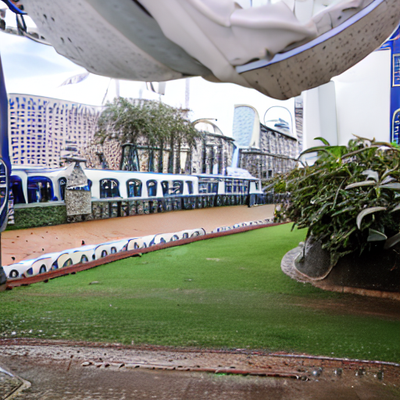} \\

     \includegraphics[height=1.51cm]{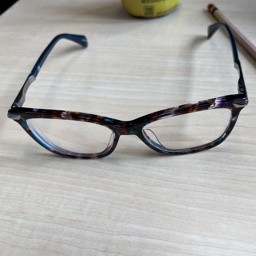} &
    \multirow{1}{*}[1.15cm]{\rotatebox[origin=c]{90}{w/o Immu.}}
    \includegraphics[height=1.51cm]{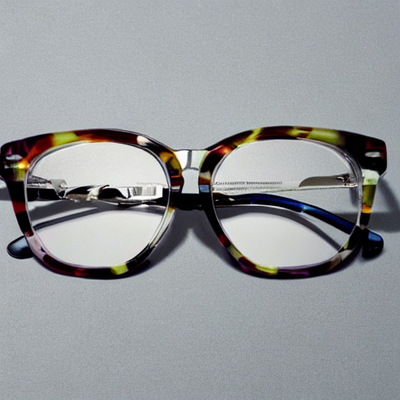} & 
    \includegraphics[height=1.51cm]{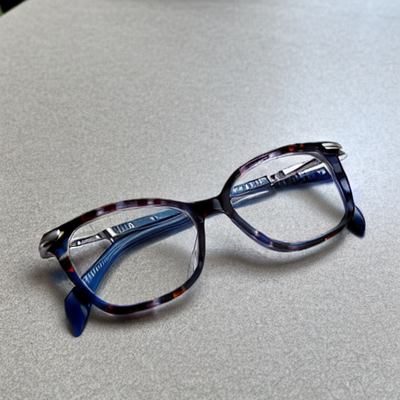} & 
    \includegraphics[height=1.51cm]{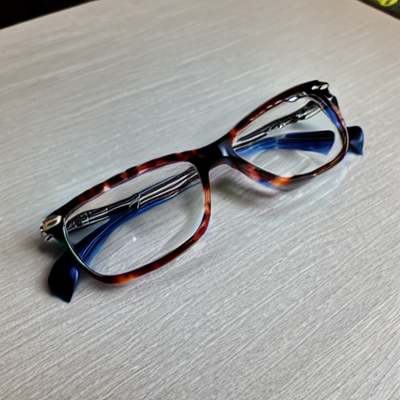} & 
    \includegraphics[height=1.51cm]{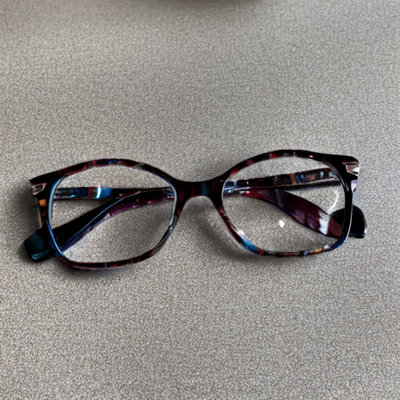} \\
    \includegraphics[height=1.51cm]{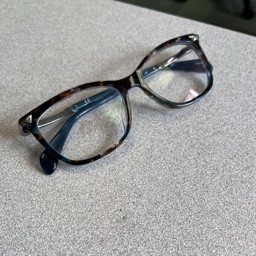} &
     \multirow{2}{*}[1.1cm]{\rotatebox[origin=c]{90}{w/ MIMA}}
     \includegraphics[height=1.51cm]{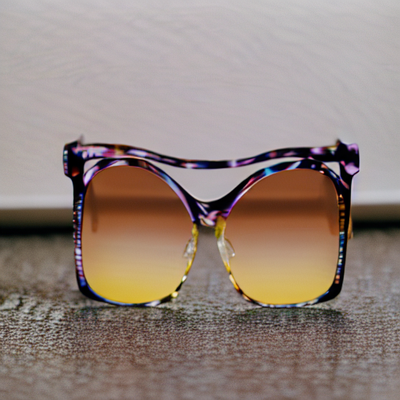} & 
     \includegraphics[height=1.51cm]{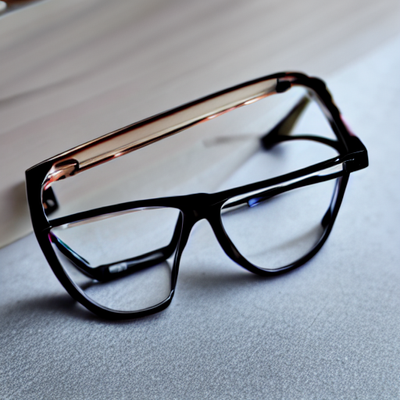} & 
     \includegraphics[height=1.51cm]{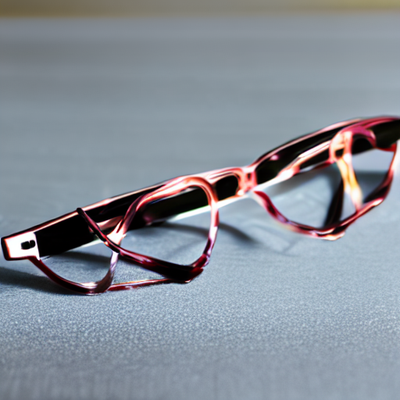} & 
     \includegraphics[height=1.51cm]{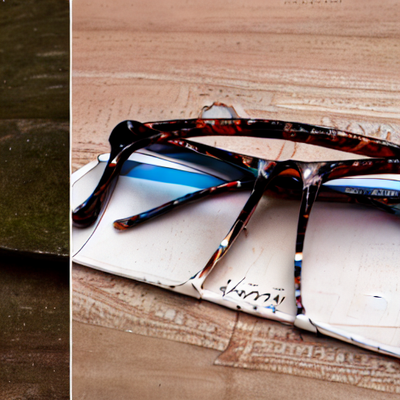}
    \end{tabular}
        \vspace{-0.15cm}
    \caption{Qualitative results with and without MIMA against three concepts across four personalization methods.
    }
    \vspace{-0.3cm}
    \label{fig:qual_per_single}
\end{figure}

\subsection{Multi-concept Personalization Immunization }
Following IMMA, we evaluate MIMA against learning unique/personalized concepts under four adaptation methods: Textual Inversion (TI)~\cite{gal2022textual}, DreamBooth~\cite{ruiz2022dreambooth} (DB),  DreamBooth LoRA, and Custom Diffusion (CD)~\cite{kumari2022customdiffusion}. %

\myparagraph{Experiment details.} We conduct the experiments on thirteen unique concepts from~\citet{kumari2022customdiffusion}, including pets, furniture, scenes, decor items, \etc. Each of them contains four to six real-world images of a personalized/unique concept. To form the concept sets, we randomly select two to three concepts among them. %
For MIMA training, we pair each unique concept with a unique token in the prompt. We train MIMA with personalized images and the prompt containing the unique token. 
For adaptation, we consider the four aforementioned adaptation methods on top of the same immunized weights to study the effect. The evaluation prompt for all concepts is ``A photo of $[V^\ast]$'', with \textit{different} special tokens during MIMA training phrases. The regularization concept is set to each target concept's category name, \eg, \textit{``cat''} or \textit{``plant''}. As in the re-learning task, we generated 200 images for the regularization of MIMA.

\myparagraph{Evaluation metrics.} 
Beyond \msgr, we also want to show that the model maintains its capacity of being fine-tuned to generate \textit{other} concepts. Hence, we introduce the Mean Relative Similarity Gap Ratio (\mrsgr). This metric measures the performance gap between the target and other concepts for models with and without immunization, where the performance is measured as the average similarity between generations from models with and without MIMA across all concepts. 

Formally, we denote $(\rvx^\gI_{[n]}, \rvx^\gA_{[n]})$ as the generated images, after adaptation, with and without MIMA for $n$-th target concept set $\gC$. $(\rvx^\gI_{o,[n']}, \rvx^\gA_{o,[n']})$ are generated images with and without MIMA on $n'$-th \textit{other unique concept} in the set of other concepts $\gC_{\text{o}}$. We define $\texttt{MRSGR}(\{(\rvx^\gI_{[n]}, \rvx^\gA_{[n]})\}, \{(\rvx^\gI_{o, [n']}, \rvx^\gA_{o, [n']})\})$ as 
\bea
        \label{eq:rsgr}
\frac{\overbrace{\bar\gM(\{(\rvx^\gI_{o, [n']}, \rvx^\gA_{o, [n']})\})}^{\text{Other concept} 
 (\uparrow)} - \overbrace{\bar\gM(\{(\rvx^\gI_{[n]}, \rvx^\gA_{[n]})\})}^{\text{Target concepts} (\downarrow)}}{{\bar\gM(\{(\rvx^\gI_{o, [n']}, \rvx^\gA_{o, [n']})\})}},
\eea
with average similarity $\bar\gM$ over the image pairs defined as:
\bea
\bar\gM (\{(\rvx^\gI_{[n]}, \rvx^\gA_{[n]})\}) = \frac{1}{|\gC|} \sum_{n=1}^{|\gC|} \gM(\rvx^\gI_{[n]}, \rvx^\gA_{[n]}) \text{~~and~}\\ 
\bar\gM (\{(\rvx^\gI_{o,[n']}, \rvx^\gA_{o, [n']})\}) = \frac{1}{|\gC_\text{o}|} \sum_{n'=1}^{|\gC_\text{o}|} \gM(\rvx^\gI_{o,[n']}, \rvx^\gA_{o,[n']}).
\eea
A larger \mrsgr indicates a better effect at preserving the other concepts when immunizing the model against the target concepts. To show one single immunized model is effective against multiple adaptation methods, we report the averaged \msgr and \mrsgr over the four adaptation methods.

\myparagraph{Personalization  results.}
In~\tabref{tab:sgr_per}, we report \msgr of immunization against personalization adaptation on five 2-concept and 3-concept sets. We observe positive ratios across all sets and all evaluation metrics. Overall, MIMA has the largest ratios across different sets and evaluation metrics, which indicates that MIMA 
most effectively protects the pre-trained model. All the results in the tables are reported at the $40^{\tt th}$ step for all adaptations. 

To show that MIMA maintains the capacity to learn other personalized concepts with the adaptation methods, we report \mrsgr in~\tabref{tab:rsgr_per}.  We denote concepts other than the target concepts as ``other concepts'' within the thirteen personalization concepts. As we can see, \mrsgr of MIMA is the largest across all concept sets and metrics, which means MIMA is better at preserving the model's ability to personalize other concepts. Additionally, we provide the \msgr metric against adaptation training steps in~\figref{fig:sgr_per}. We can observe a solid gap between with and without MIMA. The gap is larger than that of \bsla and \bslb, which shows that MIMA is more effective than the baselines at immunizing the model.
In~\figref{fig:rsgr_per}, we show that \mrsgr stays positive and is higher than baselines'. 

Finally, we show the generated images with and without MIMA's adaptation in~\figref{fig:qual_per_single}. Compared with the reference images in the first column, models with MIMA do not generate the exact personal item or generate an unrelated image. In other words, MIMA protects the model from being adapted to personal concepts. 

\section{Conclusion}
In this work, we aim to mitigate the risk associated with the open-sourcing of text-to-image models by studying the mitigation method based on the model immunization paradigm of IMMA~\cite{zheng2023imma}. We generalized the setting by considering multi-concept immunization, which takes the immunization algorithms one step closer to the real-world scenario. We then propose MIMA, a multi-concept immunization algorithm that makes a pre-trained model difficult to fine-tune on \textit{multiple} harmful concepts. MIMA leverages a differntaible merge layer that combines model weights to achieve multi-concept immunization. Empirically, MIMA outperforms two IMMA-inspired baselines and is effective in protecting the pre-trained model over various malicious adaptation settings.

\section*{Acknowledgements}
This project is supported in part by an NSF Award \#2420724 and the Ross-Lynn Research Scholar Grant.
\bibliography{egbib}

\clearpage
\appendix

{\noindent\bf \LARGE Appendix
}
\vspace{0.25cm}

\noindent The appendix is organized as follows:
\begin{itemize}[topsep=2pt]
    \item We provide the additional results of MIMA in~\secref{sec:add_res}.
    \item We provide the derivation of model merging optimization in~\secref{sec:derivation}.
    \item We document the experiment details in~\secref{sec:exp_detail}. We will release the code upon acceptance of the paper.
\end{itemize}

\section{Additional results}
\label{sec:add_res}

\myparagraph{MIMA against re-learning object results.}
In~\tabref{tab:sgr_erase_obj}, we report the \msgr of re-learning sets of objects after they were erased by UCE~\cite{gandikota2024unified}.
When averaged over all similarity metrics, MIMA outperforms the baselines with gains of 22.59\% over \bsla and 14.82\% over \bslb.

\begin{figure*}[h]
    \centering
    \setlength{\tabcolsep}{0.5pt}
    \begin{tabular}{cc@{\hskip 3pt}|@{\hskip 3pt}ccc}
    \multicolumn{2}{c}{\bf 2-concept} & \multicolumn{3}{c}{\bf 3-concept } \\
     chain saw & French horn & gas pump & golf ball & garbage truck  \\
     \includegraphics[height=2.0cm, trim={10 0.4cm 10 0.4cm},clip]{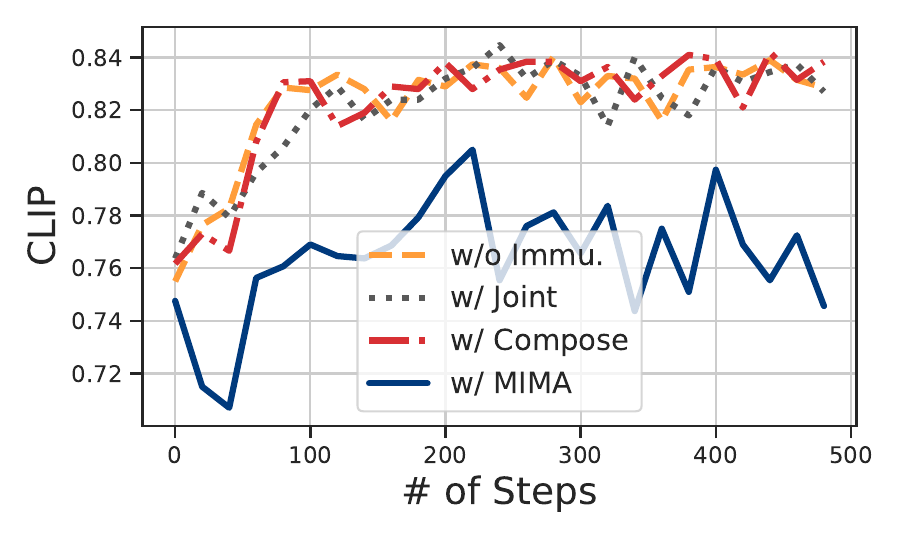} & \includegraphics[height=2.0cm, trim={10 0.4cm 10 0.4cm},clip]{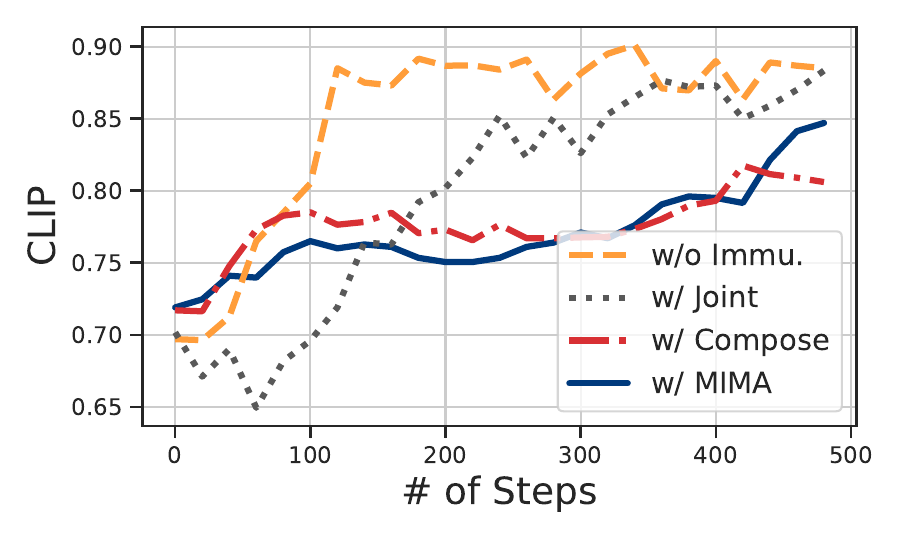} &
     \includegraphics[height=2.0cm, trim={10 0.4cm 10 0.4cm},clip]{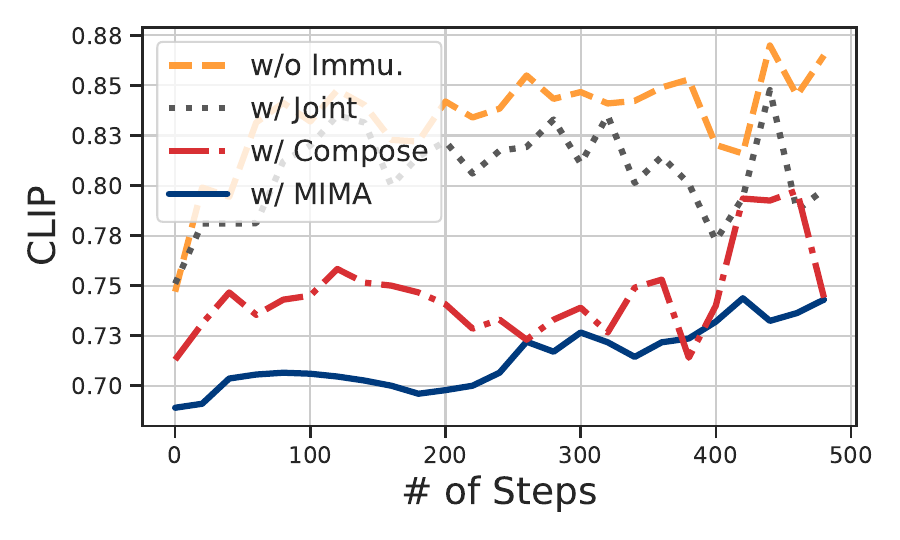} & \includegraphics[height=2.0cm, trim={10 0.4cm 10 0.4cm},clip]{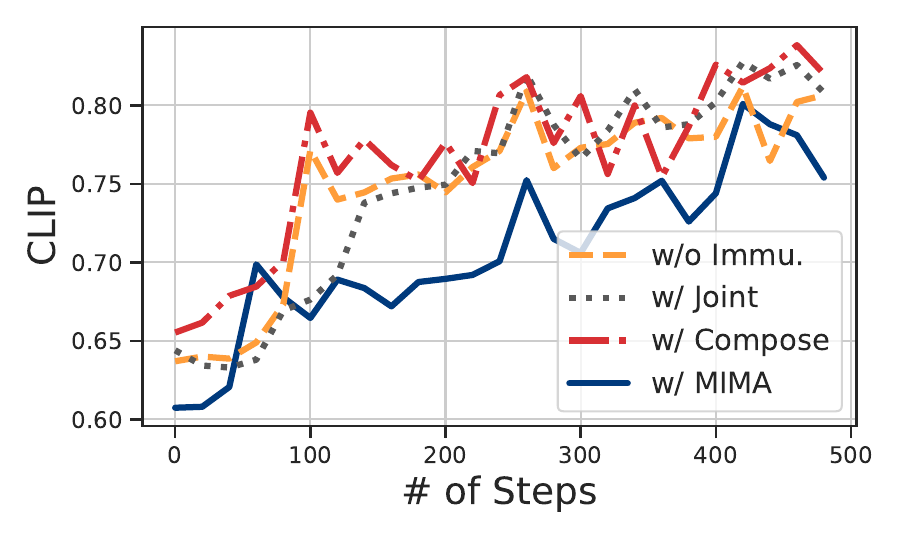} &
     \includegraphics[height=2.0cm, trim={10 0.4cm 10 0.4cm},clip]{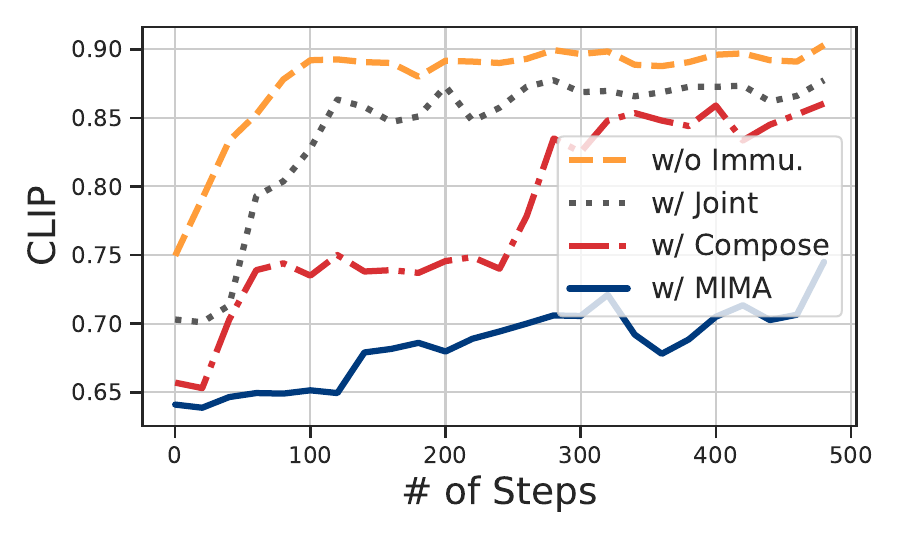} \\
     \includegraphics[height=2.0cm, trim={10 0.4cm 10 0.4cm},clip]{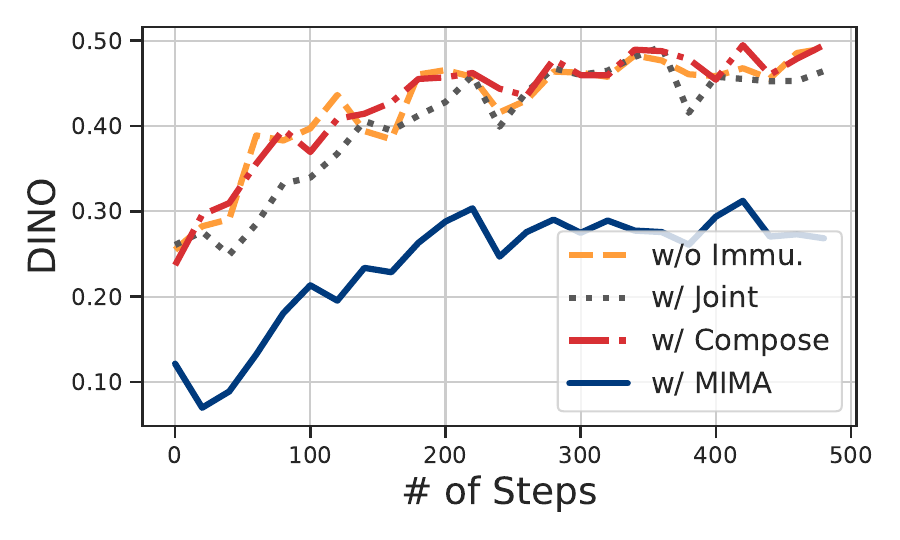} & \includegraphics[height=2.0cm, trim={10 0.4cm 10 0.4cm},clip]{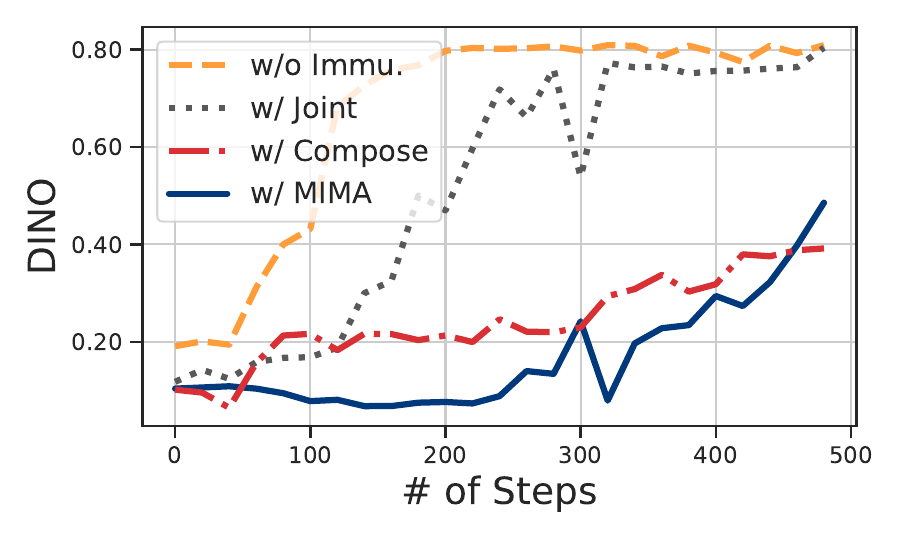} & \includegraphics[height=2.0cm, trim={10 0.4cm 10 0.4cm},clip]{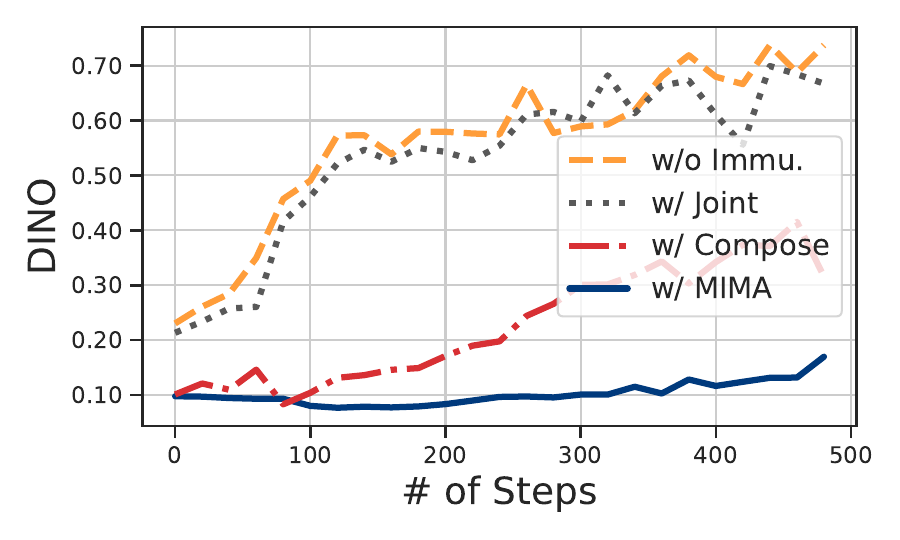} & \includegraphics[height=2.0cm, trim={10 0.4cm 10 0.4cm},clip]{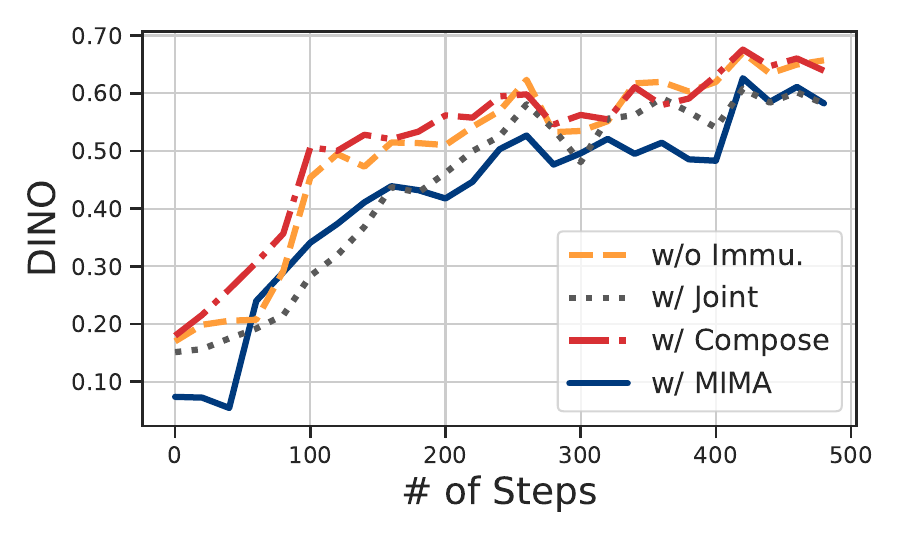} & 
     \includegraphics[height=2.0cm, trim={10 0.4cm 10 0.4cm},clip]{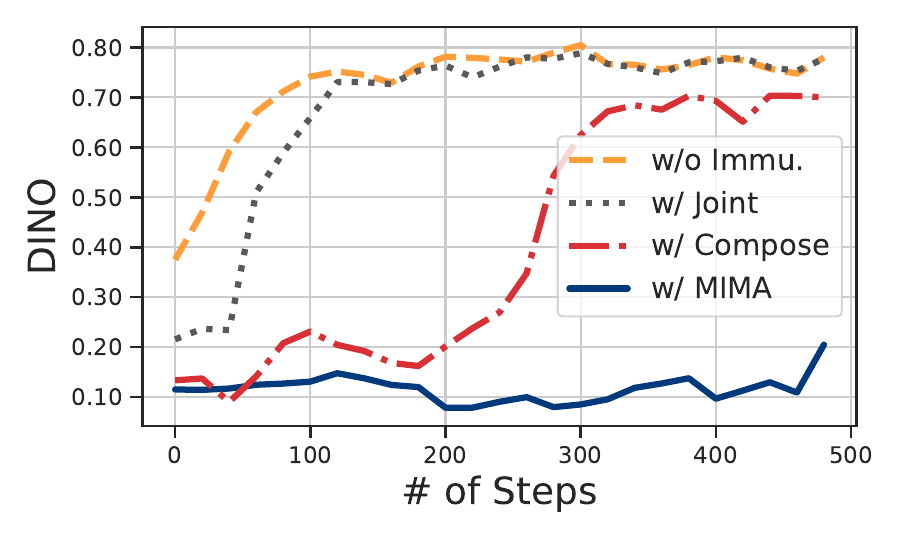} \\
     \includegraphics[height=2.0cm, trim={10 0.4cm 10 0.4cm},clip]{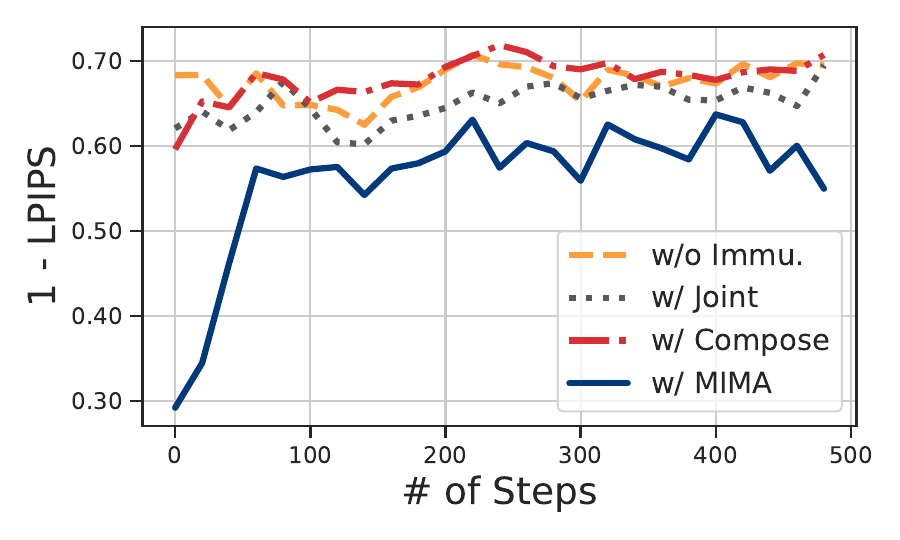} & \includegraphics[height=2.0cm, trim={10 0.4cm 10 0.4cm},clip]{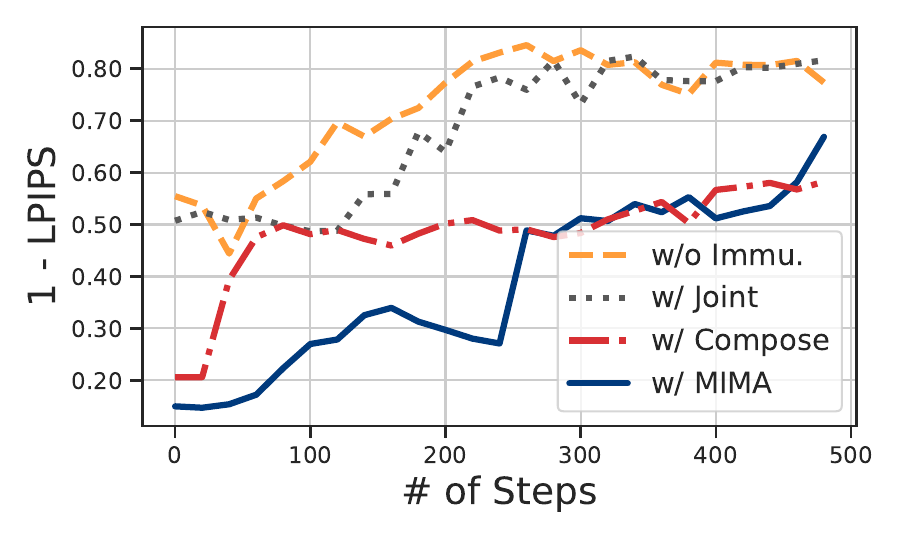} & \includegraphics[height=2.0cm, trim={10 0.4cm 10 0.4cm},clip]{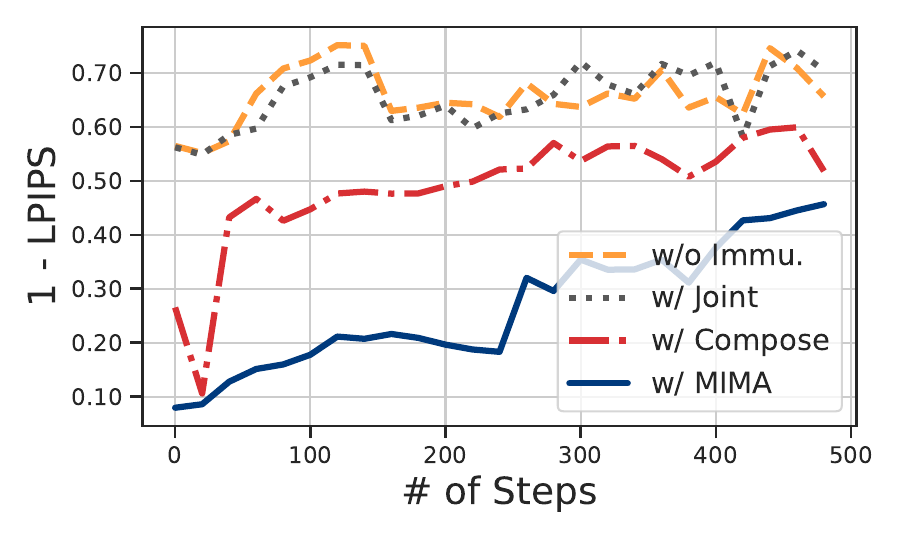} & \includegraphics[height=2.0cm, trim={10 0.4cm 10 0.4cm},clip]{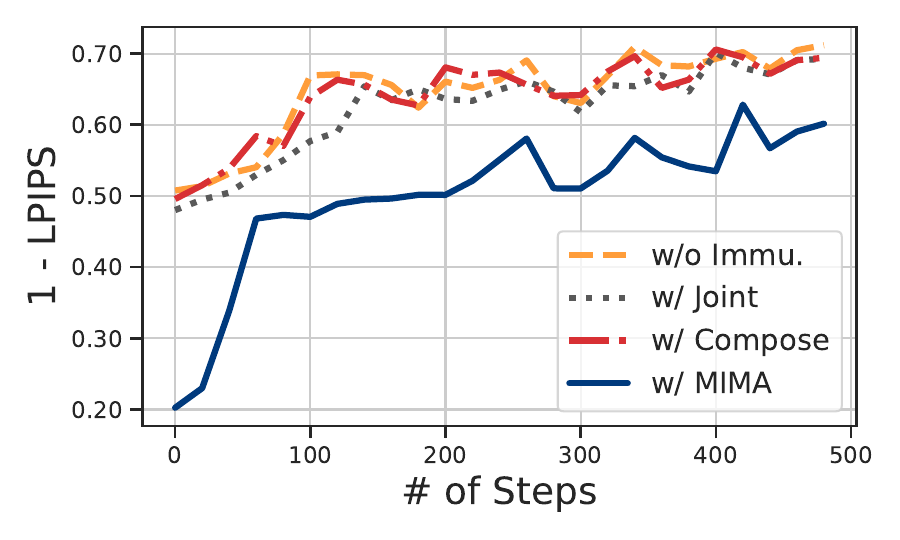} & 
     \includegraphics[height=2.0cm, trim={10 0.4cm 10 0.4cm},clip]{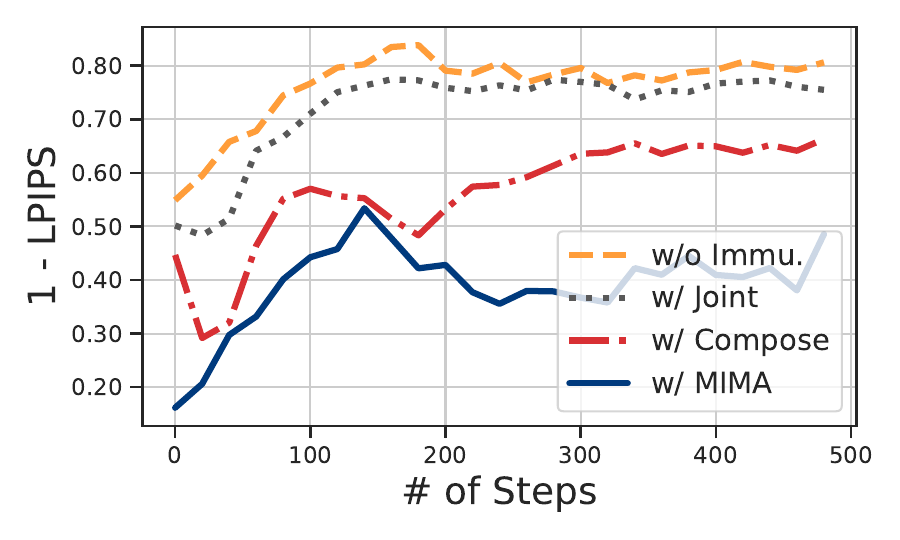}\\
    \end{tabular}
    \vspace{-0.15cm}
    \caption{{ Similarity \vs epochs for LoRA on objects.} Models with MIMA achieve lower similarity throughout LoRA's steps.
    }
    \vspace{-0.25cm}
    \label{fig:sgr_erase_obj}
\end{figure*}

In~\figref{fig:sgr_erase_obj}, we present the similarity metric \vs adaptation steps for object re-learning. Again, we observe that MIMA generally has the largest gap from the model without immunization, followed by Compose, then Joint.

In~\tabref{tab:cls_obj_erase}, we report classification accuracy to further evaluate the generation quality. Specifically, we report the average accuracy of the three target concepts with and without immunization using the ResNet50 model pre-trained on ImageNet. If the immunization is successful then the accuracy should be low, as the generations would not contain the object.
\begin{table}[h]
\footnotesize
\centering
\setlength{\tabcolsep}{2pt}
\resizebox{1.\linewidth}{!}{
\begin{tabular}{llccccc@{\hskip 10pt}ccccc}
\specialrule{.15em}{.05em}{.05em}
           \multicolumn{2}{c}{\multirow{2}{*}{Group \#}} & \multicolumn{5}{c}{2-concept} & \multicolumn{5}{c}{3-concept} \\ \cline{3-7} \cline{8-12}
           && 1  & 2 & 3 & 4  & 5 & 1  & 2 & 3 & 4 & 5  \\
\hline
\multirow{3}{*}{\bsla}
&\texttt{(C)}     & 0.94     & 2.78          & 2.19        & 1.50        & -1.95    & 0.83 & 1.40       & 0.70   & 2.08        &\bf 5.08              \\
 &\texttt{(D)}    & 2.44     & 7.73         & 2.31       & 11.5          & -3.41    & 4.81 & 3.84     & 3.30  & 6.35           & 7.87            \\
  &\texttt{(L)}     & 3.66     & -0.35         & 3.13       & 4.96          & 6.65  & 3.57 & 2.14     & -1.93   & 2.12           & 3.37            \\
 \hline
\multirow{3}{*}{\bslb} 
&\texttt{(C)}     & 3.41     & 10.5          & 2.62        & 1.95           & 5.49 & 7.73 & 1.17       & -2.30     & 6.24         & 1.92              \\
 &\texttt{(D)}    & 30.4     & 32.1         & 5.77       & 24.0          & 29.9 & 16.9& 4.51     & 3.67     & 23.8          & 10.2            \\
  &\texttt{(L)}   & 14.8     & 18.0         & 9.83       &  4.05         & 21.3  & 6.65 & 4.68     & 2.82  & 17.2           & 3.53           \\
 \hline
\multirow{3}{*}{Ours} 
&\texttt{(C)}     &\bf 8.03     &\bf 11.0          &\bf 11.7  &\bf 7.69   &\bf 17.3 &\bf 11.7  & \bf 10.5       &\bf 5.94    &\bf 8.28        & 3.04               \\
 &\texttt{(D)}    &\bf 49.5    &\bf 37.5         &\bf 45.7  &\bf 52.5   &\bf 55.2 &\bf 48.5  &\bf 28.3     &\bf 37.1   &\bf 34.3           &\bf 22.9            \\
 &\texttt{(L)}    &\bf 21.2    &\bf 20.5         &\bf 25.8   &\bf 21.1  &\bf 49.5 &\bf 37.1  &\bf 20.5     &\bf 21.1   &\bf 32.7           &\bf 11.3            \\
\specialrule{.15em}{.05em}{.05em}
\end{tabular}
}
    \vspace{-0.15cm}
\caption{\msgr$\uparrow$(\%) on objects for UCE with LoRA. MIMA shows an average \msgr improvement of 22.59\% over \bsla and 14.82\% over \bslb across all three similarity metrics.
}
\vspace{-0.15cm}
\label{tab:sgr_erase_obj}
\end{table}

For the model without immunization, we observe that LoRA relearns the object as indicated by the high classification accuracy. In contrast, the model with MIMA fails to learn them back. 

Next, to ensure that the immunized model remains useful, 
we check whether MIMA maintains the generation capability on \textit{other objects}. We define \textit{other concepts} as the seven classes that are not used as the three target concepts. %
As shown in the right block of~\tabref{tab:cls_obj_erase}, the average accuracy of the other concepts is similar with and without MIMA, indicating that the immunized model maintains its usability. 
\begin{figure}[h]
    \hspace{-.5cm}
    \centering
    \footnotesize
    \setlength{\tabcolsep}{1.2pt}
    \renewcommand{\arraystretch}{1.2}
    \begin{tabular}{lcc@{\hskip 5pt}cc}
    & \multicolumn{2}{c}{Stable Diffusion} & \multicolumn{2}{c}{Re-learning}\vspace{-3pt}  \\
    & Reference  & Erased & w/o Immu. &  w/ MIMA \\
    \multirow{2}{*}[1.1cm]{\rotatebox[origin=c]{90}{cassette}} & 
    \includegraphics[height=1.8cm,width=1.8cm]{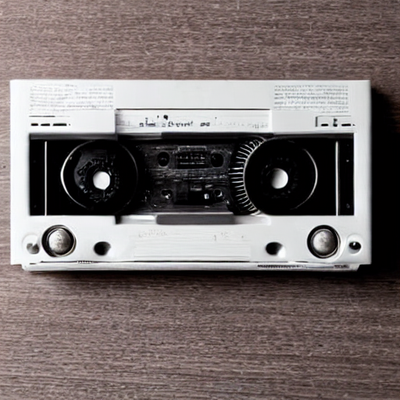} & 
    \includegraphics[height=1.8cm,width=1.8cm]{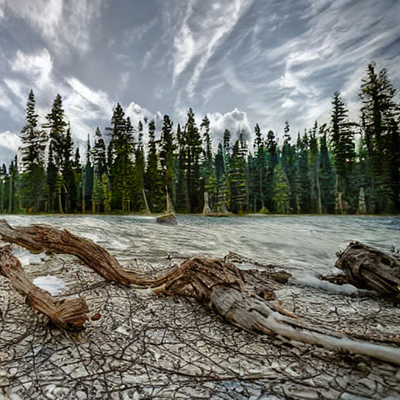}  &
    \includegraphics[height=1.8cm,width=1.8cm]{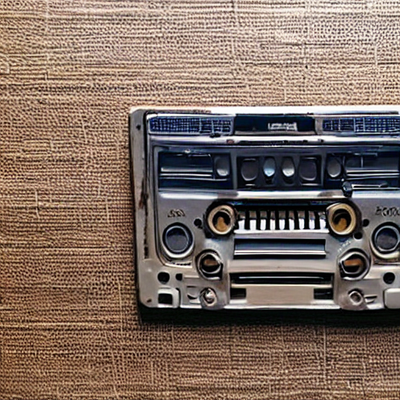}  &
    \includegraphics[height=1.8cm,width=1.8cm]{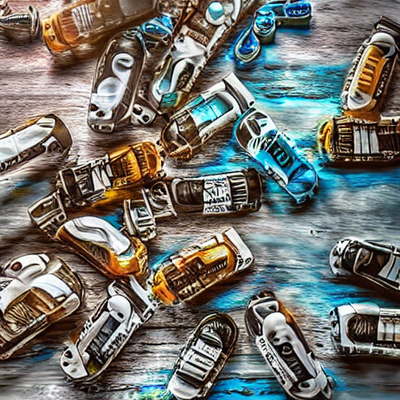}  \\
    \multirow{2}{*}[1.25cm]{\rotatebox[origin=c]{90}{chain saw}} & 
    \includegraphics[height=1.8cm,width=1.8cm]{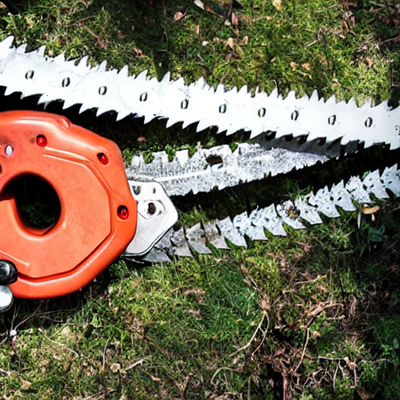} & 
    \includegraphics[height=1.8cm,width=1.8cm]{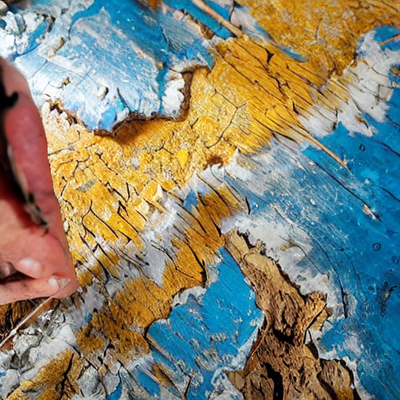}  &
    \includegraphics[height=1.8cm,width=1.8cm]{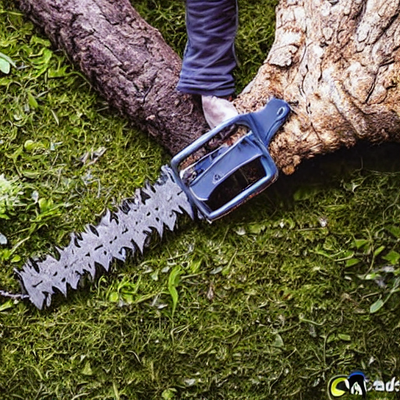}  &
    \includegraphics[height=1.8cm,width=1.8cm]{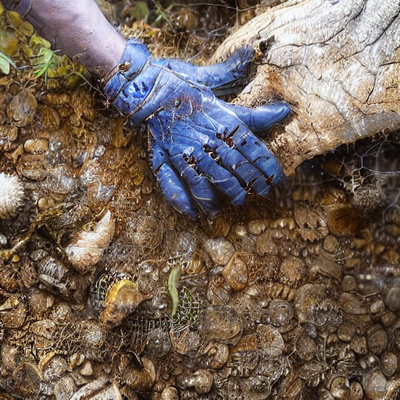}  \\
    \multirow{2}{*}[1cm]{\rotatebox[origin=c]{90}{church}} & 
    \includegraphics[height=1.8cm,width=1.8cm]{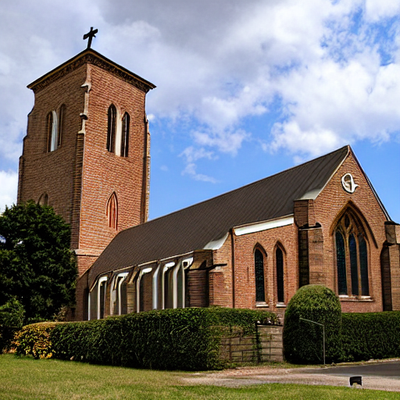} & 
    \includegraphics[height=1.8cm,width=1.8cm]{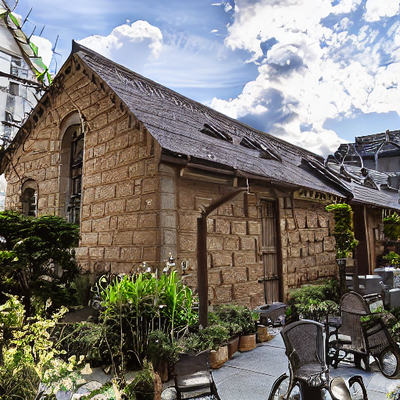}  &
    \includegraphics[height=1.8cm,width=1.8cm]{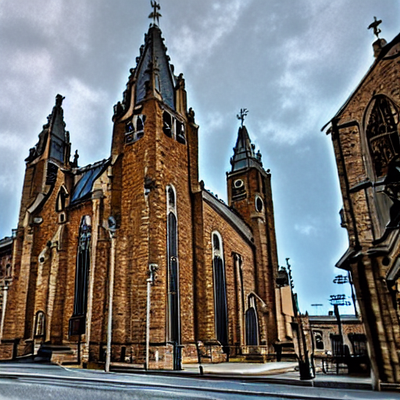}  &
    \includegraphics[height=1.8cm,width=1.8cm]{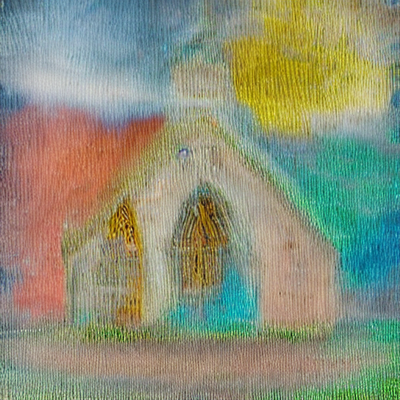} \\
    \end{tabular}
        \vspace{-0.15cm}
    \captionof{figure}{Qualitative result of MIMA against re-learning objects. Both UCE and MIMA are adapted to all three concepts. 
    }
    \vspace{-0.3cm}
    \label{fig:qual_erase_obj}
\end{figure}

Finally, qualitative results in~\figref{fig:qual_erase_obj} also confirm that MIMA successfully protects against re-learning as the generated images do not contain the target concepts. 
\begin{table}[h]
\centering
\footnotesize %
\setlength{\tabcolsep}{2.3pt}
\resizebox{\columnwidth}{!}{
\begin{tabular}{lccccc|cccc}
\specialrule{.15em}{.05em}{.05em}
 &&\multicolumn{4}{c|}{Acc. of  target ($\downarrow$)} & \multicolumn{4}{c}{Acc. of others ($\uparrow$)} \\ \cline{3-10}
Mod. & UCE    &  w/o Immu.    &  \bsla  & \bslb    & MIMA     & UCE      & \bsla  & \bslb            & MIMA                 \\\hline
1  & 3.20       &  34.1        &  0.67           &    0.67        &\bf  0.00   & 88.5    & 86.9 &\bf 89.3 & 87.7 \\
2  & 1.27       &  38.7        &  23.5           &   25.5       &\bf 0.20    & 71.7    &\bf 74.7 & 67.0 & 64.4   \\
3  & 1.33       &  44.3         &  22.6           &  26.6       &\bf 4.1    & 84.8    &\bf 90.0 & 79.7 & 79.2   \\ \hline
Avg.   &  1.93      &  39.0        &  15.6        &   17.6   &\bf 1.42 & 81.7 &\bf 83.9 & 78.7 & 77.1    \\
\specialrule{.15em}{.05em}{.05em}
\end{tabular}
}
    \vspace{-0.15cm}
\caption{{Acc. (\%) of object-erased models on 500 images.} 
{\it Col.~1:} Original UCE model with the target concepts erased. {\it Col.~2:} UCE without Immunization after LoRA. We observe that the target concept is successfully relearned. {\it Col.~3-5:} UCE with Immunization after LoRA. {\it Col.~6-9:} Acc. of other objects of UCE before and after Immunization. %
}
\vspace{-0.15cm}
\label{tab:cls_obj_erase}
\end{table}

\begin{figure}[h]
    \vspace{-0.2cm}
    \centering
    \scriptsize
    \setlength{\tabcolsep}{1.1pt}
    \renewcommand{\arraystretch}{1.2}
    \resizebox{1.0\linewidth}{!}{%
    \begin{tabular}{c@{\hskip 6pt}ccc}
      Reference & { w/o Immu.} & {w/ MIMA}\\
    \multirow{1}{*}[1cm]{\rotatebox[origin=c]{90}{cat}}
    \includegraphics[height=1.8cm]{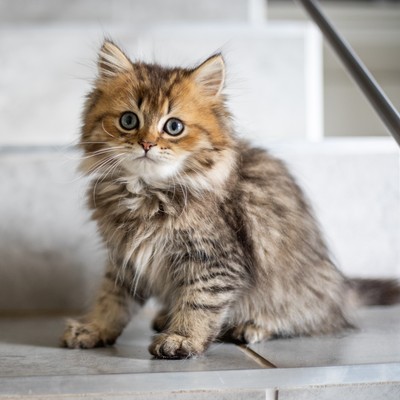} &
    \multirow{1}{*}[1.2cm]{\rotatebox[origin=c]{90}{cat \& pot}}
    \includegraphics[height=1.8cm]{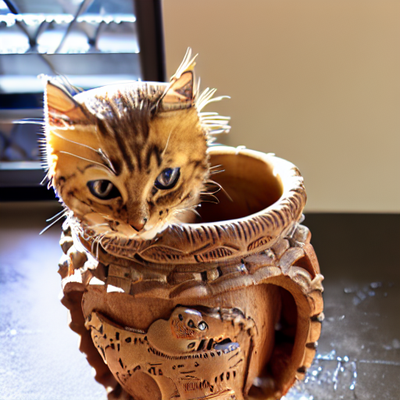} & 
    \includegraphics[height=1.8cm]{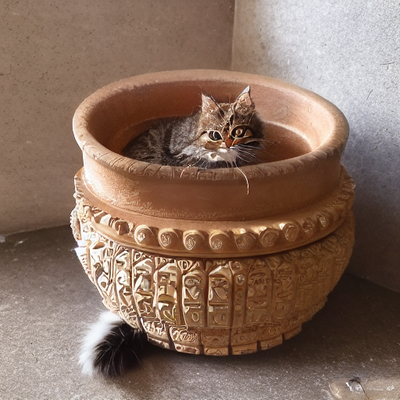} &  \\
    \multirow{1}{*}[1.0cm]{\rotatebox[origin=c]{90}{flower}}
    \includegraphics[height=1.8cm, width=1.5cm]{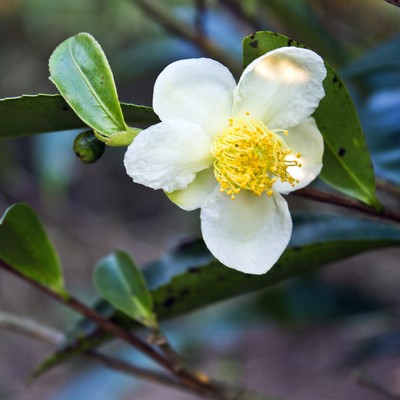} &
     \multirow{2}{*}[1.35cm]{\rotatebox[origin=c]{90}{flower \& pot}}
     \includegraphics[height=1.8cm]{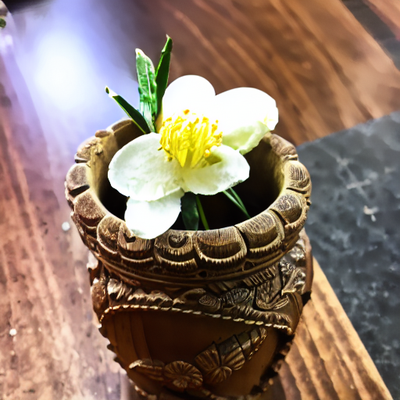} & 
     \includegraphics[height=1.8cm]{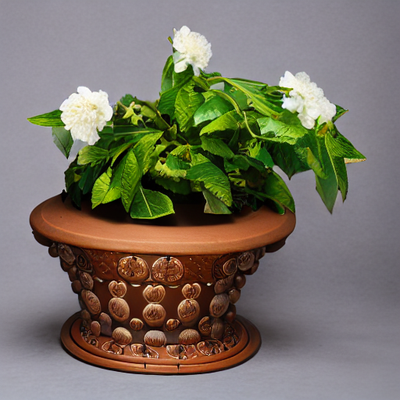} &\\
     \multirow{1}{*}[0.8cm]{\rotatebox[origin=c]{90}{pot}}
    \includegraphics[height=1.8cm, width=1.5cm]{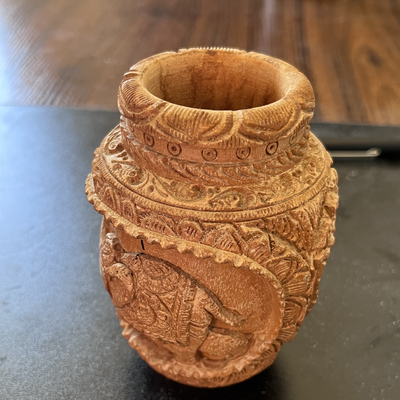} &
     \multirow{2}{*}[1.35cm]{\rotatebox[origin=c]{90}{cat \& flower}}
     \includegraphics[height=1.8cm]{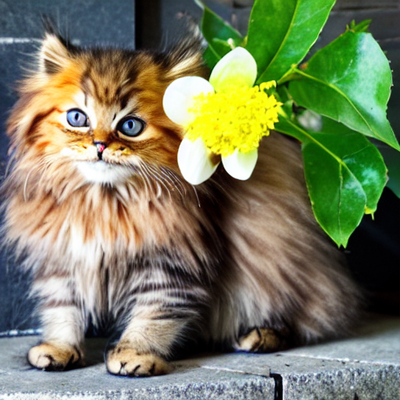} & 
     \includegraphics[height=1.8cm]{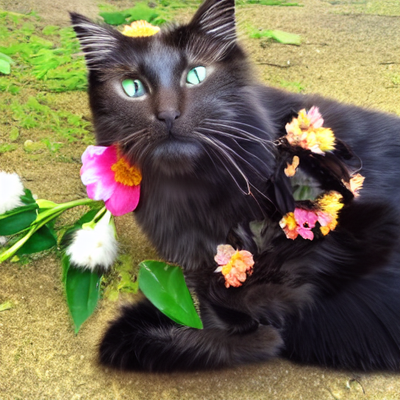} &\\
    \end{tabular}
    }
    \captionof{figure}{{Custom Diffusion on two concepts w/ and w/o MIMA.} Without any immunization, Custom Diffusion successfully learns the target two concepts concurrently. With MIMA, Custom Diffusion fails to learn the specific content in the reference images, \eg, the model with IMMA generates a black cat instead of the cat in the reference image, and the color of the flower is also incorrect. (See third row).}
    \label{fig:qual_per_multi}
\end{figure}

\noindent\myparagraph{Adapt Custom Diffusion after MIMA.}
We provide the results of preventing Custom Diffusion from personalizing two concepts in~\figref{fig:qual_per_multi}. We observe that with MIMA, Custom Diffusion fails to make the model learn the specific target contents in the reference images. 

\begin{figure}[h]
    \centering
    \scriptsize
    \setlength{\tabcolsep}{1.2pt}
    \renewcommand{\arraystretch}{1.2}
    \begin{tabular}{lcc@{\hskip 5pt}cc}
    & \multicolumn{2}{c}{Stable Diffusion} & \multicolumn{2}{c}{Re-learning}\vspace{-3pt}  \\
    & Reference  & Erased & w/o Immu. &  w/ MIMA \\
    \multirow{2}{*}[1.8cm]{\rotatebox[origin=c]{90}{Kelly Mckernan}} & 
    \includegraphics[height=1.8cm,width=1.8cm]{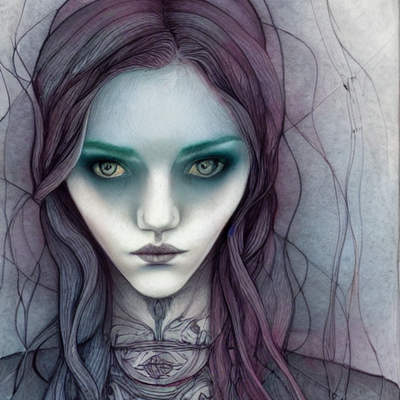} & 
    \includegraphics[height=1.8cm,width=1.8cm]{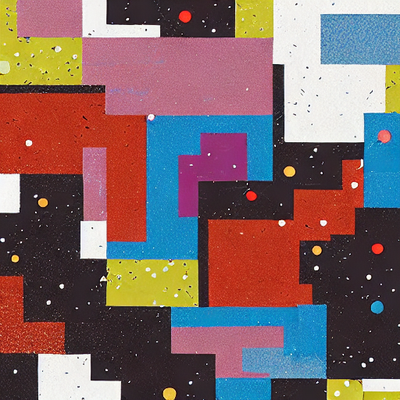}  &
    \includegraphics[height=1.8cm,width=1.8cm]{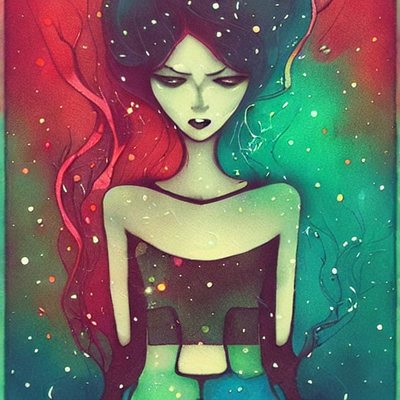}  &
    \includegraphics[height=1.8cm,width=1.8cm]{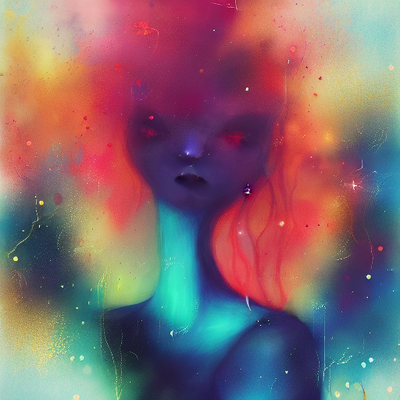}  \\
    \multirow{2}{*}[1.55cm]{\rotatebox[origin=c]{90}{Thomas Kinkade}} & 
    \includegraphics[height=1.8cm,width=1.8cm]{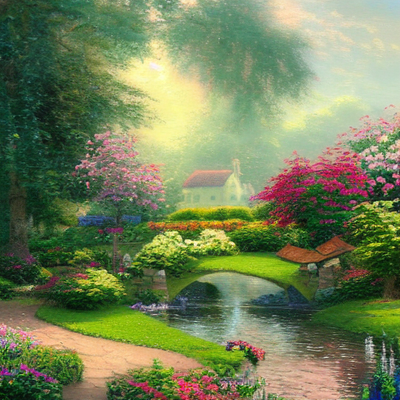} & 
    \includegraphics[height=1.8cm,width=1.8cm]{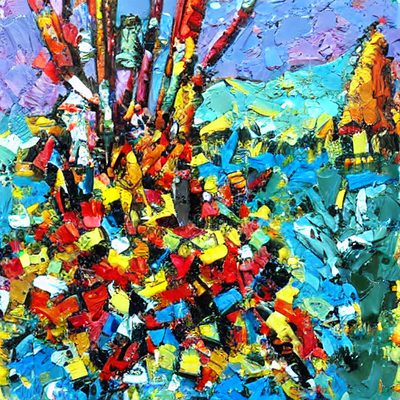}  &
    \includegraphics[height=1.8cm,width=1.8cm]{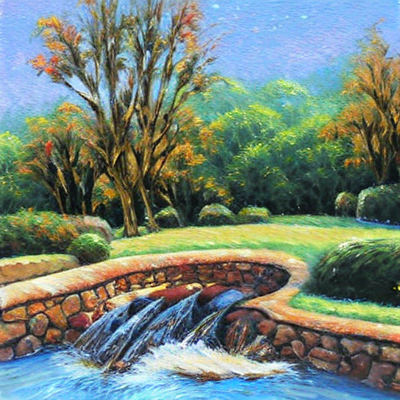}  &
    \includegraphics[height=1.8cm,width=1.8cm]{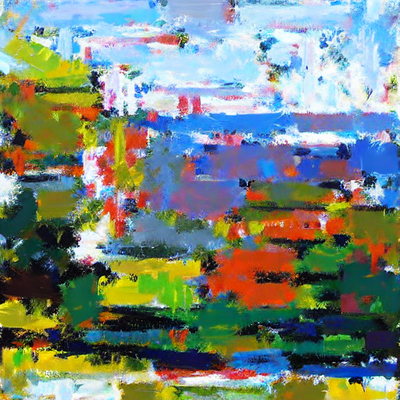}  \\
    \multirow{2}{*}[1.2cm]{\rotatebox[origin=c]{90}{Van Gogh}} & 
    \includegraphics[height=1.8cm,width=1.8cm]{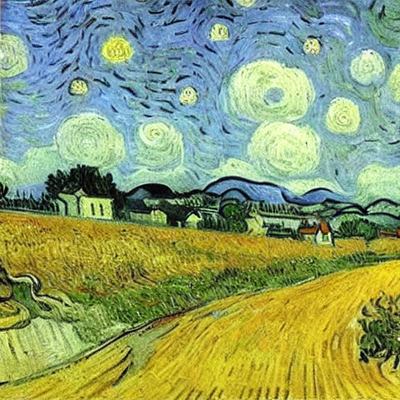} & 
    \includegraphics[height=1.8cm,width=1.8cm]{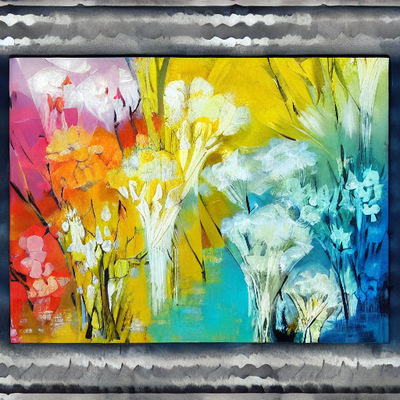}  &
    \includegraphics[height=1.8cm,width=1.8cm]{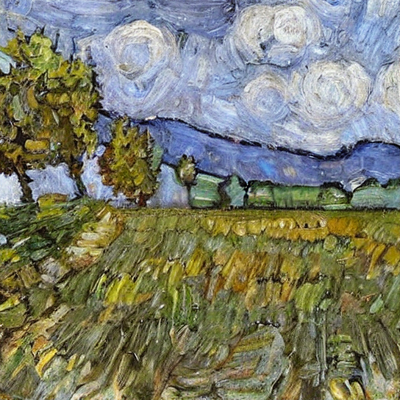}  &
    \includegraphics[height=1.8cm,width=1.8cm]{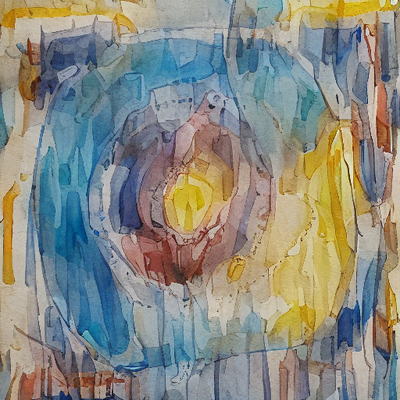} \\
    \end{tabular}
    \vspace{-0.3cm}
    \caption{{Additional qualitative result of MIMA against re-learning artistic styles.} Both Erased and MIMA are adapted to all three concepts on a single model.
    }
    \label{fig:supp_qual_erase_style}
\end{figure}

\begin{figure}[h]
    \centering
    \scriptsize
    \setlength{\tabcolsep}{1.2pt}
    \renewcommand{\arraystretch}{1.2}
    \begin{tabular}{lcc@{\hskip 5pt}cc}
    & \multicolumn{2}{c}{Stable Diffusion} & \multicolumn{2}{c}{Re-learning}\vspace{-3pt}  \\
    & Reference  & Erased & w/o Immu. &  w/ MIMA \\
    \multirow{2}{*}[1.2cm]{\rotatebox[origin=c]{90}{gas pump}} & 
    \includegraphics[height=1.8cm,width=1.8cm]{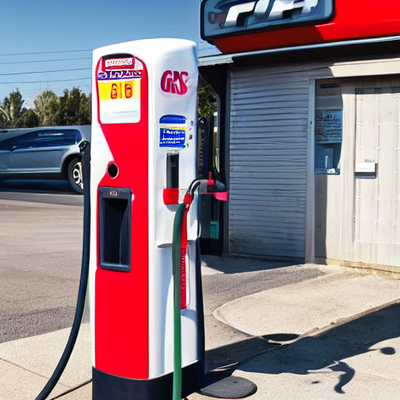} & 
    \includegraphics[height=1.8cm,width=1.8cm]{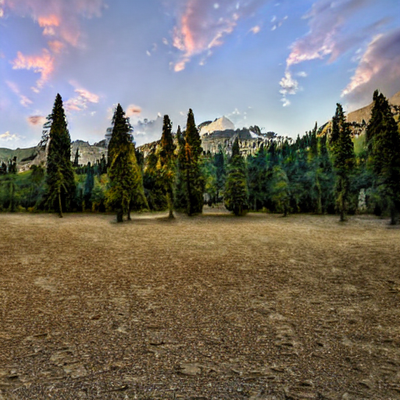}  &
    \includegraphics[height=1.8cm,width=1.8cm]{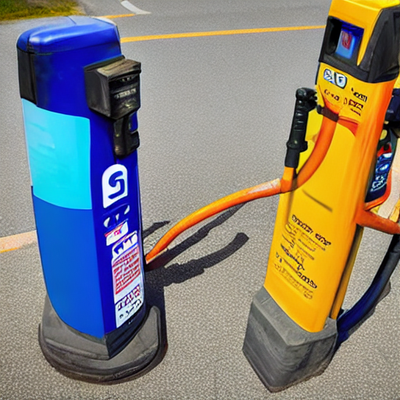}  &
    \includegraphics[height=1.8cm,width=1.8cm]{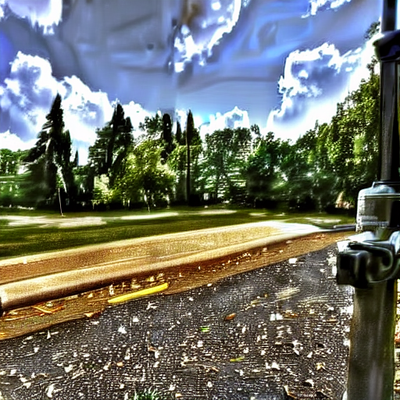}  \\
    \multirow{2}{*}[1.25cm]{\rotatebox[origin=c]{90}{golf ball}} & 
    \includegraphics[height=1.8cm,width=1.8cm]{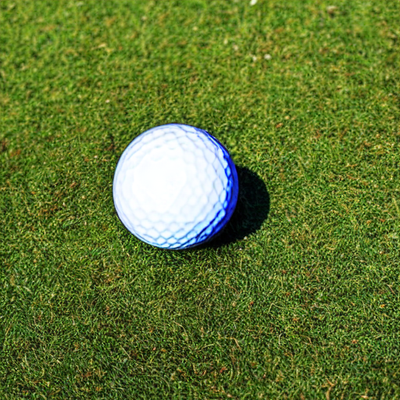} & 
    \includegraphics[height=1.8cm,width=1.8cm]{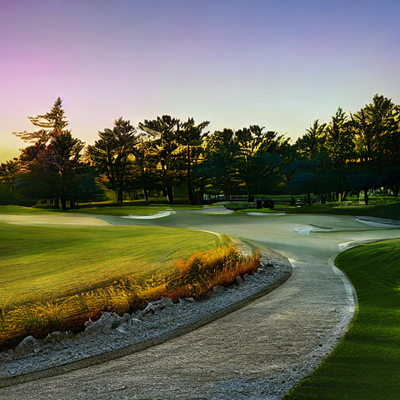}  &
    \includegraphics[height=1.8cm,width=1.8cm]{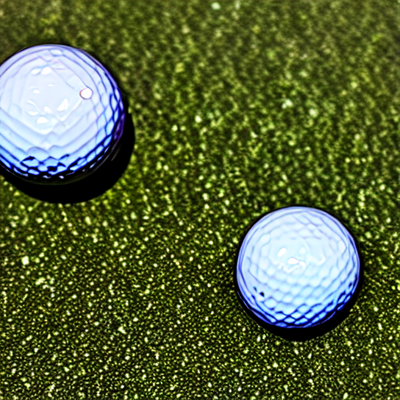}  &
    \includegraphics[height=1.8cm,width=1.8cm]{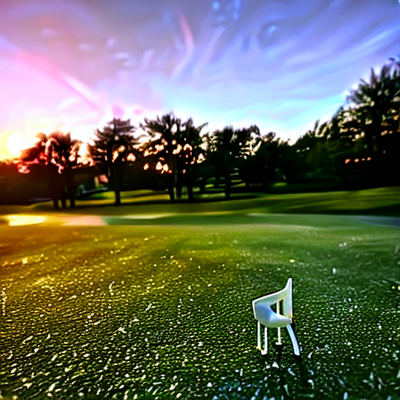}  \\
    \multirow{2}{*}[1.3cm]{\rotatebox[origin=c]{90}{parachute}} & 
    \includegraphics[height=1.8cm,width=1.8cm]{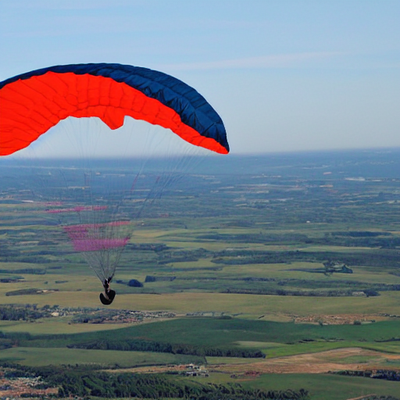} & 
    \includegraphics[height=1.8cm,width=1.8cm]{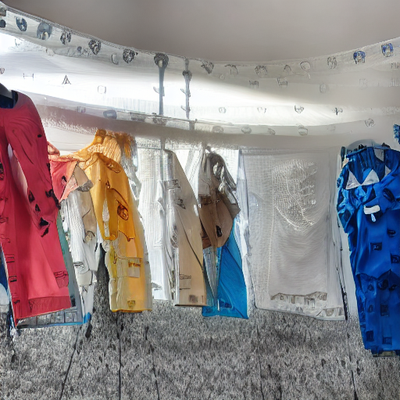}  &
    \includegraphics[height=1.8cm,width=1.8cm]{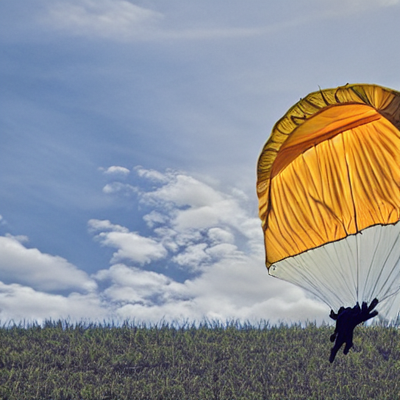}  &
    \includegraphics[height=1.8cm,width=1.8cm]{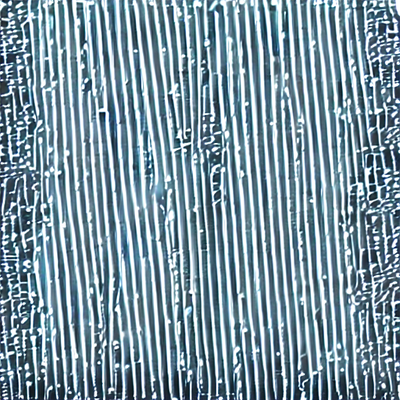} \\
    \end{tabular}
    \vspace{-0.3cm}
    \caption{{Additional qualitative result of MIMA against re-learning objects.} Both Erased and MIMA are adapted to all three concepts on a single model.
    }
    \label{fig:supp_qual_erase_object}
\end{figure}

\begin{figure}[t]
    \centering
    \small
    \setlength{\tabcolsep}{1.3pt}
    \renewcommand{\arraystretch}{1.2}
    \begin{tabular}{c@{\hskip 6pt}ccccc}
      Reference & TI & DB & +LoRA & CD\\
    \includegraphics[height=1.5cm]{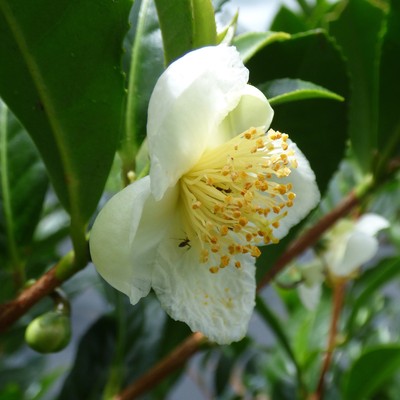} &
    \multirow{1}{*}[1.45cm]{\rotatebox[origin=c]{90}{w/o Immu.}}
    \includegraphics[height=1.5cm]{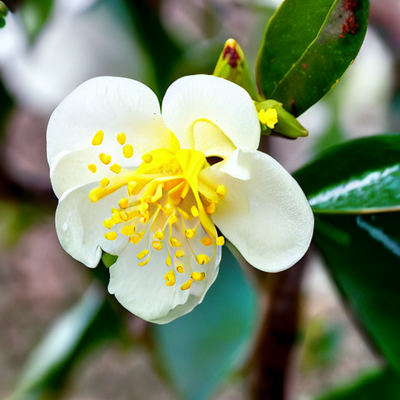} & 
    \includegraphics[height=1.5cm]{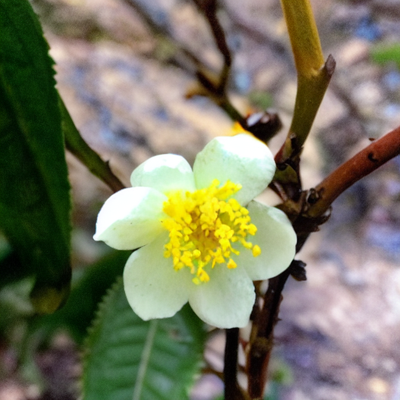} & 
    \includegraphics[height=1.5cm]{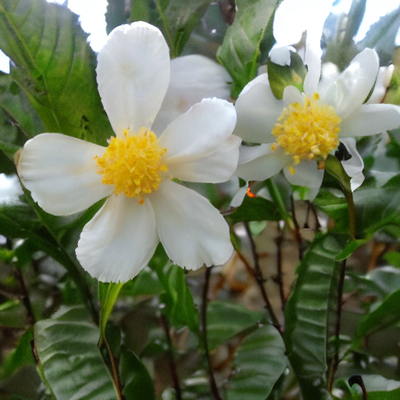} & 
    \includegraphics[height=1.5cm]{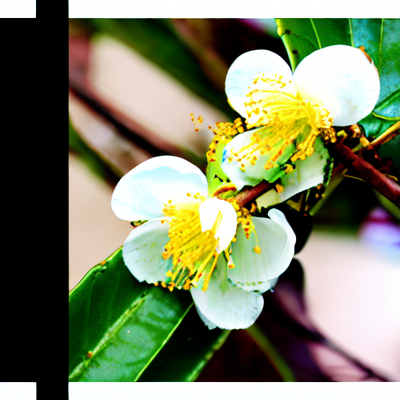} \\
    \includegraphics[height=1.5cm, width=1.7cm]{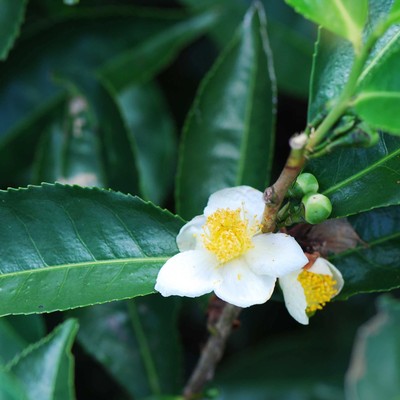} &
     \multirow{2}{*}[1.35cm]{\rotatebox[origin=c]{90}{ w/ MIMA}}
     \includegraphics[height=1.5cm]{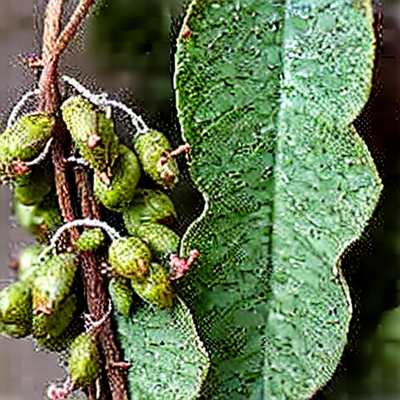} & 
    \includegraphics[height=1.5cm]{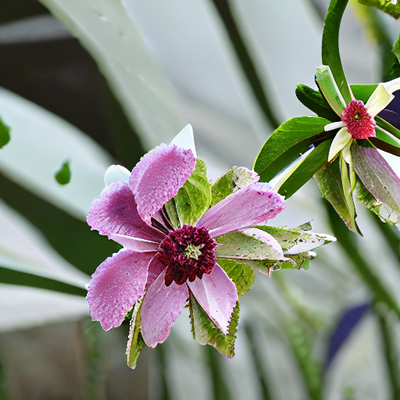} & 
    \includegraphics[height=1.5cm]{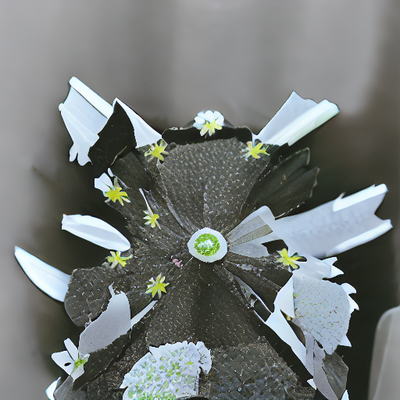} & 
    \includegraphics[height=1.5cm]{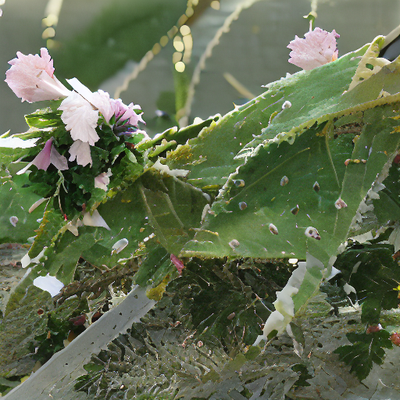} \\
    \includegraphics[height=1.5cm]{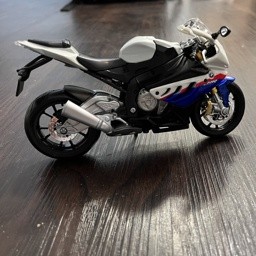} &
    \multirow{1}{*}[1.45cm]{\rotatebox[origin=c]{90}{w/o Immu.}}
    \includegraphics[height=1.5cm]{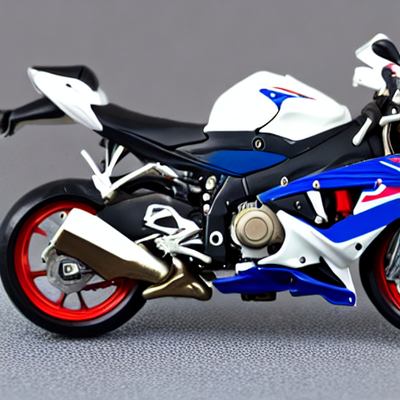} & 
    \includegraphics[height=1.5cm]{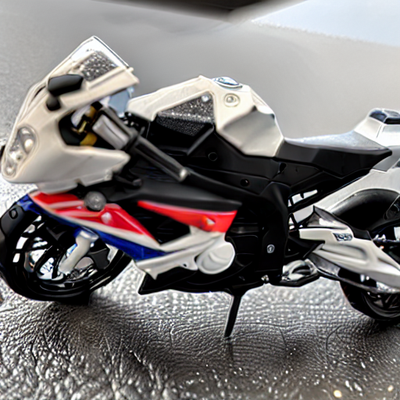} & 
    \includegraphics[height=1.5cm]{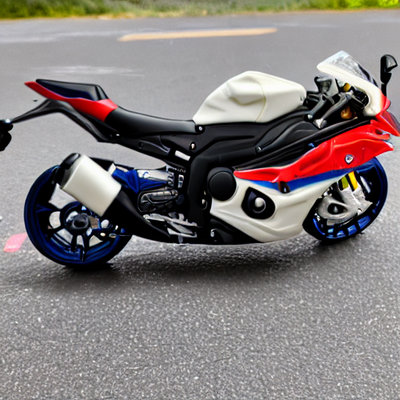} & 
    \includegraphics[height=1.5cm]{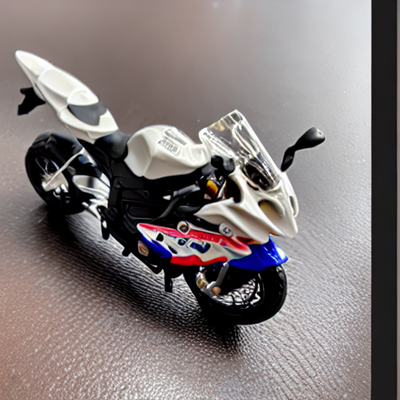} \\
    \includegraphics[height=1.5cm]{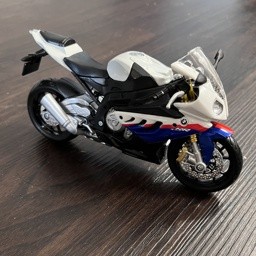} &
     \multirow{2}{*}[1.35cm]{\rotatebox[origin=c]{90}{w/ MIMA}}
     \includegraphics[height=1.5cm]{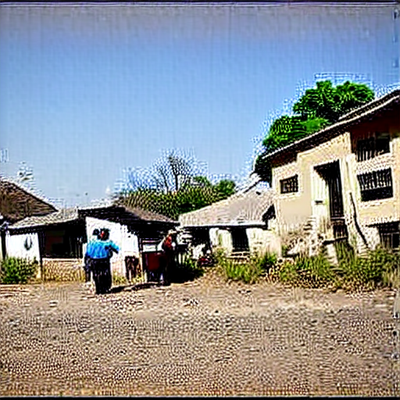} & 
    \includegraphics[height=1.5cm]{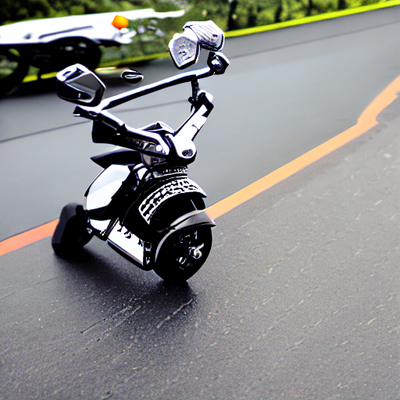} & 
    \includegraphics[height=1.5cm]{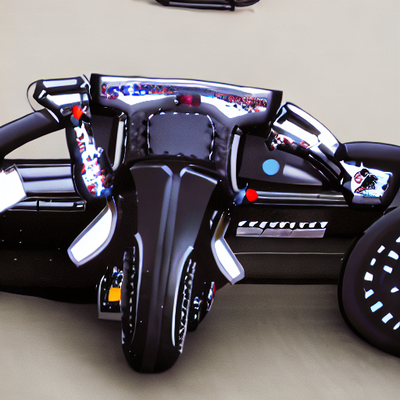} & 
    \includegraphics[height=1.5cm]{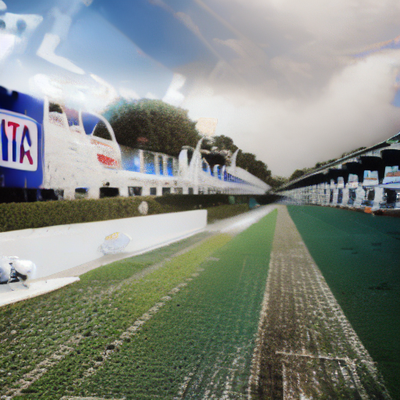} \\
     \includegraphics[height=1.5cm]{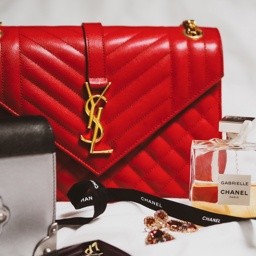} &
    \multirow{1}{*}[1.45cm]{\rotatebox[origin=c]{90}{w/o Immu.}}
    \includegraphics[height=1.5cm]{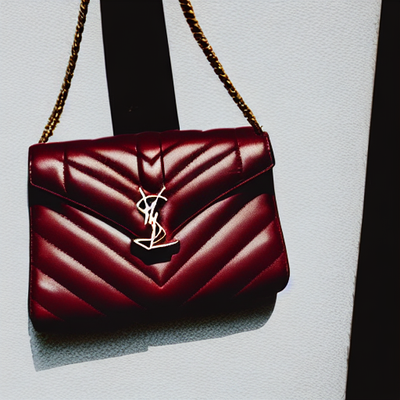} & 
    \includegraphics[height=1.5cm]{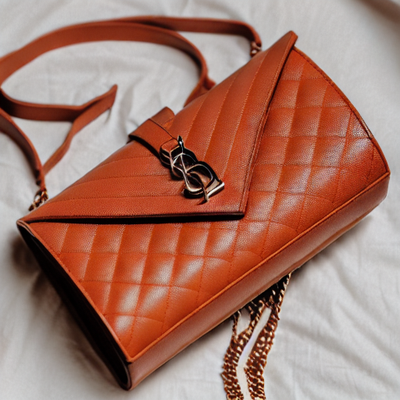} & 
    \includegraphics[height=1.5cm]{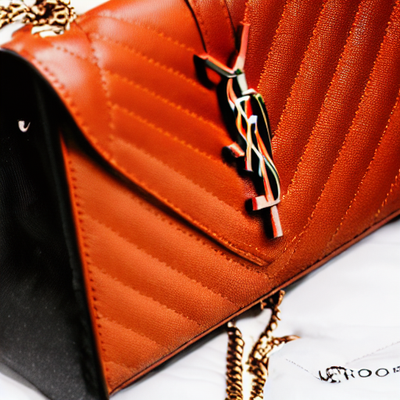} & 
    \includegraphics[height=1.5cm]{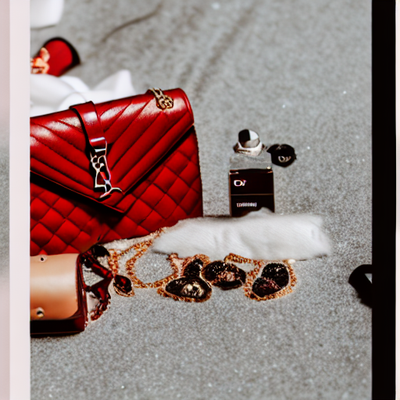} \\
    \includegraphics[height=1.5cm]{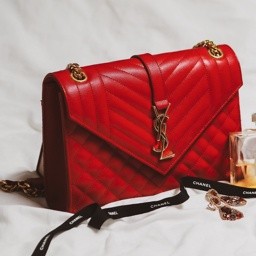} &
     \multirow{2}{*}[1.35cm]{\rotatebox[origin=c]{90}{w/ MIMA}}
     \includegraphics[height=1.5cm]{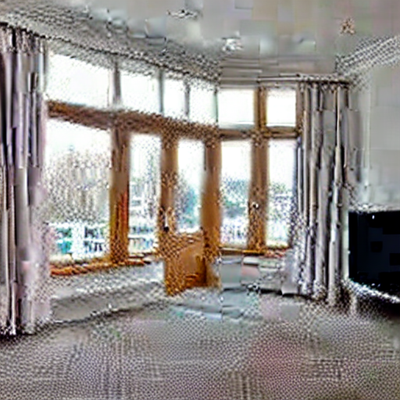} & 
    \includegraphics[height=1.5cm]{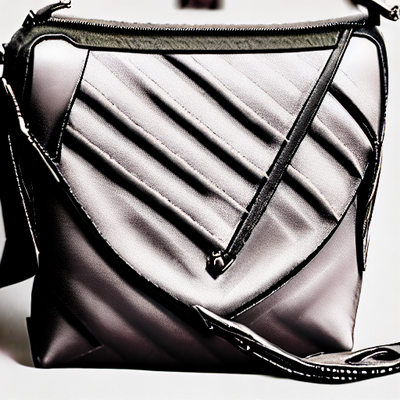} & 
    \includegraphics[height=1.5cm]{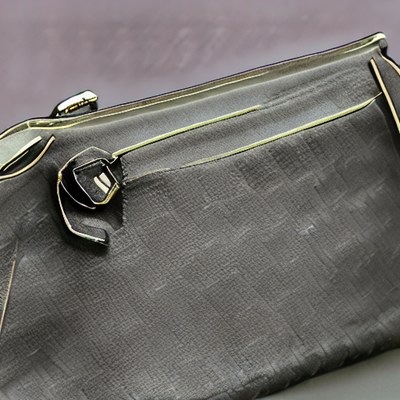} & 
    \includegraphics[height=1.5cm]{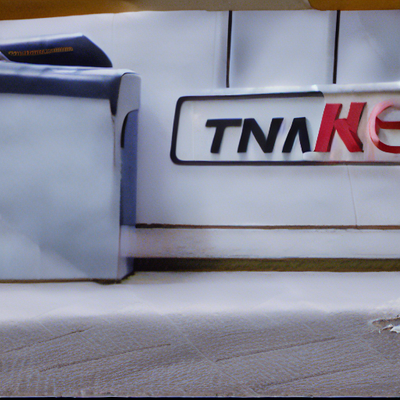} \\
    \end{tabular}
    \vspace{-0.1cm}
    \caption{{Additional personalization adaptation w/ and w/o MIMA against three concepts. %
    }
    }
    \label{fig:supp_qual_per}
    \vspace{-0.3cm}
\end{figure}

\begin{figure}[t]
    \centering
    \small
    \setlength{\tabcolsep}{1.3pt}
    \renewcommand{\arraystretch}{1.2}
    \begin{tabular}{c@{\hskip 6pt}ccccc}
      Reference & TI & DB & +LoRA & CD\\
    \includegraphics[height=1.5cm]{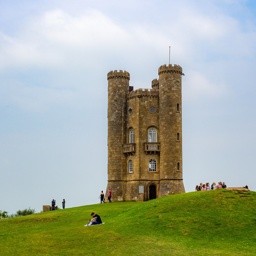} &
    \multirow{1}{*}[1.35cm]{\rotatebox[origin=c]{90}{ w/o Immu.}}
    \includegraphics[height=1.5cm]{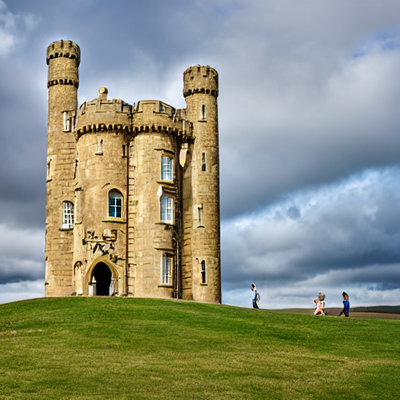} & 
    \includegraphics[height=1.5cm]{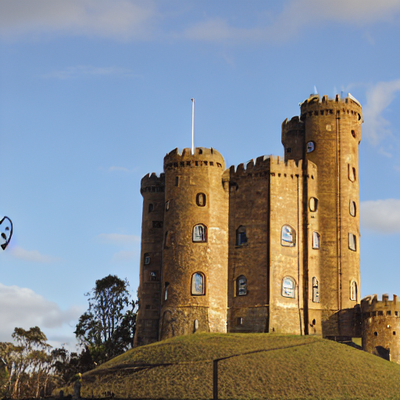} & 
    \includegraphics[height=1.5cm]{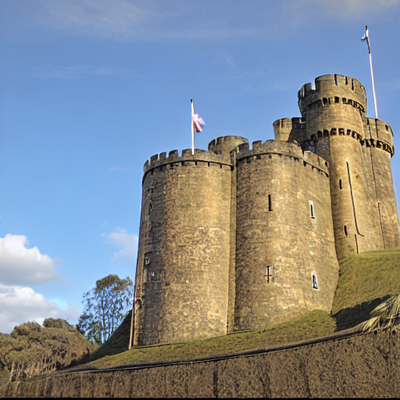} & 
    \includegraphics[height=1.5cm]{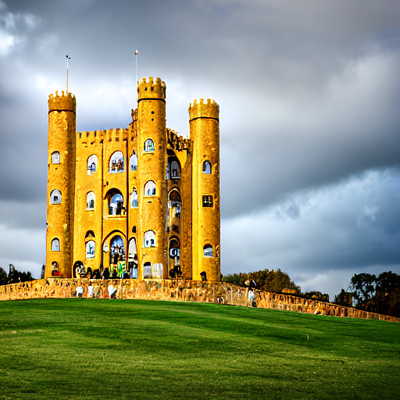} \\
    \includegraphics[height=1.5cm]{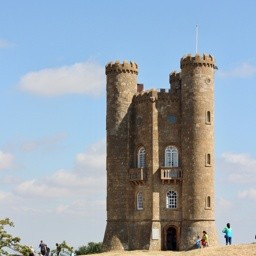} &
    \multirow{1}{*}[1.35cm]{\rotatebox[origin=c]{90}{ w/ MIMA}}
    \includegraphics[height=1.5cm]{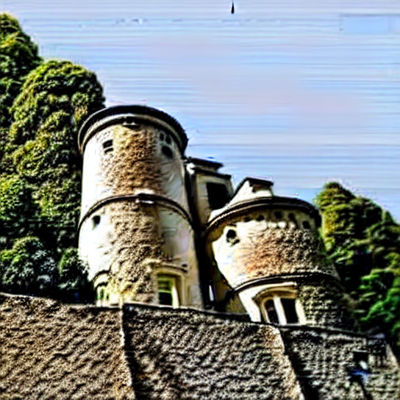} & 
    \includegraphics[height=1.5cm]{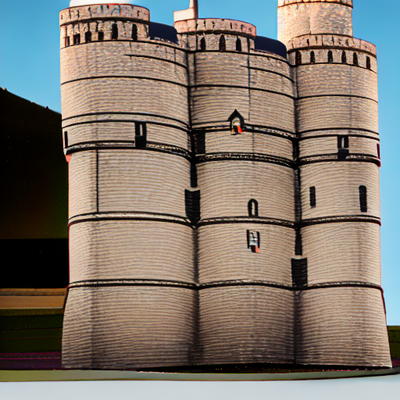} & 
    \includegraphics[height=1.5cm]{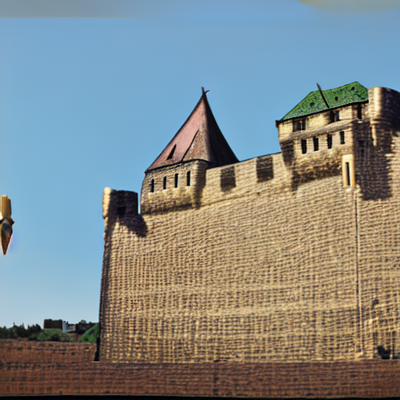} & 
    \includegraphics[height=1.5cm]{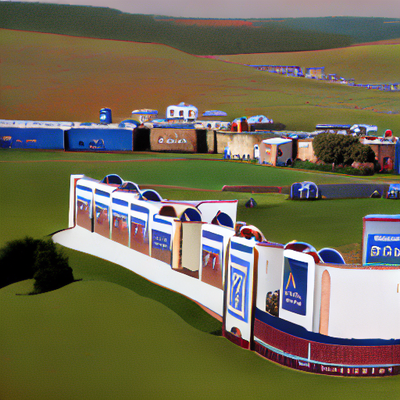} \\
    \includegraphics[height=1.5cm, width=1.7cm]{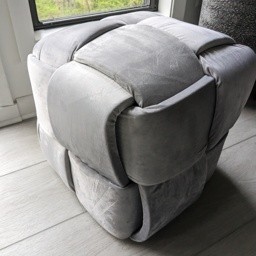} &
     \multirow{2}{*}[1.35cm]{\rotatebox[origin=c]{90}{ w/o Immu.}}
     \includegraphics[height=1.5cm]{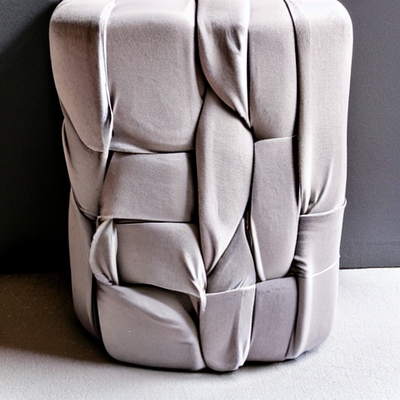} & 
    \includegraphics[height=1.5cm]{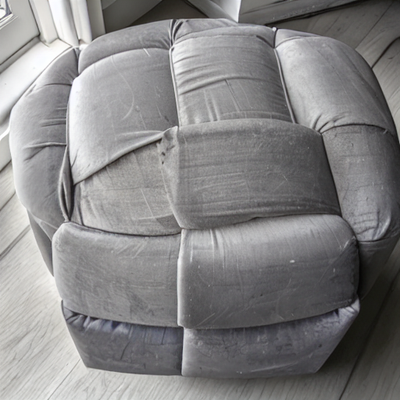} & 
    \includegraphics[height=1.5cm]{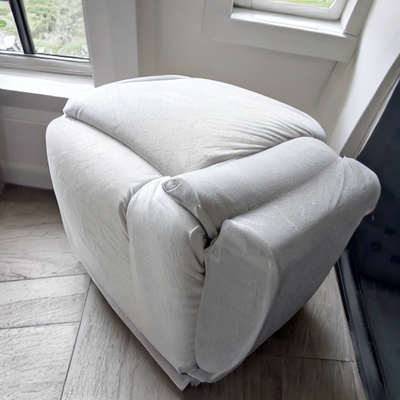} & 
    \includegraphics[height=1.5cm]{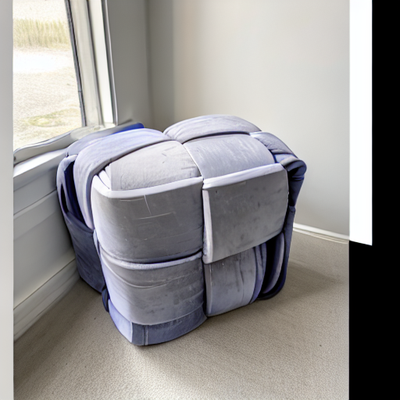} \\
    \includegraphics[height=1.5cm, width=1.7cm]{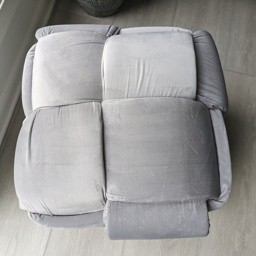} &
     \multirow{2}{*}[1.35cm]{\rotatebox[origin=c]{90}{ w/ MIMA}}
     \includegraphics[height=1.5cm]{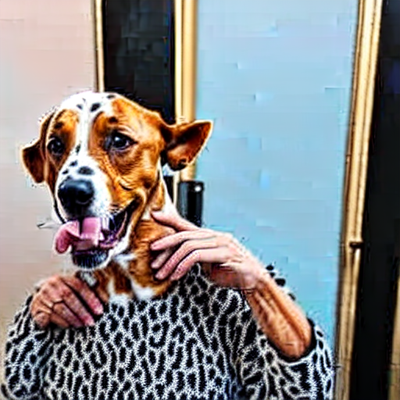} & 
    \includegraphics[height=1.5cm]{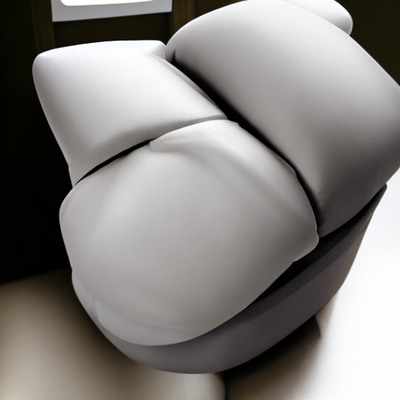} & 
    \includegraphics[height=1.5cm]{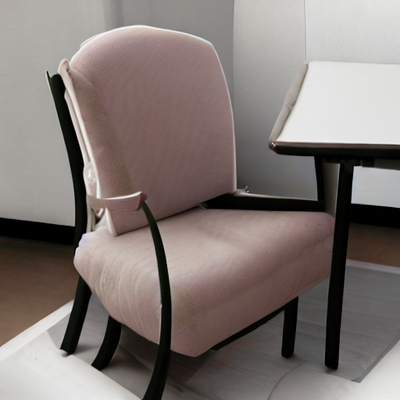} & 
    \includegraphics[height=1.5cm]{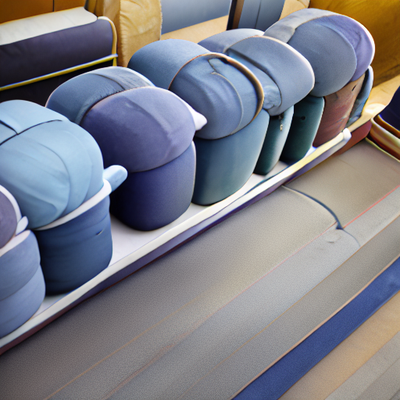} \\
    \includegraphics[height=1.5cm, width=1.7cm]{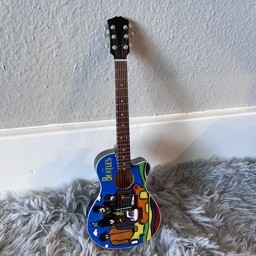} &
     \multirow{2}{*}[1.35cm]{\rotatebox[origin=c]{90}{ w/o Immu.}}
     \includegraphics[height=1.5cm]{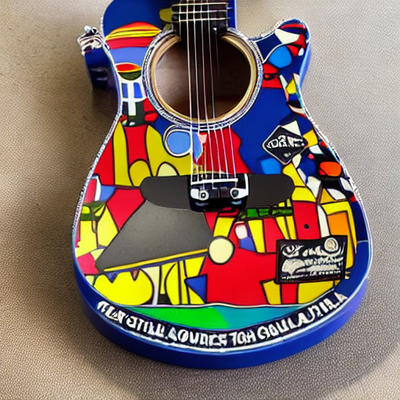} & 
    \includegraphics[height=1.5cm]{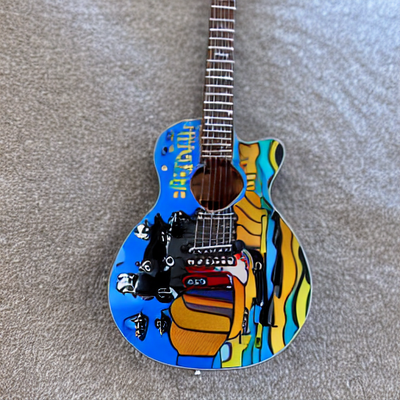} & 
    \includegraphics[height=1.5cm]{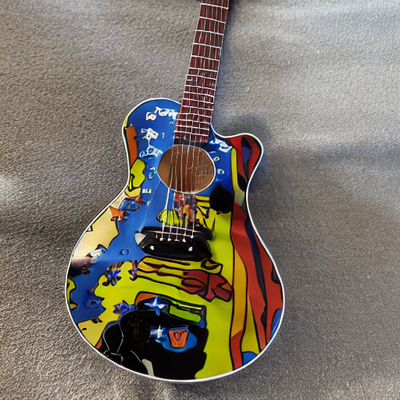} & 
    \includegraphics[height=1.5cm]{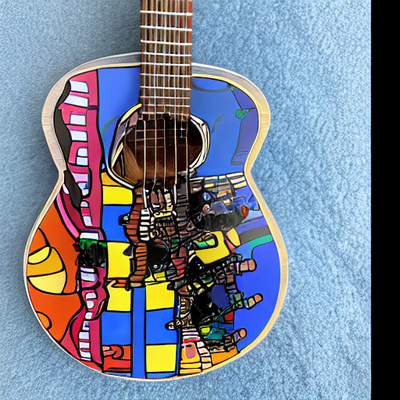} \\
    \includegraphics[height=1.5cm, width=1.7cm]{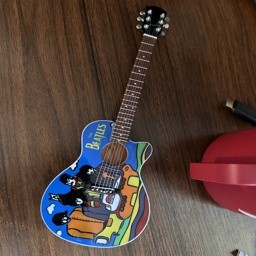} &
     \multirow{2}{*}[1.35cm]{\rotatebox[origin=c]{90}{ w/ MIMA}}
     \includegraphics[height=1.5cm]{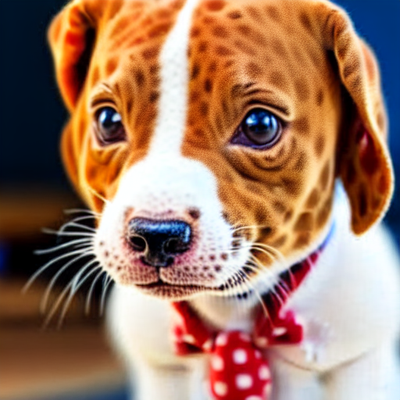} & 
    \includegraphics[height=1.5cm]{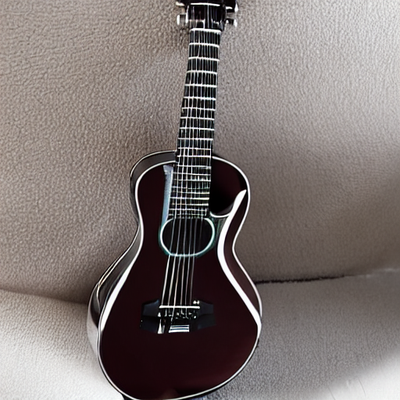} & 
    \includegraphics[height=1.5cm]{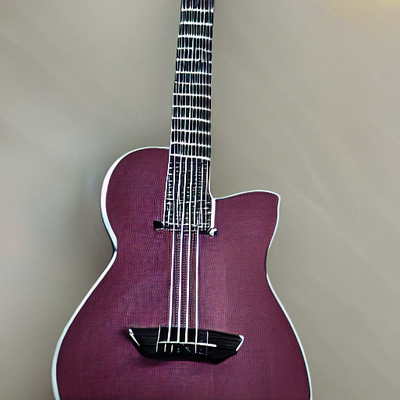} & 
    \includegraphics[height=1.5cm]{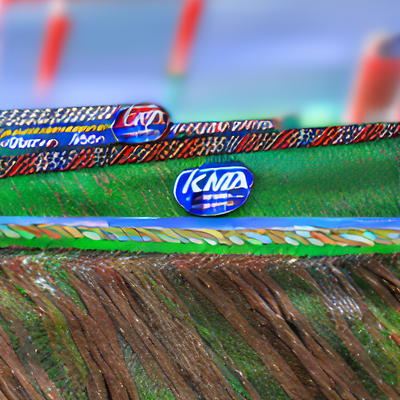} \\
    \end{tabular}
    \vspace{-0.1cm}
    \caption{{Additional personalization adaptation w/ and w/o MIMA against three concepts. %
    }
    }
    \label{fig:supp_qual_per2}
    \vspace{-0.3cm}
\end{figure}

\noindent \myparagraph{Additional qualitative results of MIMA.}
We also provide additional qualitative results against style and object relearning in~\figref{fig:supp_qual_erase_style} and~\figref{fig:supp_qual_erase_object}. The result of MIMA against personalization is in~\figref{fig:supp_qual_per} and \figref{fig:supp_qual_per2}. We can observe that MIMA successfully protects the model against multi-concept re-learning and personalization.

\begin{figure}[h]
\centering
    \setlength{\tabcolsep}{0.1pt}
    \begin{tabular}{lcccc}
      \includegraphics[height=1.55cm, trim={0 0.35cm 0 0.35cm},clip]{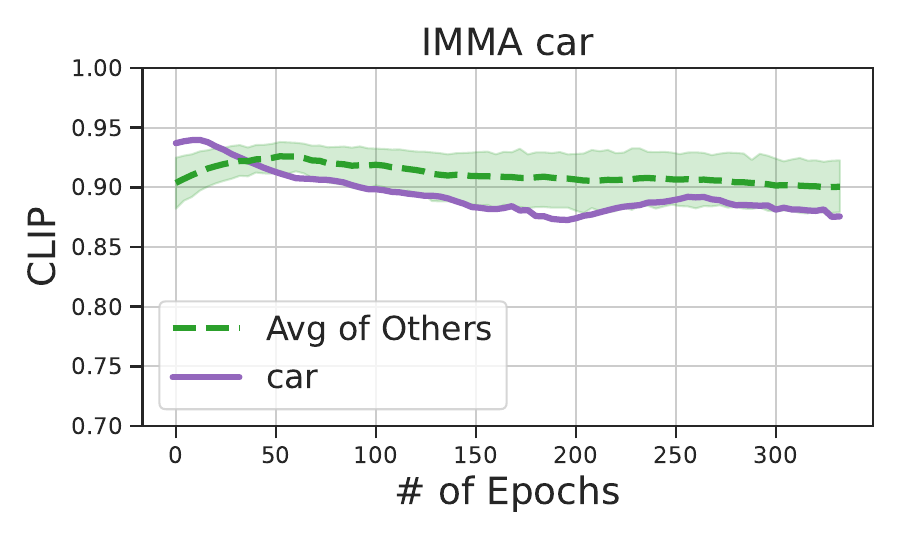} & \includegraphics[height=1.55cm, trim={0 0.35cm 0 0.35cm},clip]{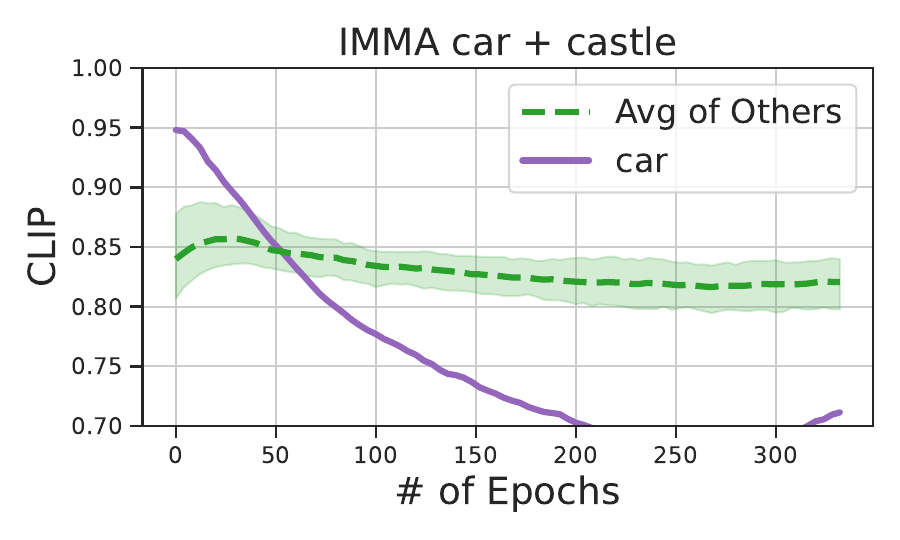} &\includegraphics[height=1.55cm, trim={0 0.35cm 0 0.35cm},clip]{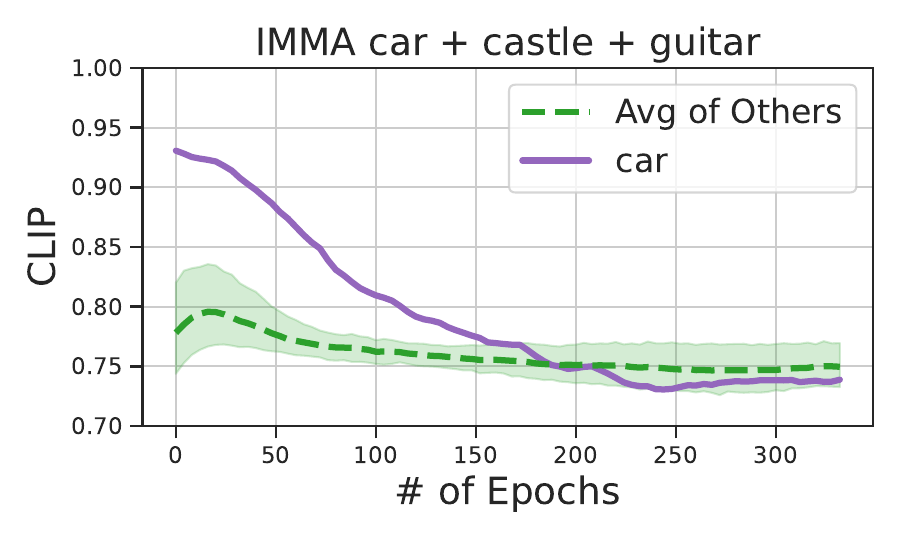} \\
     \includegraphics[height=1.55cm, trim={0 0.35cm 0 0.35cm},clip]{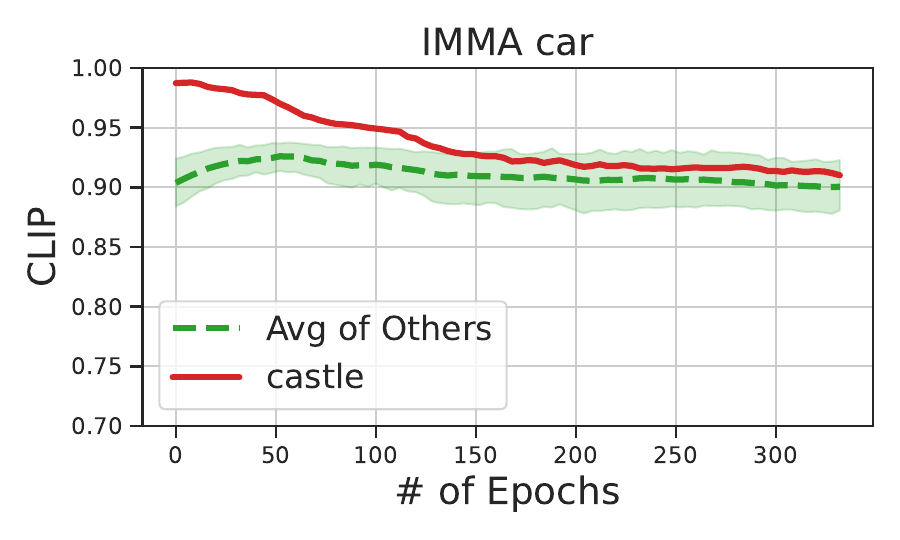} & \includegraphics[height=1.55cm, trim={0 0.35cm 0 0.35cm},clip]{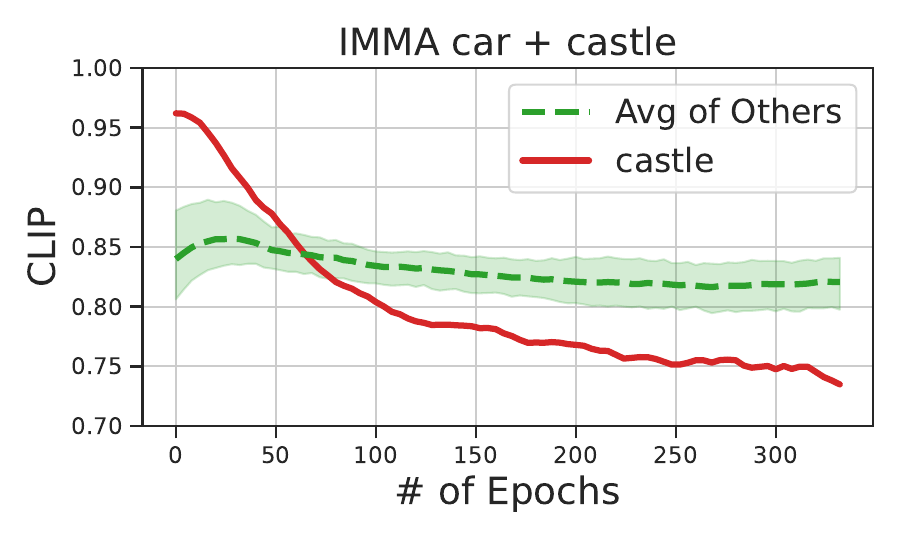}& \includegraphics[height=1.55cm, trim={0 0.35cm 0 0.35cm},clip]{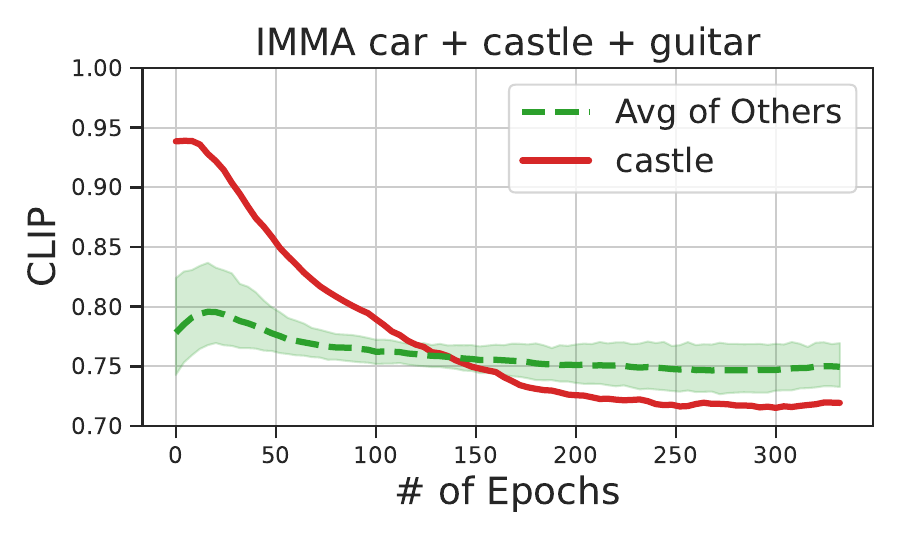}\\
     \includegraphics[height=1.55cm, trim={0 0.35cm 0 0.35cm},clip]{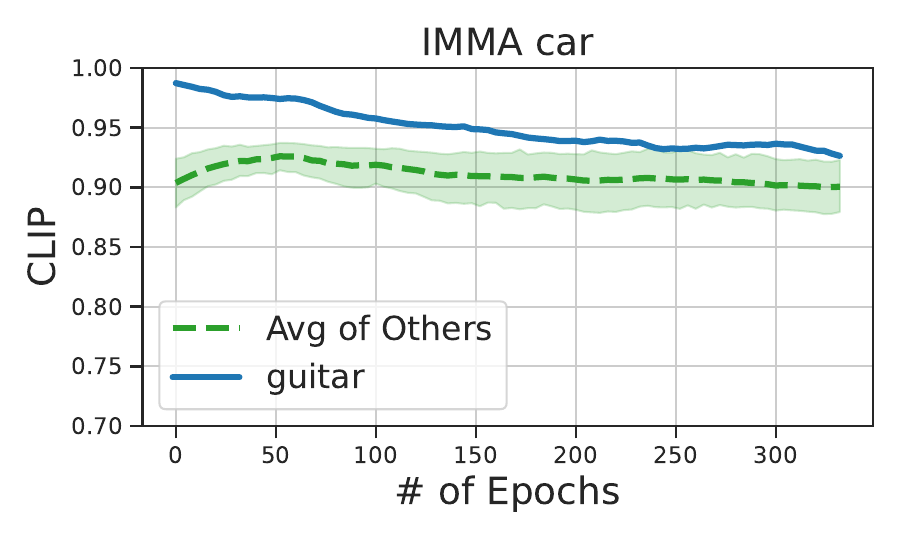} & \includegraphics[height=1.55cm, trim={0 0.35cm 0 0.35cm},clip]{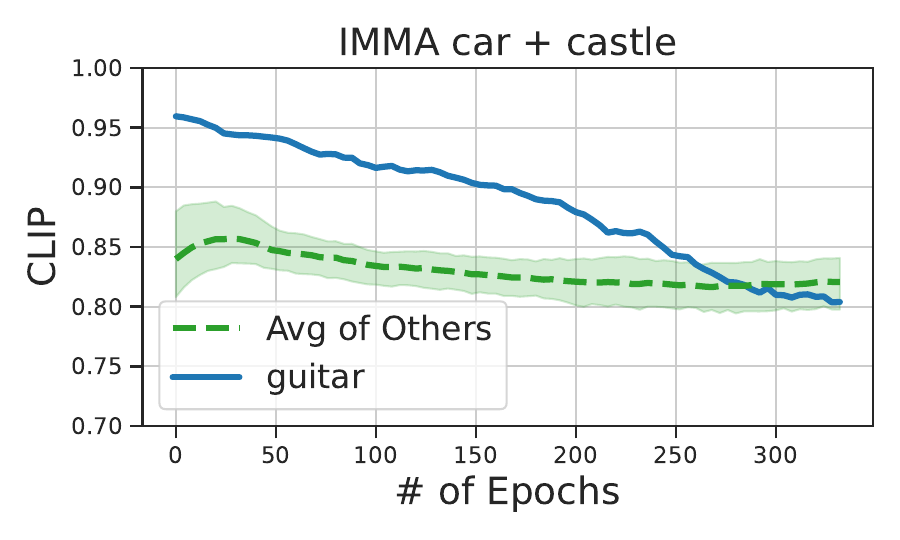} & \includegraphics[height=1.55cm, trim={0 0.35cm 0 0.35cm},clip]{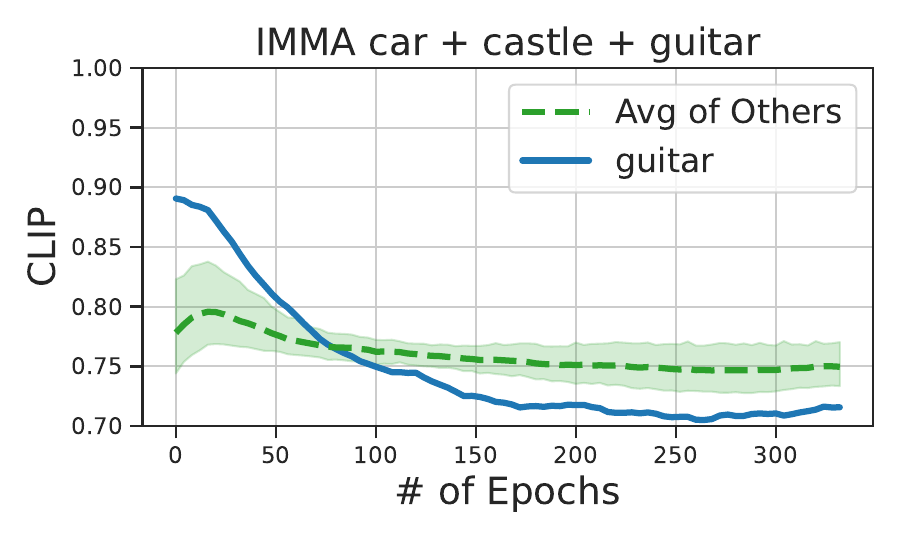} \\
    \end{tabular}
    \vspace{-0.4cm}
    \captionof{figure}{{CLIP similarity on other concepts vs. target concept with IMMA sequentially.} Each column shows IMMA protecting the concept of car, car + castle, and car + castle + guitar.
    }
    \label{fig:seq_imma_rsgr}
    \vspace{-0.4cm}
\end{figure}

\noindent \paragraph{An additional baseline of Sequential IMMA.} We provide results of running IMMA sequentially on \textit{car, castle} and \textit{guitar}. In~\figref{fig:seq_imma_rsgr}, we can observe that when running IMMA multiple times on different concepts, the CLIP similarity between generations with and without IMMA will drop both on the target concept and other concepts. This indicates that sequentially running IMMA will degrade the performance of generating other concepts and is not a promising approach. 

\section{Derivation of model merging layer}
\label{sec:derivation}
We provide a derivation for obtaining the differentiable model merging layer. Specifically, we show that solving the model merging optimization function~\equref{eq:optim_layer} is equivalent to solving a linear system as expressed in~\equref{eq:linear_reg_form}. 

Recall that the optimization function is formulated as follows:
\bea
\nonumber
&\varphi^\star = \argmin_{\varphi}  ||\rmC_{\text{reg}} \varphi  - \rmC_{\text{reg}} \rmW^p||_F^2 \text{~~s.t.~~}\rmC \varphi = \rmO^\ast, \\
\nonumber
&\text{~where~} \rmC \triangleq {\tt Concat}(\gC)\in \mathbb{R}^{(N \cdot l) \times c}\text{~,~} \\ 
\nonumber
&\rmC_{\text{reg}} \triangleq {\tt Concat}(\gC_{\text{reg}})\in \mathbb{R}^{(N' \cdot l) \times c}\text{~,~} \\ 
\nonumber
&\text{~~~~~~~~and~}\rmO^\ast \triangleq {\tt Concat}\left(\{\rvc_{[n]}\rmW_{[n]}\}_{n=1}^N\right) \in \mathbb{R}^{(N \cdot l) \times d}.
\eea
The Lagrange form of this optimization problem is given by:
\[
L(\varphi, \rmM) = ||\rmC_{\text{reg}} \varphi  - \rmC_{\text{reg}} \rmW^p||_F^2 - \text{tr}((\rmC \varphi - \rmO^\ast)\rmM^\top)
\]
where $\rmM \in \mathbb{R}^{(N \times l) \times d}$ represents the matrix of Lagrange multipliers associated with the constraint. 

To find the optimal value of \(\varphi\), we take the derivative of the Lagrangian \(L(\varphi, \rmM)\) with respect to \(\varphi\) and set it equal to zero:
\bea
\frac{\partial L(\varphi, \rmM)}{\partial \varphi} = 2\rmC_{\text{reg}}^{\top} \rmC_{\text{reg}} \varphi - 2\rmC_{\text{reg}}^{\top} \rmC_{\text{reg}} W^{p} - \rmC^{\top} \rmM = 0
\eea
Simplifying, we obtain:
\bea
\rmC_{\text{reg}}^{\top} \rmC_{\text{reg}} \varphi = \rmC_{\text{reg}}^{\top} \rmC_{\text{reg}} W^{p} + \frac{1}{2} \rmC^{\top} \rmM
\eea
By letting $\mQ \triangleq \rmC_{\text{reg}}^\top \rmC_{\text{reg}}
~\text{ and }~
\vt \triangleq  \rmC_{\text{reg}}^\top \rmC_{\text{reg}} \rmW^p + \frac{1}{2}\rmC^\top \rmM$, we can obtain $\varphi^\star$ by solving the linear system
\bea
\mQ \varphi = \vt
\eea
Therefore, we have:
\bea
\varphi^\star = \rmQ^{-1}\rvt = \left(\rmC_{\text{reg}}^{\top} \rmC_{\text{reg}}\right)^{-1} \left(\rmC_{\text{reg}}^{\top} \rmC_{\text{reg}} W^{p} + \frac{1}{2} \rmC^{\top} \rmM\right).
\eea
Next, to show that $\rmM$ is in the form of~\equref{eq:optim_m}, we substitute the expression for \(\varphi^\star\) into the constraint \(\rmC \varphi = \rmO^\star\):
\bea
\rmC \varphi^\star = \rmC \left(\rmC_{\text{reg}}^{\top} \rmC_{\text{reg}}\right)^{-1} \left(\rmC_{\text{reg}}^{\top} \rmC_{\text{reg}} W^{p} + \frac{1}{2} \rmC^{\top} \rmM\right) = \rmO^\star.
\eea
Thus, the solution for \(\rmM\) is:
\bea
\rmM = 2\left(\rmC(\rmC_{\text{reg}}^\top\rmC_{\text{reg}})C^\top\right)^{-1}(\rmO^* - \rmC \rmW^p).
\eea

\section{Experiment details}
\label{sec:exp_detail}
Our implementation uses the public codebase of Custom Diffusion~\cite{kumari2022customdiffusion}. The pre-trained diffusion model is downloaded from the checkpoint of Stable Diffusion V1-4 (\url{https://huggingface.co/CompVis/stable-diffusion-v1-4}).

\myparagraph{Training details.}
For both tasks, the lower-level learning rate is carefully chosen. In the training of MIMA against personalization, we use a learning rate of \(2 \times 10^{-5}\) with 500 training steps and a batch size equal to the number of target concepts, ensuring that each batch includes at least one training image for each target concept. The training epoch of 500 is selected from the range \(\{100, 200, 300, 400, 500, 600, 700, 800, 900, 1000\}\), and the learning rate is chosen from \(\{1 \times 10^{-5}, 2 \times 10^{-5}, 3 \times 10^{-5}, 5 \times 10^{-5}\}\), balancing the trade-off between \texttt{MSGR} and \texttt{MRSGR}. For the MIMA training against re-learning, the learning rate is \(3 \times 10^{-5}\), with 300 training steps and a batch size of 1. The hyperparameters for MIMA against personalization are chosen from the same pool as those used in the experiments against re-learning. In each experiment, we fix the random seed and run the process once to obtain the results reported in the tables.

\myparagraph{Datasets.} Our datasets are collected from the following resources: (i) ImgaeNet~\cite{deng2009imagenet} as in UCE~\cite{gandikota2024unified} for object relearning. (ii) Eight artistic styles as in UCE~\cite{gandikota2024unified} for style relearning. (iii) CustomConcept101~\cite{kumari2022customdiffusion} for personalization adaptation.
We provide the concept names of each concept set in~\tabref{tab:concept_names}.
\begin{table*}[t]
    \centering

    \begin{tabular}{lcccc}
    \specialrule{.15em}{.05em}{.05em}
    && Re-learning (Style) & Re-learning (Object) & Personalization \\ \hline
    \multirow{5}{*}{\rotatebox[origin=c]{90}{2-concept}} & 1   &  Monet+Fagan & chain saw+French horn & purse+plant   \\
    & 2   &  Mckernan+VanGogh & church+springer & purse+woodenpot\\
    & 3   &  Eng+Mckernan & horn+garbage truck & sofa+woodenpot \\
    & 4   &  Kinkade+Picasso & gas pump+golf ball & plant+castle\\
    & 5   &  VanGogh+Monet & parachute+tench & motorbike+lighthouse\\ \hline
    \multirow{5}{*}{\rotatebox[origin=c]{90}{3-concept}} & 1   &  Mckernan+Fagan+Picasso & cassette player+chain saw+church & glasses+castle+guitar   \\
    & 2   &  Eng+Kinkade+Mckernan &  springer+horn+garbage truck & car+glasses+castle\\
    & 3   &  Kinkade+Monet+Fagan & gas pump+golf ball+garbage truck & castle+chair+guitar \\
    & 4   &  VanGogh+Monet+Picasso & chain saw+springer+tench & woodenpot+motorbike+guitar\\
    & 5   &  Eng+Monet+Edlin & church+horn+golf ball & plant+woodenpot+guitar\\
    \specialrule{.15em}{.05em}{.05em}
    \end{tabular}
        \caption{Concept names of each concept set.}
    \label{tab:concept_names}
\end{table*}

\myparagraph{Backbone model for evaluation.} We use `ViT-B/32' for CLIP, `ViT-S/16' for DINO, and AlexNet for LPIPS following standard practice.

\begin{figure}[h]
    \centering
    \includegraphics[width=\linewidth]{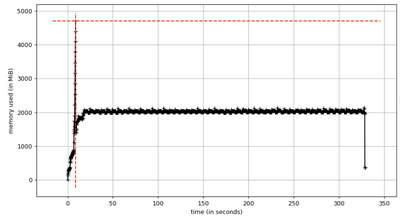}
    \caption{Memory usage over time during the experiment. The plot illustrates the memory consumption (in MiB) as the code progresses, showing a rapid increase at the start, stabilizing around 2,000 MiB, with a peak near 5,000 MiB.}
    \label{fig:memory}
\end{figure}

\myparagraph{Computational efficiency analysis.}
All experiments were conducted on one single NVIDIA A100 GPU in Ubuntu. We provide the running time and memory usage of running 100 steps in~\figref{fig:memory} where each step includes three concepts and one image for each concept. As shown, memory usage stabilized around 2,000 MiB, with each iteration taking approximately 2.60 seconds.

\end{document}